\documentclass[letterpaper,11pt]{article}

\usepackage[margin=1in]{geometry}
\usepackage{lmodern}
\usepackage{caption}
\DeclareCaptionFont{tableeight}{\fontsize{8.0}{9.5}\selectfont}
\DeclareCaptionFont{tablesix}{\fontsize{6.0}{7.0}\selectfont}

\usepackage{amsmath,amsfonts,bm}

\def\eqref#1{equation~\ref{#1}}

\def\1{\bm{1}}

\DeclareMathAlphabet{\mathsfit}{\encodingdefault}{\sfdefault}{m}{sl}
\SetMathAlphabet{\mathsfit}{bold}{\encodingdefault}{\sfdefault}{bx}{n}

\newcommand{\sigmoid}{\sigma}

\usepackage[utf8]{inputenc}
\usepackage[T1]{fontenc}
\usepackage{amsmath,amssymb,amsthm,mathtools}
\usepackage{ifthen,bbm,nicefrac,xfrac}
\usepackage{scalerel,accents,fix-cm}
\usepackage{changepage,float}
\usepackage[shortlabels]{enumitem}
\usepackage{csquotes,comment}

\usepackage{graphicx,array,booktabs,tabularx,multirow,nicematrix}
\usepackage[caption=false]{subfig}
\usepackage{xcolor,tcolorbox}

\usepackage{caption} 
\usepackage{tikz,pgfplots}
\usepgfplotslibrary{groupplots}

\usetikzlibrary{
    patterns,
    intersections,
    arrows,
    arrows.meta,
    positioning,
    decorations.markings
}
\usepgfplotslibrary{fillbetween}

\usepackage[suppress]{color-edits}
\addauthor{cp}{cyan}
\addauthor{wt}{purple}
\addauthor{rx}{blue}
\addauthor{ai}{red}

\usepackage[ruled,vlined,linesnumbered]{algorithm2e}

\usepackage{natbib,thm-restate}

\usepackage[
    colorlinks=true,
    linkcolor=blue!70!black,
    citecolor=blue!70!black,
    urlcolor=CUHKgold,
    breaklinks=true
]{hyperref}

\usepackage{cleveref}

\usepackage{soul}

\theoremstyle{plain}

\newtheorem{definition}{Definition}[section]

\crefname{claim}{claim}{claims}
\Crefname{algocf}{Algorithm}{Algorithms}

\allowdisplaybreaks

\newcommand{\xhdr}[1]{%
    \par\vspace{0.1em}%
    \noindent\textbf{#1.}\enspace\ignorespaces}

\newcommand{\squishlist}{%
    \begin{itemize}[
        itemsep=3pt,
        parsep=1pt,
        topsep=1pt,
        partopsep=0pt,
        leftmargin=1em,
        labelsep=0.5em
    ]}

\newcommand{\squishend}{\end{itemize}}

\definecolor{highlightred}{RGB}{180,0,0}

\tikzset{
    vecArrow/.style={
        thick,
        decoration={
            markings,
            mark=at position 1 with {
                \arrow[semithick]{open triangle 60}
            }
        },
        double distance=1.4pt,
        shorten >=5.5pt,
        preaction={decorate},
        postaction={
            draw,
            line width=1.4pt,
            white,
            shorten >=4.5pt
        }
    },
    innerWhite/.style={
        semithick,
        white,
        line width=1.4pt,
        shorten >=4.5pt
    }
}

\newcommand{\binIdx}{b}

\newcommand{\numBins}{B}

\newcommand{\QuestionSet}{\mathcal{Q}}

\newcommand{\evalSet}{\mathcal{I}}

\newcommand{\binTargets}{\mathcal{B}}

\newcommand{\forecastOutcome}{y}

\newcommand{\forecastProb}{p}
\newcommand{\forecastVector}{\boldsymbol{\forecastProb}}

\newcommand{\evalForecast}{\forecastVector}

\newcommand{\aggRule}{f}

\newcommand{\brierScore}[2][]{%
    \text{\normalfont\scshape BS}%
    \ifthenelse{\equal{#1}{}}{}{_{#1}}%
    \!\left[#2\right]}

\newcommand{\brierIndex}[2][]{%
    \text{\normalfont\scshape BI}%
    \ifthenelse{\equal{#1}{}}{}{_{#1}}%
    \!\left[#2\right]}

\newcommand{\ECE}[2][]{%
    \text{\normalfont\scshape ECE}%
    \ifthenelse{\equal{#1}{}}{}{_{#1}}%
    \!\left[#2\right]}

\newcommand{\logit}{\text{\normalfont\scshape logit}}
\renewcommand{\sigmoid}{\sigma}

\newcommand{\aggWeight}{w}

\providecommand{\priormean}{\mu}

\newcommand{\outcome}{\forecastOutcome}   

\newcommand{\loss}[2][]{\ell\ifthenelse{\equal{#1}{}}{}{_{#1}}\!\left(#2\right)}

\newcommand{\coveredEvents}[1]{\mathcal{E}_{#1}}
\newcommand{\modelSubset}{S}
\newcommand{\modelUniverse}{\mathcal{M}}
\newcommand{\groupThreshold}{N}

\newcommand{\precision}{\phi}
\newcommand{\bayesRule}{g}

\definecolor{resultshade}{RGB}{242,236,247}

\usepackage{array}
\usepackage{booktabs}
\usepackage{longtable}
\usepackage{makecell}
\usepackage{pdflscape}

\newcommand{\questionSymbol}{Q}
\newcommand{\questionIdx}{i}
\newcommand{\subquestionIdx}{j}
\newcommand{\numQuestions}{n}
\newcommand{\numSubquestions}{s}

\newcommand{\forecasterIdx}{m}
\newcommand{\poolSize}{k}
\newcommand{\candidatePool}{P}
\newcommand{\numEvalSubquestions}{N}

\newcommand{\robustCoeff}{\alpha}
\newcommand{\quasiMeanRule}{f}
\newcommand{\meanGenerator}{g}
\newcommand{\meanArgument}{x}
\newcommand{\realNumbers}{\mathbb{R}}

\newcommand{\eceSumIdx}{b}

\newcommand{\sigmoidArgument}{z}

\usetikzlibrary{backgrounds}
\newsavebox{\appendixtablebox}

\newcommand{\fitappendixtable}[1]{%
  \sbox{\appendixtablebox}{\input{#1}}%
  \ifdim\wd\appendixtablebox>\linewidth
    \resizebox{\linewidth}{!}{\usebox{\appendixtablebox}}%
  \else
    \resizebox{\wd\appendixtablebox}{!}{\usebox{\appendixtablebox}}%
  \fi
}

\usepackage{xurl}

\newcommand{\setsize}[1]{\left|#1\right|}
\newcommand{\abs}[1]{\left|#1\right|}

\newcommand{\prob}[2][]{%
    \text{Pr}%
    \ifthenelse{\equal{#1}{}}{}{_{#1}}%
    \!\left[
        \def\given{\middle|}
        \let\givenn\given
        #2
    \right]}

\newcommand{\expect}[2][]{%
    \mathbb{E}%
    \ifthenelse{\equal{#1}{}}{}{_{#1}}%
    \!\left[
        \def\given{\middle|}
        \let\givenn\given
        #2
    \right]}

\newcommand{\tparen}{\big}

\newcommand{\tprob}[2][]{%
    \text{Pr}%
    \ifthenelse{\equal{#1}{}}{}{_{#1}}%
    \tparen[{
        \def\given{\tparen|}
        \let\givenn\given
        #2
    }\tparen]}

\newcommand{\texpect}[2][]{%
    \mathbb{E}%
    \ifthenelse{\equal{#1}{}}{}{_{#1}}%
    \tparen[{
        \def\given{\tparen|}
        \let\givenn\given
        #2
    }\tparen]}

\newcommand{\sprob}[2][]{%
    \text{Pr}%
    \ifthenelse{\equal{#1}{}}{}{_{#1}}%
    [#2]}

\newcommand{\sexpect}[2][]{%
    \mathbb{E}%
    \ifthenelse{\equal{#1}{}}{}{_{#1}}%
    [#2]}

\usepackage{graphicx}

\usepackage{tikz}
\usepackage{pgfplots}
\usetikzlibrary{plotmarks}
\pgfplotsset{compat=1.18}

\definecolor{fbpurple}{HTML}{4F207F}
\definecolor{fbgold}{HTML}{E6A400}
\definecolor{fbteal}{HTML}{1F788A}
\definecolor{fbblue}{HTML}{2F5597}
\definecolor{fbbrick}{HTML}{B24745}
\usepackage{booktabs}
\usepackage{multirow}
\usepackage{graphicx}
\usepackage[table]{xcolor}

\definecolor{forecastpurple}{HTML}{4F207F}
\providecolor{forecastgold}{HTML}{EFAB02}
\providecolor{forecastteal}{HTML}{148A8A}
\providecolor{forecastred}{HTML}{C83E3E}
\providecolor{forecastpurple}{HTML}{4F207F}
\providecolor{forecastgold}{HTML}{EFAB02}
\providecolor{forecastteal}{HTML}{148A8A}
\providecolor{forecastred}{HTML}{C83E3E}
\providecolor{forecastblue}{HTML}{2F6BBA}
\definecolor{forecastgray}{HTML}{969AA3}
\definecolor{CUHKpurple}{HTML}{9B26B6}  
\definecolor{CUHKgold}{HTML}{FFB81C}    

\colorlet{bsgrid}{black!10}
\colorlet{bslinear}{forecastpurple}
\colorlet{bslogodds}{forecastgold!90!black}
\colorlet{bsindividual}{forecastblue!50!white}

\pgfplotsset{
  forecast axes front/.style={
    set layers=standard,
    axis line style={black!65,/pgfplots/on layer=axis foreground},
    tick style={black!65,/pgfplots/on layer=axis foreground},
  },
}

\definecolor{o-green}{HTML}{10A37F}
\definecolor{a-orange}{HTML}{D97757}
\definecolor{meta-blue}{HTML}{0668E1}
\definecolor{ali-purple}{HTML}{615CED}
\definecolor{mistral-orange}{HTML}{FA520F}

\setcitestyle{authoryear,round,citesep={;},aysep={,},yysep={;}}
\hypersetup{
    linkcolor=blue!70!black,
    citecolor=blue!70!black,
    urlcolor=blue,
    pdfauthor={Cheng Peng, Ruixi Luo, Zhi Chen, Wei Tang},
    pdftitle={Two Heads Are Better Than One: Aggregating Weaker LLMs for Better Forecasts}
}
\renewcommand{\xhdr}[1]{\paragraph{#1.}}

\title{Two Heads Are Better Than One:\\
Aggregating Weaker LLMs for Better Forecasts}

\author{
Cheng Peng\thanks{Chinese University of Hong Kong.
Email: \texttt{chengpeng@link.cuhk.edu.hk}}
\and
Ruixi Luo\thanks{Chinese University of Hong Kong.
Email: \texttt{luorx@link.cuhk.edu.hk}}
\and
Zhi Chen\thanks{Chinese University of Hong Kong.
Email: \texttt{zhi.chen@cuhk.edu.hk}}
\and
Wei Tang\thanks{Chinese University of Hong Kong.
Email: \texttt{weitang@cuhk.edu.hk}}
}
\date{}

\begin{document}

\maketitle

\begin{abstract}
Large language models (LLMs) are increasingly used to forecast real-world events, but access to the strongest individual forecaster may be costly or otherwise constrained. 
We study weak-to-strong forecast aggregation: can individually weaker LLM forecasters be aggregated to outperform a stronger forecaster? 
Using ForecastBench \citep{KBYJ-25}, we evaluate 70 LLM forecasters across 16 comparison groups, each with more than 1,000 shared subquestions, yielding 1,121 weaker-model pairs. 
Within each group, we identify the strongest individual by test Brier score and evaluate aggregates composed exclusively of weaker forecasters, with aggregation weights learned on separate training data. 
We find substantial evidence of weak-to-strong improvement. 
Learned linear pooling identifies a weaker pair that matches or outperforms the strongest individual in 11 of 16 groups and comes within 5\% of its Brier score in all 16 groups. 
We also find that these improvements do not rely on having a near-best constituent and are generally accompanied by good calibration. 
Additional analyses show that adding more models does not consistently improve performance, and competitive weaker-model aggregates also remain available under practical constraints. 


\end{abstract}

\clearpage

\section{Introduction}
\label{sec:intro}
Forecasting is an important and fundamental task in modern economic decision-making, informing decisions ranging from production and inventory planning to investment, financial markets, and public policy, etc. 
Accordingly, how to produce good forecasts has been studied for decades across statistics, economics, operations research, and machine learning (see, e.g., \citealp{W-60,A-01,H-04,GFQW-23,GSC-24}).
While much of this classical literature focuses on forecasting future values from historical data, many important decisions instead require forecasts about future real-world events, 
for which relevant information may be distributed across heterogeneous textual, numerical, and contextual sources. 

Recent remarkable advances in large language models (LLMs), which are trained on large-scale text corpora, have created new possibilities for automated forecasting of future events.
LLMs can draw on broad cross-domain knowledge, reason over textual evidence, and, when combined with retrieval or search, incorporate up-to-date information relevant to future events.
These capabilities have motivated a rapidly growing literature on LLM-based forecasting, with recent work developing and evaluating them on real-world future events (see, e.g., \citealp{ZXJ-22,HZYS-24,KBYJ-25,D-26,YML-26,ZLC-26}).

\wtedit{As more LLM forecasters become available, however, the relevant question need not be which individual forecaster is the most accurate.} 
When forecasts from multiple models are available, we can instead combine them into a single forecast.
Forecast aggregation has a long history in statistics and forecasting, where combining predictions from different sources can improve predictive accuracy \citep{BG-69,C-89}. 
Recent work finds similar benefits for LLM forecasters. 
\citet{STP-24}, for example, aggregate probabilistic forecasts from twelve LLMs and show that the resulting LLM crowd can achieve accuracy comparable to a human forecasting crowd. 
ForecastBench also considers ensemble forecasts constructed from multiple LLMs and prompting strategies \citep{KBYJ-25}. 
More recent studies further show that individual forecasting accuracy alone does not determine a model's value in an aggregate: weaker but more diverse forecasters may contribute complementary information when their forecasts differ from those of stronger models \citep{AJSD-26}, and learned aggregation methods can outperform individual LLM forecasters and classic aggregation rules \citep{D-26}. 

\wtedit{These findings suggest that what a collection of LLM forecasters can achieve may not be fully captured by their individual performance rankings.}
This distinction becomes particularly important when access to the strongest individual forecaster cannot be taken for granted.
A stronger model may be more expensive to use, available only through an external API, or incompatible with settings that require open-weight or locally deployable models. 
In such cases, the relevant question is not whether combining several forecasts, including the strongest one, can further improve performance. Rather, we ask a more stringent question of weak-to-strong aggregation: {\em  Can two or more LLM forecasters that are individually weaker than a stronger forecaster be aggregated to match or outperform that stronger forecaster?}

The possibility of such weak-to-strong improvement is suggested by a broader literature showing that weaker models can collectively outperform stronger ones.
Classical boosting results show that weak learners can be combined to construct a substantially stronger model \citep{S-90}. 
Related work on LLMs similarly shows that combining multiple outputs or weaker models can improve on \aiedit{the performance} of stronger individual models (see, e.g., \citealp{WW-22,DLTTM-24,HLLC-25}).
These results provide evidence that individual model strength does not necessarily determine the strength of an aggregate. 
Yet whether this weak-to-strong phenomenon extends to probabilistic forecasting of real-world events remains largely unexplored. 
We therefore systematically study whether proper aggregation can turn two individually weaker LLM forecasters into a forecast that matches or outperforms a stronger individual forecaster.


\xhdr{Our contributions and results}
We summarize our main contributions and results as follows:
\begin{itemize}[
    leftmargin=1.2em,
    labelsep=0.4em,
    itemsep=1pt,
    topsep=2pt,
    parsep=0pt,
    partopsep=0pt
]
    \item We introduce and systematically study weak-to-strong forecast aggregation for LLMs: whether individually weaker LLM forecasters can be aggregated to match or outperform a stronger forecaster without using its forecast. 
    Using ForecastBench \citep{KBYJ-25}, we organize model configurations into representative forecasters, construct comparison groups with sufficient shared forecast coverage, identify the strongest individual in each group as the benchmark, and evaluate aggregates formed only from weaker forecasters. 
    Our main experiments cover $70$ LLM forecasters, $16$ comparison groups, each with more than $1,000$ shared subquestions, and $1,121$ weaker-model pairs, and compare both learned and fixed aggregation rules.

    \item We find substantial evidence of weak-to-strong improvement. 
    At the group level, there exists an aggregation rule (i.e., learned linear pooling) \aiedit{that} identifies a weaker pair that outperforms the strongest model in 11 of 16 groups and comes within 5\% of its Brier score in all 16 groups. 
    Moreover, we find that these gains do not rely on having a near-best constituent and are generally accompanied by good calibration. 
    Our additional analysis shows that learning the aggregation weights matters, while adding more models does not consistently improve performance; 
    the gains also vary across question categories and remain available under lower-input-price and open-weight constraints. 
    Finally, our robustness checks, control experiment, and case study show that the weak-to-strong pattern persists under a substantially higher shared-coverage threshold, cannot generally be explained by individual recalibration alone, and can depend on the choice of aggregation rule.
\end{itemize}
\xhdr{Paper organization}
\Cref{sec:prelim} introduces the forecasting setting, evaluation metrics, and aggregation rules. 
\Cref{sec:experimental-setup} describes the dataset, construction of comparison groups, and experimental protocol. 
In \Cref{sec:main-results}, we present our main results, including pairwise and multi-model aggregation, and category-specific analysis. 
\Cref{sec:conclusion} concludes. 
The appendix provides 
\cpedit{robustness checks}, practical-constraint analysis, case studies, and supplementary results.

\subsection{Further Related Work}
\label{apx:related}

\xhdr{LLMs' forecasting and benchmarking}
Recent work has explored the use of large language models for automated forecasting of real-world events
\citep{JKK-21,ZXJ-22,SP-23,FPT-24,HZYS-24,PZM-24,PBCM-24,YSH-24,AGR-25}.
Early studies generally find that off-the-shelf LLMs lag strong human forecasters, motivating methods that
improve forecasting through better prompting, information retrieval, reasoning, fine-tuning, and calibration.
In particular, \citet{HZYS-24} develop a retrieval-augmented forecasting system that combines
information retrieval, structured reasoning, and multiple model predictions, and show that the resulting
system approaches the performance of competitive human forecasting crowds.
Subsequent work further studies specific ingredients of LLM forecasting, including forecasting strategies
and reasoning \citep{PBCM-24}, retrieval of temporally appropriate information
\citep{YSH-24}, temporal generalization and the degradation of forecasting accuracy over time
\citep{DTR-25,ZCG-25}, and logical consistency of probabilistic forecasts
\citep{PSA-25}.
\citet{YML-26} further study how internalized knowledge, information sources, retrieval, and the
ability to integrate available evidence affect LLMs' predictive performance.

A related but distinct literature applies LLMs and transformer-based models to statistical time-series
forecasting
\citep{DKM-23,GFQW-23,NNSK-23,RAW-23,
DKSZ-24,GSC-24,JWM-24,WLK-24}.
Our focus, as in the event-forecasting literature above, is instead on probabilistic judgmental forecasting
of real-world events, for which relevant evidence may come from heterogeneous textual and contextual
sources rather than a well-defined historical time series.

Alongside advances in forecasting methods, a growing literature has developed increasingly challenging
benchmarks for evaluating LLMs' predictive capabilities.
ForecastQA \citep{JKK-21} is an early event-forecasting dataset based on future events described
in news articles, while Autocast \citep{ZXJ-22} incorporates questions from forecasting
competitions and prediction platforms.
Since then, benchmark construction has expanded along several dimensions, including extracting and
representing forecastable events from news and temporal information
\citep{ZCY-24,WZY-25}, developing open-ended future-event prediction tasks
\citep{GPW-26}, and constructing forecasting benchmarks that place greater emphasis on temporal
generalization and realistic information environments \citep{WBH-25}.
More recently, the literature has shifted toward dynamic and live evaluation, in part to reduce benchmark
contamination and avoid relying on uncertain model knowledge cutoffs
\citep{BWH-25,KBYJ-25,ZLC-26}.

ForecastBench \citep{KBYJ-25} continuously collects unresolved forecasting questions from
prediction markets, forecasting platforms, and real-world datasets, and evaluates LLMs and human forecasters
as these questions resolve.
FutureBench \citep{BWH-25} and FutureX \citep{ZLC-26} further extend live
evaluation toward agentic forecasting.
In particular, FutureX evaluates LLMs equipped with reasoning, search, and deep-research capabilities on
a broad range of frequently updated events drawn from diverse real-world sources.
Prophet Arena \citep{YML-26} complements these benchmarks by continuously evaluating
probabilistic forecasts on live prediction-market events under a multi-horizon and modularized forecasting
pipeline, and by assessing models along multiple dimensions, including forecasting loss, calibration, and
economic value.
Together, these benchmarks provide increasingly rich and contamination-resistant environments for comparing
LLM forecasters.
Our work builds on this literature using forecasts collected by ForecastBench, but shifts attention from
evaluating individual forecasters to understanding what can be gained by combining forecasts from multiple
LLMs.

\xhdr{Forecast aggregation}
Recent work has explored how probabilistic forecasts produced by large language models can be aggregated and how model diversity affects the performance of LLM crowds
\citep{HZYS-24,STP-24,ASK-25,KBYJ-25,AJSD-26,D-26,T-26}.
Early studies often incorporate aggregation as one component of a broader forecasting system or evaluate relatively simple pooling rules.
\citet{HZYS-24} develop a retrieval-augmented forecasting system that searches for relevant information, generates multiple forecasts, and aggregates them as part of an end-to-end forecasting pipeline.
\citet{STP-24} aggregate probabilistic forecasts from twelve LLMs using the median and show that the resulting LLM crowd achieves accuracy comparable to a human forecasting crowd.
\citet{KBYJ-25} construct LLM ensemble baselines from nine forecasts per question, obtained from three LLMs under three prompting strategies, and compare the median, geometric mean, and geometric mean of log odds.
More recently, \citet{ASK-25} develop the AIA Forecaster, in which a supervisor agent reconciles disparate forecasts for the same event as part of a system that combines agentic search, forecast reconciliation, and statistical calibration.

More recent work examines more directly which LLM forecasts should be combined and why aggregation can help.
\citet{AJSD-26} study model selection for forecasting ensembles under a fixed sampling budget and show that individual forecasting accuracy alone does not determine ensemble performance: forecasts from frontier models can be highly correlated, while somewhat weaker but more diverse models can contribute substantial complementary information.
Relatedly, \citet{JSS-26} document an accuracy--correlation effect between LLM and human forecasts, showing that more accurate LLM forecasts also tend to be more correlated with human forecast aggregates.
\citet{D-26} compare\aiedit{s} classical and learned aggregation methods using probabilistic forecasts from fifteen LLMs and find\aiedit{s} that learned logistic-regression and neural-network aggregators outperform both individual models and classical pooling methods; their symbolic-regression analysis further identifies disagreement across model predictions as an important source of the aggregation gain.
Together, these findings suggest that the value of an LLM crowd depends not only on the forecasting accuracy of its constituent models, but also on the dependence and complementarity among their predictions.
Our work builds on this literature but focuses on a more stringent question: whether two LLMs that are each individually weaker than a third, stronger model can nevertheless be aggregated to match or outperform that stronger model.

More broadly, forecast aggregation has a long history in statistics, economics, and decision theory
\citep{S-61,BG-69,GZ-86,C-89}.
For probability forecasts, the literature studies linear, logarithmic, multiplicative, and transformed pooling methods
\citep{B-82,G-84,GWZ-84,RG-10,BMT-14,SBF-14}.
\citet{RG-10} study linear pooling and show why recalibration can be necessary even when the component forecasts are calibrated.
\citet{SBF-14} develop a simple logit-based model in which a single parameter controls how strongly the aggregate is extremized, while \citet{BMT-14} identify probability-scale compression and the aggregation of partially independent information as two reasons why aggregated forecasts may benefit from extremization.
A related information-based literature explicitly models the information underlying individual forecasts and how information overlap affects aggregation
\citep{EPSU-16,SPU-16,LR-21}.
In particular, \citet{SPU-16} model partially overlapping information across forecasters and relate the appropriate degree of extremization to informational diversity.
Related work also studies what additional information must be elicited from forecasters to permit effective aggregation \citep{FCK-15}.
These approaches motivate our comparison between learned linear pooling and learned log-odds pooling.

A complementary theoretical literature studies robust forecast aggregation when the aggregator lacks full knowledge of the information structure or the dependence across information sources
\citep{ABS-18,DIL-21,LR-22,NR-22,KWW-24,FMNW-25,GHHKSY-25,GK-25,CPT-26,TZ-26}.
Our experiments complement this literature by evaluating both robust and data-driven aggregation rules on probabilistic forecasts produced by heterogeneous LLMs.

\section{Preliminaries}
\label{sec:prelim}

We study whether aggregating individually weaker forecasters can outperform a stronger forecaster. 
In this section, we describe the forecasting task and introduce our performance metric\aiedit{;} we then introduce aggregation rules considered in this work.

\subsection{Forecasting Questions and Forecasts}
\label{sec:prediction-task}

\xhdr{Forecasting questions}
We use the dataset from ForecastBench~\citep{KBYJ-25}, which includes
\emph{standard questions} and \emph{combination questions}.
We denote the question set by $\QuestionSet=\{\questionSymbol_\questionIdx\}_{ \questionIdx\in [\numQuestions]}$, where $\numQuestions$ denotes the total number of  questions in $\QuestionSet$.
Each question~$\questionSymbol_\questionIdx$ contains a number $\numSubquestions_\questionIdx$ of binary subquestions $\{\questionSymbol_{\questionIdx\subquestionIdx}\}_{\subquestionIdx\in [\numSubquestions_\questionIdx]}$, each of which resolves to Yes or No. We treat a standard question as the case $\numSubquestions_\questionIdx=1$, with $\questionSymbol_\questionIdx=\{\questionSymbol_{\questionIdx1}\}$. We use this common representation throughout the paper.

\begin{itemize}[
    leftmargin=1.2em,
    labelsep=0.4em,
    itemsep=1pt,
    topsep=2pt,
    parsep=0pt,
    partopsep=0pt
]
    \item \textbf{Standard question:}
    ``Will it rain tomorrow in City~A?''
    The question itself is its sole subquestion.

    \item \textbf{Combination question:}
    A question about tomorrow's weather in City~A and City~B may
    contain subquestions, such as ``Will it rain in both cities?''
    and ``Will it rain in City~A but not in City~B?''
\end{itemize}

The binary outcome
$\outcome_{\questionIdx\subquestionIdx}\in\{0,1\}$ equals $1$ if subquestion~$\questionSymbol_{\questionIdx\subquestionIdx}$
resolves to Yes and $0$ otherwise.

\xhdr{Forecasters and forecasts}
A \emph{forecaster} is a model configuration specified by its model
version, prompt, and information condition. For a given forecast
date and resolution date, the forecast $\forecastProb_{\questionIdx\subquestionIdx}^{\forecasterIdx}\in[0,1]$
is the probability that forecaster~$\forecasterIdx$ assigns to
$\outcome_{\questionIdx\subquestionIdx}=1$ before observing the outcome.

\subsection{Evaluation Metrics}
\label{sec:evaluation}

We evaluate forecasts along two dimensions: \emph{accuracy}, which
measures how closely predicted probabilities match realized
outcomes, and \emph{calibration}, which measures whether forecasted
probabilities agree with observed outcome frequencies. For evaluation metric\aiedit{s}, we use the Brier score for accuracy and expected calibration error (ECE) for calibration.
We evaluate all individual and aggregated forecasts in each comparison on the same pool of questions. 
We use the evaluation set $\evalSet\subseteq[\numQuestions]$ to denote the indices of the questions included in the evaluation pool. 
For each forecaster $\forecasterIdx$, we denote the vector of its forecasts for all evaluated questions and their subquestions by
$\evalForecast^{\forecasterIdx}
=\bigl(\forecastProb_{\questionIdx\subquestionIdx}^{\forecasterIdx}\bigr)
_{\questionIdx\in\evalSet,\,
\subquestionIdx\in[\numSubquestions_{\questionIdx}]}$.

\xhdr{Brier score}
The Brier score~\citep{B-50} is a scoring rule that evaluates a probability forecast against its realized outcome. 
The expected score is minimized when \aiedit{a} forecaster forecasts the true outcome probability.
We compute the Brier score for each question by averaging the squared losses of its subquestions:
\[
    \brierScore[\questionIdx]{\evalForecast^{\forecasterIdx}}
    =
    \frac{1}{\numSubquestions_\questionIdx}
    \sum\nolimits_{\subquestionIdx\in [\numSubquestions_\questionIdx]}
    \bigl(\forecastProb_{\questionIdx\subquestionIdx}^{\forecasterIdx}-\outcome_{\questionIdx\subquestionIdx}\bigr)^2~.
\]
The overall Brier score is the average across questions
$\brierScore[\evalSet]{\evalForecast^{\forecasterIdx}}
=
\frac{1}{\setsize{\evalSet}}
\sum\nolimits_{\questionIdx\in\evalSet}
\brierScore[\questionIdx]{\evalForecast^{\forecasterIdx}}$.
Here each question therefore receives equal weight, regardless of its number of subquestions. 
A lower Brier score indicates better accuracy. 

\xhdr{Expected calibration error}
A reliable forecast should match the observed frequencies\aiedit{;} 
that is, among predictions assigning probability $0.7$ to Yes, approximately $70\%$ should resolve to Yes. 
This property is referred to as \emph{calibration} in the machine learning literature (see, e.g., \citealp{GPSW-17}).
Following the literature (see, e.g., \citealp{YML-26}), we estimate calibration error by pooling predictions with similar probabilities. 
In particular, we partition $[0,1]$ into $\numBins=10$ equal-width
bins and index them by $\binIdx\in[\numBins]$. We denote the total number of evaluated subquestions by $\numEvalSubquestions
    =\sum\nolimits_{\questionIdx\in\evalSet} \numSubquestions_{\questionIdx}$.
For forecaster~$\forecasterIdx$, we define $\binTargets_{\binIdx}^{\forecasterIdx}$ as the set containing
the subquestion indices $(\questionIdx,\subquestionIdx)$ such that the forecast $\forecastProb_{\questionIdx\subquestionIdx}^{\forecasterIdx}$ falls in bin~$\binIdx$. 
For each nonempty $\binTargets_{\binIdx}^{\forecasterIdx}$ containing indices $(\questionIdx,\subquestionIdx)$, we denote the mean \aiedit{of} the corresponding forecasts $\forecastProb_{\questionIdx\subquestionIdx}^{\forecasterIdx}$ by $\overline{\forecastProb}_{\binIdx}^{\forecasterIdx}$ and the mean of their observed outcomes $\outcome_{\questionIdx\subquestionIdx}$ by $\overline{\outcome}_{\binIdx}^{\forecasterIdx}$, respectively.

The empirical ECE averages the absolute differences between these
two quantities, weighted by each bin's share of subquestions:
\[
    \ECE[\evalSet]{\evalForecast^{\forecasterIdx}}
    =
    \sum\nolimits_{\eceSumIdx\in [\numBins]}
    \frac{\setsize{\binTargets_{\binIdx}^{\forecasterIdx}}}{\numEvalSubquestions}
    \abs{
        \overline{\forecastProb}_{\binIdx}^{\forecasterIdx}
        -\overline{\outcome}_{\binIdx}^{\forecasterIdx}
    }.
\]
ECE gives each evaluated subquestion
equal weight. We note that ECE and the Brier score are not equivalent: a lower ECE does not necessarily imply a lower Brier score.

\subsection{Aggregation Rules}
\label{sec:aggregation-rules}

We aggregate multiple probability forecasts for the same
subquestion into a single forecast. We refer to the selected
set of forecasters as a \emph{pool} and denote its size by
$\poolSize$. Within a pool, we index forecasters by
$\forecasterIdx\in[\poolSize]$ and collect their forecasts in
the vector
$\forecastVector=(\forecastProb^{1},\ldots,\forecastProb^{\poolSize})$.
We then define an aggregation rule
$\aggRule:[0,1]^{\poolSize}\to[0,1]$ as a deterministic function
that maps this vector to a single probability forecast.

Our main analysis focuses on two data-driven aggregation rules, {\em linear pooling} and {\em log-odds pooling}, whose coefficients are learned from historical forecasts. 
These rules allow the contribution of each forecaster to be adapted to the candidate pool and form the basis of our main experiments. 
We introduce them first in \Cref{sec:learned-rules}. For comparison, we also consider parameter-free classical pooling rules and, for pairwise aggregation, fixed rules from the robust forecast aggregation literature~\citep{ABS-18,KWW-24,GHHKSY-25,CPT-26}.


\subsubsection{Learned Aggregation Rules}
\label{sec:learned-rules}

Our primary aggregation rules use historical data to learn a separate nonnegative coefficient for each forecaster in the pool. These coefficients allow the fitted rule to adjust both the relative contributions of the forecasts and the overall scale of their combination. 
We denote the
coefficient assigned to forecaster~$\forecasterIdx$ by
$\aggWeight_{\forecasterIdx}$ and collect the coefficients in
the vector
$\boldsymbol{\aggWeight}=(\aggWeight_1,\ldots,\aggWeight_{\poolSize})
\in\realNumbers_+^{\poolSize}$.
Using these coefficients, we consider \emph{log-odds pooling}
and \emph{linear pooling}.\footnote{
We follow the literature (see, e.g., \citealp{ABS-18}) to adopt the same boundary convention for all aggregation rules.
If at least one forecast is $0$ and none is $1$, we return $0$;
if at least one forecast is $1$ and none is $0$, we return $1$.
When both $0$ and $1$ appear, we return $0.5$.}

\xhdr{Log-odds pooling}
We combine the forecasts in log-odds space using the learned
coefficients, then convert the result back to a probability: in particular, letting $\logit(\meanArgument)=\ln(\meanArgument/(1-\meanArgument))$ and $\sigmoid(z) = 1/(1+\exp(-z))$,
we obtain 
$\aggRule^{\mathrm{log}}_{\boldsymbol{\aggWeight}}(\forecastVector)
=\sigmoid\!(
    \sum\nolimits_{\forecasterIdx\in[\poolSize]}
    \aggWeight_{\forecasterIdx}\logit(\forecastProb^{\forecasterIdx}))$.

\xhdr{Linear pooling}
We also combine the forecasts directly in probability space to obtain 
$\aggRule^{\mathrm{lin}}_{\boldsymbol{\aggWeight}}(\forecastVector)
=\min\!\{1,
    \sum\nolimits_{\forecasterIdx\in[\poolSize]}
    \aggWeight_{\forecasterIdx}\forecastProb^{\forecasterIdx}
\}$.

For each pool and each learned rule, we fit a separate coefficient
vector by minimizing the training Brier score over
$\realNumbers_+^{\poolSize}$, following the data-splitting
procedure in \Cref{sec:split-pairs}. Once fitted, we freeze the
coefficients and evaluate the same rule using Brier score and ECE on the test data. For the ECE analysis, we also maintain the
Brier-based group benchmark and candidate pools specified
in \Cref{sec:protocol-and-evaluation}. 

\subsubsection{Classical Aggregation Rules}
\label{sec:classical-rules}

We consider four classical rules that combine forecasts without
fitting parameters to historical data.

\xhdr{Quasi-arithmetic means}
We present the first three rules as
instances of the quasi-arithmetic mean~\citep{B-03,D-26}. We transform
each forecast, average the transformed values, and apply the
inverse transformation. We denote the transformation by
$\meanGenerator$, which is required to be continuous and strictly
monotone on an interval containing the forecasts. 
We write this common construction as
\[
    \quasiMeanRule_{\meanGenerator}(\forecastVector)
    =\meanGenerator^{-1}\!\left(
        \frac{1}{\poolSize}
        \sum\nolimits_{\forecasterIdx\in[\poolSize]}
        \meanGenerator(\forecastProb^{\forecasterIdx})
    \right)~.
\]
We obtain different pooling rules by choosing different transformations.
\begin{itemize}[
    leftmargin=1.2em,
    labelsep=0.4em,
    itemsep=1pt,
    topsep=2pt,
    parsep=0pt,
    partopsep=0pt
]
    \item \emph{Arithmetic mean (AM).}
    We choose $\meanGenerator(\meanArgument)=\meanArgument$ to obtain
    $\aggRule_{\mathrm{AM}}(\forecastVector)
    =\poolSize^{-1}
    \sum\nolimits_{\forecasterIdx\in[\poolSize]}
    \forecastProb^{\forecasterIdx}$. \rxedit{Conventionally, this mean is also referred to as \aiedit{the} \emph{Simple mean}.}

    \item \emph{Harmonic mean (HM).}
    We choose $\meanGenerator(\meanArgument)=1/\meanArgument$ to obtain
    $\aggRule_{\mathrm{HM}}(\forecastVector)
    =(\poolSize^{-1}
    \sum\nolimits_{\forecasterIdx\in[\poolSize]}
    1/\forecastProb^{\forecasterIdx})^{-1}$.

    \item \emph{Log-odds mean (GM).}
    This rule averages forecasts in log-odds space and converts
    the result back to a probability. We choose
    $\meanGenerator=\logit$, where
    $\logit(\meanArgument)=\ln(\meanArgument/(1-\meanArgument))$,
    and denote its inverse by
    $\sigmoid(\sigmoidArgument)=1/(1+\exp(-\sigmoidArgument))$
    for $\sigmoidArgument\in\realNumbers$. The resulting rule is
    $\aggRule_{\mathrm{GM}}(\forecastVector)
    =\sigmoid(\poolSize^{-1}
    \sum\nolimits_{\forecasterIdx\in[\poolSize]}
    \logit(\forecastProb^{\forecasterIdx}))$.
    The label GM refers to taking the geometric mean of the
    forecast odds before converting back to a probability.
\end{itemize}

\xhdr{Median}
We use the median as the fourth classical baseline when we \aiedit{aggregate} more than two models. 
Under this rule, we order the forecasts and take the middle value, averaging the two middle
values when the pool size~$\poolSize$ is even:
$\aggRule_{\mathrm{med}}(\forecastVector)
=\operatorname{median}
(\forecastProb^{1},\ldots,\forecastProb^{\poolSize})$.

\xhdr{Robust aggregation rules}
Beyond the classical aggregation rules above, for pairwise aggregation we also evaluate four rules from the robust forecast aggregation literature (see \Cref{sec:fixed-rules} for details).

\section{Experimental Setups and Dataset Statistics}
\label{sec:model-merging}
\label{sec:groups}
\label{sec:split-pairs}
\label{sec:exp}
\label{sec:experimental-setup}

\colorlet{resultshade}{forecastpurple!8}
\newcommand{\resulttablesetup}{%
    \centering
    \fontsize{8.0}{9.5}\selectfont
    \setlength{\tabcolsep}{3.2pt}%
    \renewcommand{\arraystretch}{1.05}%
}

\newsavebox{\resultTableBox}
\newcommand{\fitresulttable}[2][\linewidth]{%
  \sbox{\resultTableBox}{#2}%
  \ifdim\wd\resultTableBox>#1\relax
    \resizebox{#1}{!}{\usebox{\resultTableBox}}%
  \else
    \usebox{\resultTableBox}%
  \fi
}

In this section, we describe our experiment setups and report some dataset statistics that are relevant to this work.
We construct our experimental comparisons in two stages. We first
select representative forecasters from the available model
configurations, then organize them into groups with sufficient
shared forecast coverage. 
Within each group, we compare individual and aggregated forecasts on the same subquestions. 
We describe this construction below, followed by the training procedure and evaluation criteria.

\subsection{Data and Comparison Groups}
We use probability forecasts and resolved outcomes from
ForecastBench~\citep{KBYJ-25}. The data catalog contains $8{,}556$
questions and $22{,}162$ binary subquestions, described in \Cref{sec:prediction-task}. 
The raw forecasts come from $313$ model configurations covering 30 forecast rounds from 2024-07-21 to 2026-06-07. The corresponding questions are resolved between 2024-07-25 and 2026-09-11.
To limit the representation of closely related model configurations, we organize them into \emph{model families}. Configurations in
the same family share a model series and capability variant,
but may differ in release date, prompt, or information
condition. 
For example, GPT-5 configurations with different
prompts belong to the same family, whereas GPT-5 and GPT-5-Mini
belong to different families.
Within each family, we select the configuration with forecasts
for the largest number of distinct subquestions as its
\emph{representative}. 
This gives $91$ family representatives, whose set we denote by $\modelUniverse$. We use these representatives
to construct the comparison groups.

\xhdr{Comparison groups and shared coverage}
Forecast coverage differs across the selected forecasters.
To compare them on a common sample, we form groups whose members
have forecasted a sufficiently large set of shared subquestions.
In particular, we index forecasters by $\forecasterIdx\in\modelUniverse$.
For each forecaster~$\forecasterIdx$, we define its coverage set
$\coveredEvents{\forecasterIdx}$ as the question-subquestion pairs
$(\questionIdx,\subquestionIdx)$ for which it provides a probability forecast.

We form a \emph{comparison group} by selecting a subset of
forecasters $\modelSubset\subseteq\modelUniverse$. To compare these forecasters
on the same subquestions, we use only those forecasted by every
member. We denote this shared coverage by
$\coveredEvents{\modelSubset}
=\bigcap_{\forecasterIdx\in \modelSubset}\coveredEvents{\forecasterIdx}$.
To guarantee adequate per-group data for our evaluation, we require each group to contain at least three forecasters and at least $\groupThreshold$ shared subquestions, where $\groupThreshold$ is the coverage threshold. A group is therefore
\emph{eligible} if
$\setsize{\modelSubset}\geq 3,
\setsize{\coveredEvents{\modelSubset}}\geq\groupThreshold$.
With at least three forecasters, we can combine the forecasts
of two members and compare the result with a third forecaster
who is not part of the aggregate.

An eligible group may contain many smaller eligible groups.
To reduce the number of comparisons, we retain only
\emph{maximal groups}: eligible groups to which no additional
forecaster can be added while maintaining at least
$\groupThreshold$ shared subquestions.
Formally, an eligible group~$\modelSubset$ is maximal if
\[
    \setsize{
        \coveredEvents{\modelSubset}\cap\coveredEvents{\forecasterIdx}
    }
    <\groupThreshold
    \quad
    \text{for every }
    \forecasterIdx\in\modelUniverse\setminus \modelSubset.
\]
We keep all groups satisfying this condition.
These groups may differ in size and share some forecasters.
We evaluate them separately because merging distinct maximal
groups would reduce their shared coverage below
$\groupThreshold$.
For each group, we use the full set of shared subquestions
to construct the training and test sets;
$\groupThreshold$ specifies a minimum rather than a fixed
sample size.

\begin{table}[!htbp]
    \centering
    \fontsize{8.0}{9.5}\selectfont
    \setlength{\tabcolsep}{5pt}
    \renewcommand{\arraystretch}{1.08}
    \caption{ Overall dataset statistics by coverage threshold
    $\groupThreshold$. Each forecaster is counted once per threshold.
    The last two columns report the ranges of group sizes and shared-subquestion counts before the training-test split.
    }
    \label{tab:setup-data-statistics}
    \begin{tabular}{rrrrr}
        \toprule
        \shortstack[r]{Required shared\\subquestions\\($\groupThreshold$)}
        & \shortstack[r]{Distinct\\forecasters}
        & \shortstack[r]{No.\ of maximal\\groups}
        & \shortstack[r]{Forecasters per group\\(min-max)}
        & \shortstack[r]{Shared subquestions\\per group (min-max)} \\
        \midrule
        $1{,}000$ & $70$ & $16$ & $7$--$18$ & $1{,}006$--$2{,}969$ \\
        $4{,}000$ & $22$ & $9$  & $3$--$9$  & $4{,}082$--$5{,}907$ \\
        \bottomrule
\end{tabular}
\end{table}

\xhdr{Construction statistics}
We apply this procedure at coverage thresholds
$\groupThreshold\in\{1{,}000,4{,}000\}$.
At $\groupThreshold=1{,}000$, retaining only maximal groups
reduces the number of comparison groups from $567{,}316$ to $16$.
Across both thresholds, $70$ of the $91$ family representatives
appear in at least one retained group and constitute the
forecasters evaluated in our main experiments.
\Cref{tab:setup-data-statistics} summarizes the number of
distinct forecasters, the number of retained groups, and
the ranges of group sizes and shared coverage at each threshold.
Increasing the coverage threshold from $1{,}000$ to $4{,}000$
reduces the number of distinct forecasters from $70$ to $22$.
Within each threshold, we order groups by decreasing number
of forecasters, breaking ties by decreasing number of shared subquestions.
Detailed group-level statistics are provided in \Cref{tab:overall-group-membership} in \Cref{apx:exp-setup}.
\subsection{Experimental Protocol and Evaluation}\label{sec:protocol-and-evaluation}
\vspace{-5pt}
We evaluate whether individually weaker forecasters can be aggregated to outperform the strongest individual in their comparison group. We assess accuracy and calibration using the Brier score (BS) and expected calibration error (ECE) defined in \Cref{sec:evaluation}.
In our evaluation, we divide each group's shared subquestions into approximately equal training and test sets using a fixed random seed. 
We use the same partition for all forecasters, aggregation rules, and pool sizes within the group. Our category-specific experiments construct groups separately within each category using the same coverage criterion, but split the data by question, keeping each question's subquestions together. 
We describe additional restrictions alongside the corresponding experiments.


\xhdr{Group benchmark and candidate pools}
Within each group, we identify the \emph{strongest forecaster}
as the individual with the lowest test Brier score. We use this
forecaster as the \emph{group benchmark} and exclude it from every candidate aggregate. 
For a pool size~$\poolSize\in\{2,3,4,5\}$, we define a candidate pool~$\candidatePool \subseteq \modelSubset$ as a set of $\poolSize$ distinct forecasters, each with a strictly higher test Brier score than the benchmark.
We evaluate all such pools in every group with enough eligible
members.
Our comparison is retrospective because we use test BS to select both the benchmark and eligible forecasters: we ask whether individuals who perform worse on the evaluation sample can collectively outperform its strongest individual. 
\wtedit{For the ECE analysis, we keep the same benchmark and candidate pools.}

We compare the fixed and learned rules defined in \Cref{sec:aggregation-rules}. 
Our baselines for pools of three to five forecasters are the arithmetic mean, harmonic mean, geometric mean of odds, and median. 
Our pairwise baselines include the arithmetic mean, harmonic mean, geometric mean of odds, and four robust rules: average-prior, heuristic-prior, precision, and prior-agnostic. 
We evaluate learned log-odds and linear pooling at every pool size. We fit each learned rule separately for each pool by minimizing training BS over nonnegative coefficients, then hold these coefficients fixed for both test BS and test ECE. The fixed rules require no training.


\xhdr{Evaluation and reporting}
We evaluate every aggregate on its group's common test set. For each metric, we compare the aggregate with the group benchmark using two criteria. An aggregate \emph{matches or improves} the benchmark if its score is no greater than the benchmark’s, and is \emph{within 5\%} if its score is at most 1.05 times the benchmark's. The latter criterion includes all matching or improving aggregates.
We apply these criteria separately to BS and ECE and report two complementary summaries for each rule and pool size:
\begin{itemize}[
    leftmargin=1.2em,
    labelsep=0.4em,
    itemsep=1pt,
    topsep=2pt,
    parsep=0pt,
    partopsep=0pt
]
    \item
    Group-level rates report the fraction of eligible groups containing at least one pool that matches or improves the benchmark, or lies within 5\% of it. We include only groups with at least one eligible pool in the denominator.
    \item 
    Pool-level rates report the corresponding fractions across all eligible pool occurrences. We count each group–pool combination once, so the same model combination appearing in multiple groups is evaluated separately on each group's test set.
\end{itemize}
We additionally report whether aggregation improves on the forecasters being combined. Specifically, we report the fraction of eligible pools whose aggregate score is no greater than that of every member of the pool. For two-forecaster pools, this is reported as \emph{Improve on both members}. We apply these criteria separately to BS and ECE.

For the category-specific analyses in \Cref{sec:event-type},
we construct comparison groups within each category using
the same coverage criterion. We state any additional restrictions
alongside the corresponding experiments.

\section{Main Experiment Results}
\label{sec:main-results}

\wtedit{In this section, we focus on pairwise aggregation and compare learned and fixed aggregation rules.
We set the coverage threshold to be \(\groupThreshold=1{,}000\) which gives us 1,121 eligible pairs from 16 comparison groups.}
\rxedit{We then evaluate the additional benefit of multi-model aggregation.
We additionally examine whether successful pairs remain available under practical input-price and open-weight constraints in \Cref{apx:prac-constraint-materials}.}
\wtedit{Our results show that aggregating just two individually weaker forecasters can outperform the group benchmark, where the learned linear pooling achieve\aiedit{s} the highest overall success rates in both BS and ECE.}



\subsection{Pairwise Aggregation Results}
\label{sec:BS-results}

We first examine whether aggregating two individually weaker forecasters improves forecast accuracy and can close the gap to the group benchmark in their comparison group.
We then compare learned and fixed rules to assess the additional benefit of learning the weights. 
We also examine whether these gains persist when neither constituent is close to the benchmark, \aiedit{and} whether the BS improvements are accompanied by good calibration.

\xhdr{Aggregating two weaker models can outperform the strongest group benchmark}
\wtedit{In \Cref{tab:overall-main-results-expanded}, we report the overall two-model BS results across the 16 comparison groups and 1,121 eligible pairs. 
For each aggregation rule, we report the fraction of groups and pairs that match or improve on the group benchmark, the corresponding fractions within 5\% of the benchmark, and the fraction of pairs whose aggregate improves on both constituent forecasters.}\footnote{\rxedit{In \Cref{tab:group-best-individual-top2-pairs-brier} of \Cref{apx:pairwise-agg-materials}, we record \aiedit{the} detailed group benchmark and the best weak-pair aggregate under each learned rule for each group, for which we note that the aggregation weights here are all non-zeros, so the aggregation rules are actually utilizing all the forecasters.}}

\begin{table}[h]
\caption{Two-model BS results at $\groupThreshold=1{,}000$.
Cells report
\rxedit{ the group-level and pair-level rates as well as the ``Improve on both members'' described in \Cref{sec:protocol-and-evaluation} in the form of fractions and percentages.
}
}
\label{tab:overall-main-results-expanded}
\resizebox{0.95\linewidth}{!}{
\begin{tabular}{lccccc}
\toprule
& \multicolumn{2}{c}{Groups}
& \multicolumn{3}{c}{Eligible pairs} \\
\cmidrule(lr){2-3}\cmidrule(lr){4-6}
Aggregation rule & \shortstack{Match or improve\\benchmark}
& \shortstack{Within $5\%$\\of benchmark}
& \shortstack{Match or improve\\benchmark}
& \shortstack{Within $5\%$\\of benchmark}
& \shortstack{Improve on\\both members} \\
\midrule
\rowcolor{resultshade}
Learned linear pooling & \shortstack{11/16\\(68.75\%)} & \shortstack{16/16\\(100.00\%)} & \shortstack{315/1,121\\(28.10\%)} & \shortstack{730/1,121\\(65.12\%)} & \shortstack{1,073/1,121\\(95.72\%)} \\
\rowcolor{resultshade}
Learned log-odds pooling & \shortstack{8/16\\(50.00\%)} & \shortstack{15/16\\(93.75\%)} & \shortstack{84/1,121\\(7.49\%)} & \shortstack{467/1,121\\(41.66\%)} & \shortstack{1,022/1,121\\(91.17\%)} \\
\midrule
Simple mean (AM) & \shortstack{8/16\\(50.00\%)} & \shortstack{15/16\\(93.75\%)} & \shortstack{70/1,121\\(6.24\%)} & \shortstack{345/1,121\\(30.78\%)} & \shortstack{736/1,121\\(65.66\%)} \\
Harmonic mean (HM) & \shortstack{8/16\\(50.00\%)} & \shortstack{16/16\\(100.00\%)} & \shortstack{46/1,121\\(4.10\%)} & \shortstack{349/1,121\\(31.13\%)} & \shortstack{658/1,121\\(58.70\%)} \\
Log-odds mean (GM) & \shortstack{8/16\\(50.00\%)} & \shortstack{15/16\\(93.75\%)} & \shortstack{52/1,121\\(4.64\%)} & \shortstack{375/1,121\\(33.45\%)} & \shortstack{689/1,121\\(61.46\%)} \\
\bottomrule
\end{tabular}
}
\end{table}

\wtedit{We find that the learned rules produce successful pairs across many comparison groups. Learned linear pooling identifies at least one pair that matches or improves on the group benchmark in 11 of the 16 groups (68.75\%), and at least one pair within 5\% of the benchmark in all 16 groups.
Notably, even under a tighter 3\% threshold, the best learned-linear pair is within 3\% of the group benchmark in all 16 comparison groups (see \Cref{tab:group-best-individual-top2-pairs-brier} for detailed BSs.)
Learned log-odds pooling matches or improves on the benchmark in 8 groups (50.00\%) and comes within 5\% in 15 groups (93.75\%). These results show that pairs of individually weaker forecasters can perform close to, or better than, the group benchmark across many comparison groups.}\footnote{\wtedit{As a robustness check, we increase the shared-coverage threshold from \(\groupThreshold=1{,}000\) to \(\groupThreshold=4{,}000\) subquestions and continue to observe the same qualitative pattern of weak-to-strong aggregation; see \Cref{apx:N4000} for details.}}

\begin{figure}[h]
    \centering
    \resizebox{0.95\linewidth}{!}{
\begingroup
\begin{tikzpicture}
\begin{axis}[
forecast axes front,
width=\linewidth,height=0.45\linewidth,
xmin=0.5,xmax=16.5,ymin=0.775,ymax=0.91,
xtick={1,...,16},
xticklabels={G1,G2,G3,G4,G5,G6,G7,G8,G9,G10,G11,G12,G13,G14,G15,G16},
ytick={0.78,0.80,0.82,0.84,0.86,0.88,0.90},
yticklabel style={/pgf/number format/fixed zerofill,/pgf/number format/precision=2},
ylabel={1 $-$ Test Brier score},
tick label style={font=\fontsize{6}{7}\selectfont},
label style={font=\fontsize{6}{7}\selectfont},
scaled ticks=false,ymajorgrids=true,grid style={black!10},tick align=outside,
legend style={at={(0.5,1.015)},anchor=south,draw=none,font=\fontsize{6}{7}\selectfont,legend columns=3},
bar width=3.6pt,
]

\addplot[
forget plot,ybar,bar shift=0pt,
fill=bsindividual,draw=bsindividual,mark=none,
error bars/.cd,y dir=both,y explicit,error mark=-,
error bar style={gray,line width=0.8pt},
error mark options={mark size=0.20pt,line width=3pt}
] coordinates {
  (0.80,0.885) += (0,0.012) -= (0,0.012)
  (1.80,0.853) += (0,0.009) -= (0,0.009)
  (2.80,0.829) += (0,0.010) -= (0,0.010)
  (3.80,0.880) += (0,0.012) -= (0,0.012)
  (4.80,0.875) += (0,0.011) -= (0,0.011)
  (5.80,0.821) += (0,0.012) -= (0,0.012)
  (6.80,0.853) += (0,0.014) -= (0,0.014)
  (7.80,0.823) += (0,0.011) -= (0,0.011)
  (8.80,0.871) += (0,0.010) -= (0,0.010)
  (9.80,0.865) += (0,0.012) -= (0,0.012)
  (10.80,0.868) += (0,0.010) -= (0,0.010)
  (11.80,0.873) += (0,0.013) -= (0,0.013)
  (12.80,0.820) += (0,0.012) -= (0,0.012)
  (13.80,0.845) += (0,0.013) -= (0,0.013)
  (14.80,0.857) += (0,0.011) -= (0,0.011)
  (15.80,0.835) += (0,0.011) -= (0,0.011)
};
\addlegendimage{area legend,fill=bsindividual,draw=bsindividual}
\addlegendentry{Group benchmark}

\addplot[
forget plot,ybar,bar shift=0pt,
fill=bslinear,draw=bslinear,mark=none,
error bars/.cd,y dir=both,y explicit,error mark=-,
error bar style={gray,line width=0.8pt},
error mark options={mark size=0.20pt,line width=3pt}
] coordinates {
  (1.00,0.882) += (0,0.012) -= (0,0.012)
  (2.00,0.857) += (0,0.008) -= (0,0.008)
  (3.00,0.852) += (0,0.009) -= (0,0.009)
  (4.00,0.882) += (0,0.011) -= (0,0.011)
  (5.00,0.874) += (0,0.011) -= (0,0.011)
  (6.00,0.842) += (0,0.010) -= (0,0.010)
  (7.00,0.854) += (0,0.013) -= (0,0.013)
  (8.00,0.843) += (0,0.011) -= (0,0.011)
  (9.00,0.872) += (0,0.009) -= (0,0.009)
  (10.00,0.865) += (0,0.012) -= (0,0.012)
  (11.00,0.867) += (0,0.009) -= (0,0.009)
  (12.00,0.873) += (0,0.013) -= (0,0.013)
  (13.00,0.848) += (0,0.009) -= (0,0.009)
  (14.00,0.844) += (0,0.013) -= (0,0.013)
  (15.00,0.863) += (0,0.010) -= (0,0.010)
  (16.00,0.845) += (0,0.009) -= (0,0.009)
};
\addlegendimage{area legend,fill=bslinear,draw=bslinear}
\addlegendentry{Learned linear pooling}

\addplot[
forget plot,ybar,bar shift=0pt,
fill=bslogodds,draw=bslogodds,mark=none,
error bars/.cd,y dir=both,y explicit,error mark=-,
error bar style={gray,line width=0.8pt},
error mark options={mark size=0.20pt,line width=3pt}
] coordinates {
  (1.20,0.882) += (0,0.012) -= (0,0.012)
  (2.20,0.861) += (0,0.009) -= (0,0.009)
  (3.20,0.830) += (0,0.010) -= (0,0.010)
  (4.20,0.882) += (0,0.011) -= (0,0.011)
  (5.20,0.874) += (0,0.011) -= (0,0.011)
  (6.20,0.833) += (0,0.012) -= (0,0.012)
  (7.20,0.855) += (0,0.013) -= (0,0.013)
  (8.20,0.836) += (0,0.013) -= (0,0.013)
  (9.20,0.866) += (0,0.011) -= (0,0.011)
  (10.20,0.865) += (0,0.012) -= (0,0.012)
  (11.20,0.869) += (0,0.011) -= (0,0.011)
  (12.20,0.872) += (0,0.012) -= (0,0.012)
  (13.20,0.827) += (0,0.010) -= (0,0.010)
  (14.20,0.843) += (0,0.012) -= (0,0.012)
  (15.20,0.853) += (0,0.010) -= (0,0.010)
  (16.20,0.826) += (0,0.010) -= (0,0.010)
};
\addlegendimage{area legend,fill=bslogodds,draw=bslogodds}
\addlegendentry{Learned log-odds pooling}
\end{axis}

\end{tikzpicture}
\endgroup
    }
    \caption{Best learned pair versus the group benchmark across the 16 comparison groups at \(\groupThreshold=1{,}000\). 
    For each group, we plot test 1 $-$ BS for the group benchmark and the lowest-BS pair under learned linear and log-odds pooling, selected separately for each rule. Higher bars (1 $-$ BS) are better; groups are ordered by decreasing number of forecasters.}
    \label{fig:overall-size-sorted-thresholds}
 \end{figure}
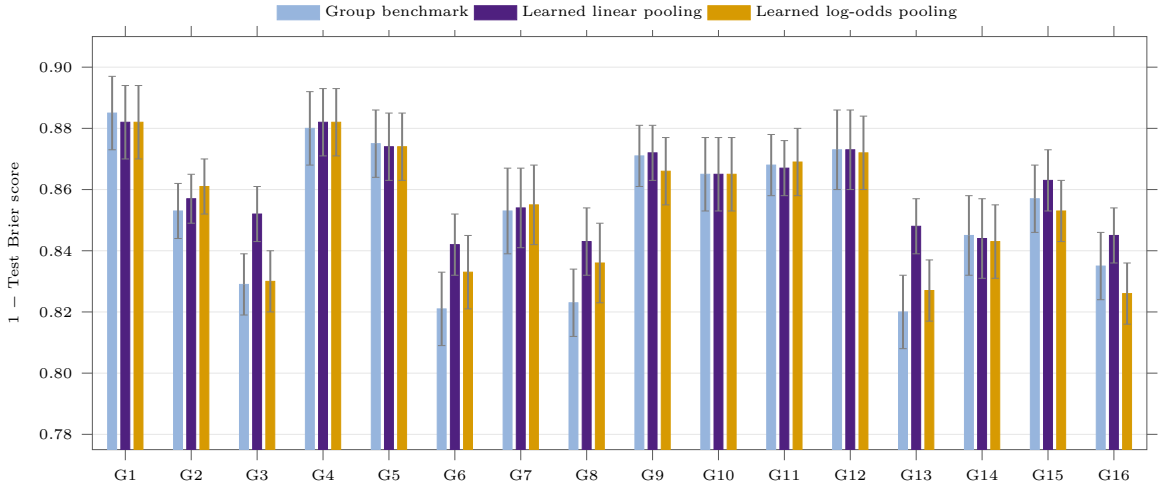

\wtedit{In \Cref{fig:overall-size-sorted-thresholds}, 
we visualize these group-level results. 
For each comparison group, we report the group benchmark together with the lowest-BS pair under each learned rule, where the best pair is selected separately for linear and log-odds pooling. 
A best-pair bar at or above the benchmark bar indicates a successful pair. We thus observe directly from \Cref{fig:overall-size-sorted-thresholds} that learned linear pooling produces a successful pair in 11 groups while learned log-odds pooling in 8 groups and pairwise aggregation is more significant in G3, G6, G8, G13 and G16. The figure also shows that their relative performance may vary across groups: linear pooling succeeds where log-odds pooling does not in G9, G12, G15, and G16, while the reverse occurs in G11. }

\xhdr{Learning the weights matters}
\wtedit{We next examine the additional benefit of learning the aggregation coefficients by comparing each learned rule with its equal-weight counterpart. 
\Cref{tab:overall-main-results-expanded}
shows that learned linear pooling matches or improves on the benchmark for 28.10\% of the eligible pairs, compared with 6.24\% for the simple mean. 
Learned log-odds pooling also improves on its equal-weight counterpart, with success rates of 7.49\% and 4.64\%, respectively. Learning also makes improvement over both constituent forecasters more common: learned linear and log-odds pooling improve on both members for 95.72\% and 91.17\% of the pairs, compared with 65.66\% and 61.46\% for their equal-weight counterparts.}\footnote{\rxedit{Apart from results in \Cref{tab:overall-main-results-expanded}, we have also compared each learned rule with its equal-weight counterpart on the same 1,121 pairs in \Cref{fig:sec4-learned-vs-means} (see \Cref{apx:pairwise-agg-materials}) and examined the robust aggregation rules in the same manner shown in this section (see \Cref{tab:overall-main-results-expanded-continued} of \Cref{apx:pairwise-agg-materials} for more details).}}

\wtedit{\xhdr{The effect is not driven only by near-best models}
We next examine whether aggregation remains effective when neither constituent is close to the benchmark. We call a forecaster {\em near-best} if its individual test BS is at most 5\% above that of the group benchmark, and classify a pair as {\em near} if it contains at least one near-best member and {\em no-near} otherwise.}

\rxedit{In \Cref{tab:no-near-pairs} (see \Cref{apx:pairwise-agg-materials}), we first report group-level results which show that successful aggregation generally does not require a near-best constituent. 
In particular, under learned linear pooling, 16 groups contain a pair within 5\% of the benchmark, and 15 of these groups contain a no-near pair meeting this criterion, compared with 9 containing a near pair.
The pair-level results also share a similar pattern mostly, with \aiedit{a} massive no-near-pair fraction. 
These results further indicate that the gains from learned linear pooling do not rely on having a constituent already close to the benchmark.}

\wtedit{
\xhdr{The BS gains are generally accompanied by good calibration} 
We next evaluate the same aggregates using ECE as a complementary metric. 
We keep the BS-best group benchmark, the BS-defined candidate pairs, and the coefficients learned by minimizing training BS fixed, and evaluate their test ECE without reselecting the benchmark or refitting the learned rules. 
In \Cref{tab:overall-ece-brier-trained} (see \Cref{apx:pairwise-agg-materials}), we report the corresponding ECE results across the same 16 comparison groups and 1,121 eligible pairs.} 
\rxedit{
In \Cref{tab:overall-ece-brier-trained}, learned linear pooling finds at least one pair that matches or improves on the group benchmark's ECE in all 16 comparison groups; 772 of the 1,121 eligible pairs (68.87\%) meet this criterion. Learned log-odds pooling does so in 13 groups and for 291 pairs (25.96\%), with a pair-level success rate similar to that of the log-odds mean (27.92\%).
}

\subsection{Multi-Model Aggregation}
\label{sec:multi-aggregation}

\wtedit{We next examine whether increasing the number of individually weaker forecasters provides substantial additional gains in the aggregated forecasts. 
In \Cref{fig:sec4-three-models} (see \Cref{apx:multi-aggregation}), we summarize and report aggregation performance as the pool size increases from two models to five models. 
\Cref{fig:sec4-three-models-a} reports the fraction of comparison groups containing at least one pool that matches or improves on the group benchmark. 
In particular,
for learned linear pooling, this fraction increases from 68.75\% (11/16 groups) with two forecasters to 75.00\% (12/16) with three, and is 68.75\% (11/16) with four and five. 
For learned log-odds pooling, the corresponding fraction increases from 50.00\% (8/16) with two forecasters to 68.75\% (11/16) with three, and remains at 68.75\% with four and five. 
Thus, increasing the pool size can create additional successful groups, but most of the increase occurs when moving from two to three forecasters and does not continue as the pool grows further.}

\rxedit{\Cref{fig:sec4-three-models-b} zooms in group G13. 
The best learned-linear pool reaches $0.848$ with two members and remains at approximately $0.848$ for exact pool sizes three through five. 
Thus, larger candidate pools do not materially raise the best observed score in this group.}

\subsection{Question-Category-Based Analysis}
\label{sec:event-type}
\wtedit{We next study whether the aggregation gains differ across question categories. 
We form comparison groups separately for Politics and Finance at the same coverage threshold \(\groupThreshold=1{,}000\), yielding 103 eligible pairs in 12 Politics groups and 287 pairs in 9 Finance groups. We evaluate the same learned rules and retain each category-specific group's BS-best individual as the benchmark.}
\rxedit{We also examine other question categories including Climate, Science, Sports and Entertainment. However, these categories do not include enough comparison groups for further investigation. 
Nevertheless, we provide the dataset statistics of $6$ question categories in \Cref{tab:event-type-data-statistics} of \Cref{apx:event-type-materials}.}

We first observe that the aggregation gains can differ substantially across question categories from \Cref{tab:event-type-threshold-1000} (see \Cref{apx:event-type-materials}). Results from \Cref{tab:event-type-threshold-1000} show that learned linear pooling is particularly effective in the Finance comparisons, both in how widely the gains appear across groups and how frequently they appear across eligible pairs \rxedit{(In \Cref{apx:event-type-materials}, we provide detailed group-level statistics in \Cref{tab:event-type-group-membership-n1000} while \aiedit{the} group benchmark and the best weak-pair aggregate under each learned rule for each group are demonstrated in \Cref{tab:group-best-individual-top2-pairs-brier-politics-and-finance})}.

\Cref{fig:event-type-size-sorted-groups} \rxedit{(see \Cref{apx:event-type-materials})} provides a group-level view of these differences. 
One possible explanation \aiedit{for why} the learned linear pooling performs so well for \aiedit{the} Finance category is the following. 
The Finance benchmarks have substantially higher BS than the Politics benchmarks: across the 9 Finance groups, benchmark BS ranges from 0.210 to 0.247, whereas across the 12 Politics groups it ranges from 0.075 to 0.111. 
The strongest group benchmark therefore performs worse in the Finance category compared to those in \aiedit{the} Politics category, which leaves more room for aggregation to close the gap. 
\rxedit{We also observe} a similar pattern for calibration, as the BS-selected benchmarks generally have higher ECE in Finance than in Politics \rxedit{(see \Cref{fig:event-type-ece-groups}} in \Cref{apx:event-type-materials}).
This may partly explain why learned linear pooling performs especially well in Finance, including its 97.91\% pair-level ECE success rate.\footnote{Because the category-specific groups differ in both forecasters and shared questions, however, we view this as a possible explanation rather than a causal effect of question category.   
}

\section{Conclusions}
\label{sec:conclusion}

In this paper, we study weak-to-strong forecast aggregation for LLMs: whether individually weaker forecasters can be aggregated to match or outperform a stronger forecaster. Using ForecastBench, we find substantial evidence of such improvement. 
Learned linear pooling identifies a weaker pair that matches or outperforms the strongest individual in 11 of 16 comparison groups and comes within 5\% of its Brier score in all 16 groups. 
These gains do not rely on having a near-best constituent and are generally accompanied by good calibration.  
Our additional analyses show that learning the aggregation weights matters, while adding more forecasters does not consistently improve performance. 
The weak-to-strong pattern also persists under a higher shared-coverage threshold and remains available under practical cost and access constraints. 

\section*{AI Disclosure}
\label{sec:AI-disclosure}
In this work, we have not used generative AI tools for 
designing or providing feedback on research methodology or experiments,
implementing methods,
assisting with translation,
cleaning and reformatting \aiedit{the} dataset,
supporting qualitative and thematic data analysis,  interpreting results.
The AI is used for proofreading and improving the readability.
We take responsibility for the final content of this work,
including text, claims or artifacts produced with the aid of generative AI.

\clearpage
\bibliographystyle{ACM-Reference-Format}
\bibliography{mybib}

\clearpage
\appendix
\captionsetup[table]{font=tablesix}

\section{Additional Aggregation Rules}
\label{sec:fixed-rules}

We also draw on the robust forecast aggregation
literature~\citep{ABS-18,KWW-24,GHHKSY-25,CPT-26}, which studies
how to combine forecasts when the forecasters' underlying
information structures are unknown to the aggregator.
We consider four fixed rules from this literature as baselines:
two prior plug-in rules, a precision rule, and a prior-agnostic robust rule.

\xhdr{Prior plug-in aggregators}
Let $\priormean\in[0,1]$ denote the common prior probability of a Yes outcome. 
Following \citet{ABS-18}, given two forecasts $\forecastProb^{1},\forecastProb^{2}$, we define
\[
    \bayesRule(\priormean;\forecastProb^{1},\forecastProb^{2})
    =
    \frac{
        \forecastProb^{1}\forecastProb^{2}(1-\priormean)
    }{
        \forecastProb^{1}\forecastProb^{2}(1-\priormean)
        +(1-\forecastProb^{1})(1-\forecastProb^{2})\priormean
    }.
\]
The \emph{average-prior} (AP) rule uses
$\widehat{\priormean}_{\mathrm{AP}}
=(\forecastProb^{1}+\forecastProb^{2})/2$, while the
\emph{heuristic-prior} (HP) rule uses
\[
    \widehat{\priormean}_{\mathrm{HP}}
    =
    0.49(\forecastProb^{1}+\forecastProb^{2})
    +0.02\,\mathbbm{1}\{\forecastProb^{1}+\forecastProb^{2}>1\}.
\]
The corresponding aggregation rules are
$\aggRule_{\mathrm{AP}}(\forecastProb^{1},\forecastProb^{2})
=\bayesRule(\widehat{\priormean}_{\mathrm{AP}};
\forecastProb^{1},\forecastProb^{2})$
and
$\aggRule_{\mathrm{HP}}(\forecastProb^{1},\forecastProb^{2})
=\bayesRule(\widehat{\priormean}_{\mathrm{HP}};
\forecastProb^{1},\forecastProb^{2})$.

\xhdr{Precision aggregator}
We also use the precision aggregator of \citet{ABS-18}, which gives
more weight to forecasts closer to $0$ or $1$. For
$\forecastProb\in(0,1)$, define
$\precision(\forecastProb)=1/[\forecastProb(1-\forecastProb)]$.
The rule is
\[
    \aggRule_{\mathrm{pre}}(\forecastProb^{1},\forecastProb^{2})
    =
    \begin{cases}
        \displaystyle
        \frac{
            \precision(\forecastProb^{1})\forecastProb^{1}
            +\precision(\forecastProb^{2})\forecastProb^{2}
        }{
            \precision(\forecastProb^{1})
            +\precision(\forecastProb^{2})
        },
        &|\forecastProb^{1}-\forecastProb^{2}|\leq 0.4,\\[1.2em]
        \displaystyle
        \frac{
            \sqrt{\precision(\forecastProb^{1})}\,\forecastProb^{1}
            +\sqrt{\precision(\forecastProb^{2})}\,\forecastProb^{2}
        }{
            \sqrt{\precision(\forecastProb^{1})}
            +\sqrt{\precision(\forecastProb^{2})}
        },
        &|\forecastProb^{1}-\forecastProb^{2}|>0.4.
    \end{cases}
\]

\xhdr{Prior-agnostic robust aggregator}
Finally, we use the prior-agnostic robust rule of \citet{CPT-26},
which combines the forecasts directly in log-odds space:
$\aggRule_{\robustCoeff}(\forecastProb^{1},\forecastProb^{2})
=
\frac{1}{
    1+
    \left(
        \frac{1-\forecastProb^{1}}{\forecastProb^{1}}
    \right)^{\robustCoeff}
    \left(
        \frac{1-\forecastProb^{2}}{\forecastProb^{2}}
    \right)^{\robustCoeff}
}$.
We refer to this rule as $\robustCoeff$-log-odds.
In our experiments, we set $\robustCoeff=0.56$, following \citet{CPT-26}.

\section{Additional Experimental Details and Group Statistics}
\label{apx:exp-setup}
In the following~\Cref{tab:overall-group-membership}, we report the size, subquestion number, and model configuration components of the groups we examined in our main experiment.

\begingroup
\setlength{\tabcolsep}{2.0pt}
\renewcommand{\arraystretch}{1.08}
\fontsize{6.0}{7.0}\selectfont
\begin{longtable}{@{}l l l@{\hspace{10pt}} >{\raggedright\arraybackslash}p{\dimexpr\textwidth-8\tabcolsep-33mm\relax}@{}}
\caption{Complete membership of the size-sorted overall groups.}
\label{tab:overall-group-membership}

\\
\toprule
Index & Size & Subquestions & Models \\
\midrule
\endfirsthead

\toprule
Index & Size & Subquestions & Models \\
\midrule
\endhead

\midrule
\multicolumn{4}{r}{\fontsize{6.0}{7.0}\selectfont Continued on next page}\\
\endfoot

\bottomrule
\endlastfoot

\multicolumn{4}{@{}l}{\textbf{Overall, threshold 1,000}}\\
\addlinespace[1pt]
1 & 18 & 1,065 &
Claude-3-7-Sonnet;
Claude-Haiku-4-5;
Claude-Opus-4-1;
Claude-Sonnet-4-5;
DeepSeek-V3.1;
GPT-4.1;
GPT-5-Mini;
GPT-5-Nano;
GPT-5.1;
Gemini-2.5-Pro;
Gemini-3-Pro-Preview;
Grok-4;
Grok-4-1-Fast-Non-Reasoning;
Grok-4-1-Fast-Reasoning;
Grok-4-Fast-Non-Reasoning;
Grok-4-Fast-Reasoning;
Qwen3-235B-A22B-Fp8-Tput;
Qwen3-235B-A22B-Thinking
\\

2 & 16 & 2,969 &
Claude-3-5-Sonnet;
Claude-3-7-Sonnet;
Claude-3-Haiku;
Claude-3-Opus;
GPT-4-Turbo;
GPT-4.1;
GPT-4o;
Llama-3.2-3B-Instruct-Turbo;
Llama-3.3-70B-Instruct-Turbo;
Llama-4-Maverick-17B-128E-Instruct-FP8;
Llama-4-Scout-17B-16E-Instruct;
Meta-Llama-3.1-405B-Instruct-Turbo;
Mistral-Large;
O4-Mini;
Qwen2.5-72B-Instruct-Turbo;
Qwen3-235B-A22B-Fp8-Tput
\\

3 & 16 & 1,902 &
Claude-2.1;
Claude-3-5-Sonnet;
Claude-3-Haiku;
Claude-3-Opus;
GPT-3.5-Turbo;
GPT-4;
GPT-4-Turbo;
GPT-4o;
Gemini-1.5-Flash;
Gemini-1.5-Pro;
Llama-2-70b-Chat-Hf;
Llama-3-70b-Chat-Hf;
Llama-3-8b-Chat-Hf;
Mixtral-8x22B-Instruct-V0.1;
Mixtral-8x7B-Instruct-V0.1;
Qwen1.5-110B-Chat
\\

4 & 16 & 1,257 &
Claude-3-7-Sonnet;
Claude-Haiku-4-5;
Claude-Opus-4-1;
Claude-Sonnet-4-5;
DeepSeek-V3.1;
GPT-4.1;
GPT-5-Mini;
GPT-5-Nano;
Gemini-2.5-Flash;
Gemini-2.5-Pro;
Grok-4;
Grok-4-Fast-Non-Reasoning;
Grok-4-Fast-Reasoning;
O3;
Qwen3-235B-A22B-Fp8-Tput;
Qwen3-235B-A22B-Thinking
\\

5 & 16 & 1,137 &
Claude-Haiku-4-5;
Claude-Opus-4-1;
Claude-Sonnet-4-5;
DeepSeek-V3.1;
GPT-4.1;
GPT-5-Mini;
GPT-5-Nano;
GPT-5.1;
Gemini-2.5-Pro;
Gemini-3-Pro-Preview;
Grok-4-1-Fast-Non-Reasoning;
Grok-4-1-Fast-Reasoning;
Grok-4-Fast-Non-Reasoning;
Grok-4-Fast-Reasoning;
Kimi-K2-Thinking;
Qwen3-235B-A22B-Thinking
\\

6 & 15 & 1,512 &
Claude-3-5-Sonnet;
Claude-3-7-Sonnet;
DeepSeek-R1;
DeepSeek-V3;
GPT-4.1;
GPT-4o;
Gemini-2.5-Flash-Preview;
Gemini-2.5-Pro-Preview;
Llama-3.3-70B-Instruct-Turbo;
Llama-4-Maverick-17B-128E-Instruct-FP8;
Llama-4-Scout-17B-16E-Instruct;
Mistral-Large;
O3;
O3-Mini;
QwQ-32B-Preview
\\

7 & 15 & 1,006 &
Claude-Haiku-4-5;
Claude-Opus-4-6;
Claude-Sonnet-4-5;
Claude-Sonnet-4-6;
DeepSeek-V3.1;
GPT-4.1;
GPT-5-Mini;
GPT-5-Nano;
GPT-5.1;
GPT-5.2;
Gemini-2.5-Pro;
Gemini-3-Flash-Preview;
Gemini-3.1-Pro-Preview;
Grok-4-1-Fast-Non-Reasoning;
Grok-4-1-Fast-Reasoning
\\

8 & 14 & 1,397 &
Claude-3-5-Sonnet;
Claude-3-7-Sonnet;
DeepSeek-R1;
DeepSeek-V3;
GPT-4o;
Gemini-2.0-Flash-Lite-001;
Gemini-2.5-Pro-Exp;
Grok-beta;
Llama-3.3-70B-Instruct-Turbo;
Llama-4-Maverick-17B-128E-Instruct-FP8;
Llama-4-Scout-17B-16E-Instruct;
Mistral-Large;
O3-Mini;
QwQ-32B-Preview
\\

9 & 13 & 1,744 &
Claude-3-5-Sonnet;
Claude-Opus-4-1;
Claude-Sonnet-4;
GLM-4.5-Air-FP8;
GPT-4.1;
GPT-5;
GPT-5-Mini;
GPT-5-Nano;
Gemini-2.5-Flash;
Gemini-2.5-Pro;
Kimi-K2-Instruct;
Magistral-Medium;
Qwen3-235B-A22B-Fp8-Tput
\\

10 & 13 & 1,124 &
Claude-Haiku-4-5;
Claude-Opus-4-1;
Claude-Sonnet-4-5;
DeepSeek-V3.1;
GPT-4.1;
GPT-5-Mini;
GPT-5-Nano;
GPT-5.1;
GPT-5.2;
Gemini-2.5-Pro;
Gemini-3-Flash-Preview;
Grok-4-1-Fast-Non-Reasoning;
Grok-4-1-Fast-Reasoning
\\

11 & 11 & 1,426 &
Claude-3-5-Sonnet;
Claude-Opus-4;
Claude-Sonnet-4;
GPT-4.1;
Kimi-K2-Instruct;
Llama-4-Maverick-17B-128E-Instruct-FP8;
Llama-4-Scout-17B-16E-Instruct;
Magistral-Medium;
O4-Mini;
Qwen3-235B-A22B-Fp8-Tput;
claude-3-5-haiku
\\

12 & 10 & 1,060 &
Claude-Haiku-4-5;
Claude-Sonnet-4;
DeepSeek-V3.1;
GPT-4.1;
GPT-5-Mini;
GPT-5-Nano;
Gemini-2.5-Pro;
Grok-4-Fast-Non-Reasoning;
Grok-4-Fast-Reasoning;
Qwen3-235B-A22B-Thinking
\\

13 & 9 & 1,557 &
Claude-3-5-Sonnet;
Claude-3-Haiku;
Claude-3-Opus;
GPT-4-Turbo;
GPT-4o;
Llama-3-70b-Chat-Hf;
Llama-3-8b-Chat-Hf;
Mistral-Large;
Qwen3-235B-A22B-Fp8-Tput
\\

14 & 9 & 1,069 &
Claude-Haiku-4-5;
Claude-Sonnet-4-5;
Claude-Sonnet-4-6;
GPT-5-Mini;
GPT-5-Nano;
GPT-5.4;
GPT-5.4-Mini;
GPT-5.4-Nano;
Gemini-3.1-Pro-Preview
\\

15 & 8 & 1,521 &
Claude-3-5-Sonnet;
Claude-Opus-4-1;
Claude-Sonnet-4;
GPT-4.1;
Kimi-K2-Instruct;
Magistral-Medium;
O4-Mini;
Qwen3-235B-A22B-Fp8-Tput
\\

16 & 7 & 1,860 &
Claude-3-5-Sonnet;
Claude-3-7-Sonnet;
DeepSeek-R1;
DeepSeek-V3;
GPT-4.5-Preview;
Llama-3.3-70B-Instruct-Turbo;
O3-Mini
\\

\midrule
\addlinespace[3pt]
\multicolumn{4}{@{}l}{\textbf{Overall, threshold 4,000}}\\
\addlinespace[1pt]

1 & 9 & 5,112 &
Claude-3-5-Sonnet;
Claude-3-7-Sonnet;
DeepSeek-R1;
DeepSeek-V3;
GPT-4o;
Llama-3.3-70B-Instruct-Turbo;
Mistral-Large;
O3-Mini;
QwQ-32B-Preview
\\

2 & 8 & 4,866 &
Claude-3-5-Sonnet;
Claude-3-7-Sonnet;
GPT-4.1;
GPT-4o;
Llama-3.3-70B-Instruct-Turbo;
Llama-4-Maverick-17B-128E-Instruct-FP8;
Llama-4-Scout-17B-16E-Instruct;
Mistral-Large
\\

3 & 7 & 5,907 &
Claude-3-5-Sonnet;
Claude-3-Haiku;
Claude-3-Opus;
GPT-4-Turbo;
GPT-4o;
Mistral-Large;
Qwen3-235B-A22B-Fp8-Tput
\\

4 & 6 & 5,342 &
Claude-3-5-Sonnet;
Claude-3-Haiku;
Claude-3-Opus;
GPT-4-Turbo;
GPT-4o;
Llama-3-70b-Chat-Hf
\\

5 & 6 & 4,634 &
Claude-3-5-Sonnet;
Claude-Sonnet-4;
GPT-4.1;
Kimi-K2-Instruct;
Magistral-Medium;
Qwen3-235B-A22B-Fp8-Tput
\\

6 & 6 & 4,409 &
Claude-3-5-Sonnet;
GPT-4.1;
Llama-4-Maverick-17B-128E-Instruct-FP8;
Llama-4-Scout-17B-16E-Instruct;
O4-Mini;
Qwen3-235B-A22B-Fp8-Tput
\\

7 & 3 & 4,289 &
Claude-Opus-4-1;
GPT-4.1;
Qwen3-235B-A22B-Fp8-Tput
\\

8 & 3 & 4,203 &
Claude-3-7-Sonnet;
GPT-4.1;
Qwen3-235B-A22B-Fp8-Tput
\\

9 & 3 & 4,082 &
Claude-Opus-4-1;
Claude-Sonnet-4;
GPT-4.1
\\

\end{longtable}
\endgroup
Of the 91 model-family representatives, 70 appear in at least one final Overall group. We list these evaluated forecasters in~\Cref{tab:all-evaluated-models}, together with a reference for each model and an indicator of whether its underlying weights are publicly downloadable. We retain the experiment identifiers so that readers can match the models to the group memberships reported above.

\begingroup
\setlength{\tabcolsep}{2.0pt}
\renewcommand{\arraystretch}{1.06}
\fontsize{6.0}{7.0}\selectfont
\newcommand{\modelcite}[1]{\citeauthor{#1}~(\citeyear{#1})}
\begin{longtable}{@{}p{35mm}@{\hspace{4pt}}p{22mm}@{\hspace{4pt}}c@{\hspace{16pt}}p{32mm}@{\hspace{4pt}}p{31mm}@{}}
\caption{The 70 evaluated forecasters in the size-sorted Overall comparison groups at $\groupThreshold=1{,}000$ and $\groupThreshold=4{,}000$. Models are grouped by provider and ordered by public release date within each provider. ``Open weight'' refers to publicly downloadable weights of the underlying model.}\label{tab:all-evaluated-models}
\\
\toprule
LLM & Reference & \shortstack{Open\\weight?} & \shortstack[l]{Comparison groups\\($\groupThreshold=1{,}000$)} & \shortstack[l]{Comparison groups\\($\groupThreshold=4{,}000$)} \\
\midrule
\endfirsthead
\toprule
LLM & Reference & \shortstack{Open\\weight?} & \shortstack[l]{Comparison groups\\($\groupThreshold=1{,}000$)} & \shortstack[l]{Comparison groups\\($\groupThreshold=4{,}000$)} \\
\midrule
\endhead
\midrule
\multicolumn{5}{r}{\fontsize{6.0}{7.0}\selectfont Continued on next page}\\
\endfoot
\bottomrule
\endlastfoot
GPT-3.5-Turbo & \modelcite{modelGPT35Turbo2023} & No & G3 & N/A \\
GPT-4 & \modelcite{modelGPT42023} & No & G3 & N/A \\
GPT-4-Turbo & \modelcite{modelGPT4Turbo2023} & No & G2, G3, G13 & G3, G4 \\
GPT-4o & \modelcite{modelGPT4o2024} & No & G2, G3, G6, G8, G13 & G1, G2, G3, G4 \\
O3-Mini & \modelcite{modelO3Mini2025} & No & G6, G8, G16 & G1 \\
GPT-4.5-Preview & \modelcite{modelGPT45Preview2025} & No & G16 & N/A \\
GPT-4.1 & \modelcite{modelGPT412025} & No & G1, G2, G4, G5, G6, G7,\newline G9, G10, G11, G12, G15 & G2, G5, G6, G7, G8, G9 \\
O3 & \modelcite{modelO4Mini2025} & No & G4, G6 & N/A \\
O4-Mini & \modelcite{modelO4Mini2025} & No & G2, G11, G15 & G6 \\
GPT-5 & \modelcite{modelGPT5Developer2025} & No & G9 & N/A \\
GPT-5-Mini & \modelcite{modelGPT5Developer2025} & No & G1, G4, G5, G7,\newline G9, G10, G12, G14 & N/A \\
GPT-5-Nano & \modelcite{modelGPT5Developer2025} & No & G1, G4, G5, G7,\newline G9, G10, G12, G14 & N/A \\
GPT-5.1 & \modelcite{modelGPT512025} & No & G1, G5, G7, G10 & N/A \\
GPT-5.2 & \modelcite{modelGPT522025} & No & G7, G10 & N/A \\
GPT-5.4 & \modelcite{modelGPT542026} & No & G14 & N/A \\
GPT-5.4-Mini & \modelcite{modelGPT54MiniNano2026} & No & G14 & N/A \\
GPT-5.4-Nano & \modelcite{modelGPT54MiniNano2026} & No & G14 & N/A \\
Claude-2.1 & \modelcite{modelClaude212023} & No & G3 & N/A \\
Claude-3-Opus & \modelcite{modelClaude32024} & No & G2, G3, G13 & G3, G4 \\
Claude-3-Haiku & \modelcite{modelClaude32024} & No & G2, G3, G13 & G3, G4 \\
Claude-3-5-Sonnet & \modelcite{modelClaude35Sonnet2024} & No & G2, G3, G6, G8, G9,\newline G11, G13, G15, G16 & G1, G2, G3, G4, G5, G6 \\
claude-3-5-haiku & \modelcite{modelClaude35Haiku2024} & No & G11 & N/A \\
Claude-3-7-Sonnet & \modelcite{modelClaude37Sonnet2025} & No & G1, G2, G4,\newline G6, G8, G16 & G1, G2, G8 \\
Claude-Opus-4 & \modelcite{modelClaude42025} & No & G11 & N/A \\
Claude-Sonnet-4 & \modelcite{modelClaude42025} & No & G9, G11, G12, G15 & G5, G9 \\
Claude-Opus-4-1 & \modelcite{modelClaudeOpus412025} & No & G1, G4, G5,\newline G9, G10, G15 & G7, G9 \\
Claude-Sonnet-4-5 & \modelcite{modelClaudeSonnet452025} & No & G1, G4, G5,\newline G7, G10, G14 & N/A \\
Claude-Haiku-4-5 & \modelcite{modelClaudeHaiku452025} & No & G1, G4, G5, G7,\newline G10, G12, G14 & N/A \\
Claude-Opus-4-6 & \modelcite{modelClaudeOpus462026} & No & G7 & N/A \\
Claude-Sonnet-4-6 & \modelcite{modelClaudeSonnet462026} & No & G7, G14 & N/A \\
DeepSeek-V3 & \modelcite{modelDeepSeekV32024} & Yes & G6, G8, G16 & G1 \\
DeepSeek-R1 & \modelcite{modelDeepSeekR12025} & Yes & G6, G8, G16 & G1 \\
DeepSeek-V3.1 & \modelcite{modelDeepSeekV312025} & Yes & G1, G4, G5,\newline G7, G10, G12 & N/A \\
Gemini-1.5-Pro & \modelcite{modelGemini15Pro2024} & No & G3 & N/A \\
Gemini-1.5-Flash & \modelcite{modelGemini15Flash2024} & No & G3 & N/A \\
Gemini-2.0-Flash-Lite-001 & \modelcite{modelGemini20FlashLite2025} & No & G8 & N/A \\
Gemini-2.5-Pro-Exp & \modelcite{modelGemini25ProExp2025} & No & G8 & N/A \\
Gemini-2.5-Pro-Preview & \modelcite{modelGeminiAPIChangelog2025} & No & G6 & N/A \\
Gemini-2.5-Flash-Preview & \modelcite{modelGeminiAPIChangelog2025} & No & G6 & N/A \\
Gemini-2.5-Flash & \modelcite{modelGemini25Stable2025} & No & G4, G9 & N/A \\
Gemini-2.5-Pro & \modelcite{modelGemini25Pro2025} & No & G1, G4, G5, G7,\newline G9, G10, G12 & N/A \\
Gemini-3-Pro-Preview & \modelcite{modelGemini32025} & No & G1, G5 & N/A \\
Gemini-3-Flash-Preview & \modelcite{modelGemini3Flash2025} & No & G7, G10 & N/A \\
Gemini-3.1-Pro-Preview & \modelcite{modelGemini31Pro2026} & No & G7, G14 & N/A \\
Kimi-K2-Instruct & \modelcite{modelKimiK22025} & Yes & G9, G11, G15 & G5 \\
Kimi-K2-Thinking & \modelcite{modelKimiK2Thinking2025} & Yes & G5 & N/A \\
Llama-2-70b-Chat-Hf & \modelcite{modelLlama22023} & Yes & G3 & N/A \\
Llama-3-8b-Chat-Hf & \modelcite{modelLlama32024} & Yes & G3, G13 & N/A \\
Llama-3-70b-Chat-Hf & \modelcite{modelLlama32024} & Yes & G3, G13 & G4 \\
Meta-Llama-3.1-405B-Instruct-Turbo & \modelcite{modelLlama312024} & Yes & G2 & N/A \\
Llama-3.2-3B-Instruct-Turbo & \modelcite{modelLlama322024} & Yes & G2 & N/A \\
Llama-3.3-70B-Instruct-Turbo & \modelcite{modelLlama332024} & Yes & G2, G6, G8, G16 & G1, G2 \\
Llama-4-Scout-17B-16E-Instruct & \modelcite{modelLlama42025} & Yes & G2, G6, G8, G11 & G2, G6 \\
Llama-4-Maverick-17B-128E-Instruct-FP8 & \modelcite{modelLlama42025} & Yes & G2, G6, G8, G11 & G2, G6 \\
Mixtral-8x7B-Instruct-V0.1 & \modelcite{modelMixtral8x7B2023} & Yes & G3 & N/A \\
Mixtral-8x22B-Instruct-V0.1 & \modelcite{modelMixtral8x22B2024} & Yes & G3 & N/A \\
Mistral-Large & \modelcite{modelMistralLarge24072024} & Yes & G2, G6, G8, G13 & G1, G2, G3 \\
Magistral-Medium & \modelcite{modelMagistralMedium2025} & No & G9, G11, G15 & G5 \\
Qwen1.5-110B-Chat & \modelcite{modelQwen15110B2024} & Yes & G3 & N/A \\
Qwen2.5-72B-Instruct-Turbo & \modelcite{modelQwen252024} & Yes & G2 & N/A \\
QwQ-32B-Preview & \modelcite{modelQwQPreview2024} & Yes & G6, G8 & G1 \\
Qwen3-235B-A22B-Fp8-Tput & \modelcite{modelQwen32025} & Yes & G1, G2, G4, G9,\newline G11, G13, G15 & G3, G5, G6, G7, G8 \\
Qwen3-235B-A22B-Thinking & \modelcite{modelQwen3Thinking25072025} & Yes & G1, G4, G5, G12 & N/A \\
Grok-beta & \modelcite{modelGrokBeta2024} & No & G8 & N/A \\
Grok-4 & \modelcite{modelGrok42025} & No & G1, G4 & N/A \\
Grok-4-Fast-Non-Reasoning & \modelcite{modelGrok4Fast2025} & No & G1, G4, G5, G12 & N/A \\
Grok-4-Fast-Reasoning & \modelcite{modelGrok4Fast2025} & No & G1, G4, G5, G12 & N/A \\
Grok-4-1-Fast-Non-Reasoning & \modelcite{modelGrok41Fast2025} & No & G1, G5, G7, G10 & N/A \\
Grok-4-1-Fast-Reasoning & \modelcite{modelGrok41Fast2025} & No & G1, G5, G7, G10 & N/A \\
GLM-4.5-Air-FP8 & \modelcite{modelGLM45Air2025} & Yes & G9 & N/A \\

\end{longtable}
\endgroup

\section{Robustness Checks and Controls}
\label{sec:calibration-ablation}

In this section, we conduct two additional analyses: \Cref{apx:N4000} checks whether our main results survive a much stricter shared-coverage requirement, 
while \Cref{apx:calibration} rules out individual recalibration as a simple alternative explanation for our weak-to-strong aggregation improvement.   

\subsection{Higher Shared-Coverage Threshold \texorpdfstring{$N = 4000$}{}}
\label{apx:N4000}


\rxedit{
In the following \Cref{tab:overall-main-results-expanded-4000}, we report statistics at $\groupThreshold=4{,}000$ following the norm of \Cref{tab:overall-main-results-expanded} in \Cref{sec:BS-results}. In particular, we report the overall two-model BS results across the $9$ comparison groups and $97$ eligible pairs. 
Results we report here \aiedit{are} similar to what we have observed at $\groupThreshold=1{,}000$. 
That is, aggregating two weaker models can still outperform the group benchmark at a higher shared-subquestion threshold.

Learned linear pooling identifies at least one pair that matches or improves on the group benchmark in 8 of the 9 groups (88.89\%), and at least one pair within 5\% of the benchmark in all 9 groups.
Learned log-odds pooling matches or improves on the benchmark in 5 groups (55.56\%) and comes within 5\% in 6 groups (66.67\%). These results show that pairs of individually weaker forecasters can perform close to, or better than, the group benchmark across many comparison groups.
}

\begin{table}[h]
\captionsetup{font=normalsize}
\caption{Two-model BS results at $\groupThreshold=4{,}000$.
Cells report
\rxedit{ the group-level and pair-level rates as well as the ``Improve on both members'' described in \Cref{sec:protocol-and-evaluation} in the form of fractions and percentages.
}
}
\label{tab:overall-main-results-expanded-4000}

\resulttablesetup
\resizebox{0.95\linewidth}{!}{
\begin{tabular}{lccccc}
\toprule
& \multicolumn{2}{c}{Groups}
& \multicolumn{3}{c}{Eligible pairs} \\
\cmidrule(lr){2-3}\cmidrule(lr){4-6}
Aggregation rule & \shortstack{Match or improve\\benchmark}
& \shortstack{Within $5\%$\\of benchmark}
& \shortstack{Match or improve\\benchmark}
& \shortstack{Within $5\%$\\of benchmark}
& \shortstack{Improve on\\both members} \\
\midrule
\rowcolor{resultshade}
Learned linear pooling & \shortstack{8/9\\(88.89\%)} & \shortstack{9/9\\(100.00\%)} & \shortstack{65/97\\(67.01\%)} & \shortstack{85/97\\(87.63\%)} & \shortstack{97/97\\(100.00\%)} \\
\rowcolor{resultshade}
Learned log-odds pooling & \shortstack{5/9\\(55.56\%)} & \shortstack{6/9\\(66.67\%)} & \shortstack{15/97\\(15.46\%)} & \shortstack{30/97\\(30.93\%)} & \shortstack{97/97\\(100.00\%)} \\
\midrule
Simple mean (AM) & \shortstack{4/9\\(44.44\%)} & \shortstack{5/9\\(55.56\%)} & \shortstack{12/97\\(12.37\%)} & \shortstack{18/97\\(18.56\%)} & \shortstack{71/97\\(73.20\%)} \\
Harmonic mean (HM) & \shortstack{3/9\\(33.33\%)} & \shortstack{5/9\\(55.56\%)} & \shortstack{4/97\\(4.12\%)} & \shortstack{12/97\\(12.37\%)} & \shortstack{42/97\\(43.30\%)} \\
Log-odds mean (GM) & \shortstack{3/9\\(33.33\%)} & \shortstack{5/9\\(55.56\%)} & \shortstack{7/97\\(7.22\%)} & \shortstack{14/97\\(14.43\%)} & \shortstack{50/97\\(51.55\%)} \\
\midrule
$\alpha$-log-odds & \shortstack{3/9\\(33.33\%)} & \shortstack{5/9\\(55.56\%)} & \shortstack{6/97\\(6.19\%)} & \shortstack{13/97\\(13.40\%)} & \shortstack{40/97\\(41.24\%)} \\
Average prior (AP) & \shortstack{3/9\\(33.33\%)} & \shortstack{5/9\\(55.56\%)} & \shortstack{3/97\\(3.09\%)} & \shortstack{13/97\\(13.40\%)} & \shortstack{42/97\\(43.30\%)} \\
Heuristic prior (HP) & \shortstack{3/9\\(33.33\%)} & \shortstack{5/9\\(55.56\%)} & \shortstack{4/97\\(4.12\%)} & \shortstack{13/97\\(13.40\%)} & \shortstack{46/97\\(47.42\%)} \\
Precision & \shortstack{3/9\\(33.33\%)} & \shortstack{5/9\\(55.56\%)} & \shortstack{6/97\\(6.19\%)} & \shortstack{13/97\\(13.40\%)} & \shortstack{49/97\\(50.52\%)} \\
\bottomrule
\end{tabular}
}
\end{table}

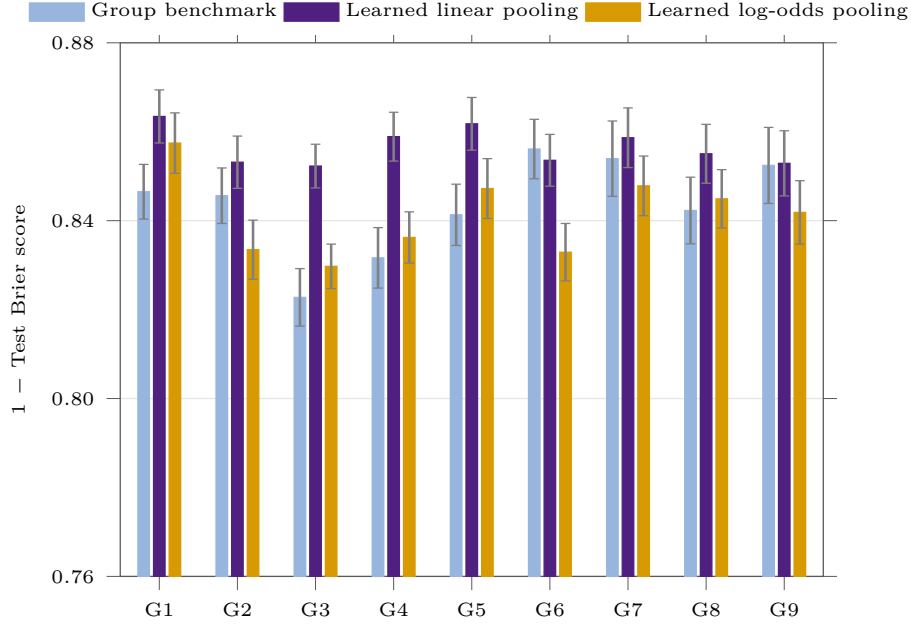
\begin{figure}[h]
   \centering
   \resizebox{0.75\linewidth}{!}{%
      \begingroup
\begin{tikzpicture}
\begin{axis}[
forecast axes front,
width=0.5625\linewidth,height=0.45\linewidth,
xmin=0.5,xmax=9.5,ymin=0.76,ymax=0.88,
xtick={1,...,9},
xticklabels={G1,G2,G3,G4,G5,G6,G7,G8,G9},
ytick={0.76,0.80,0.84,0.88},
yticklabel style={/pgf/number format/fixed zerofill,/pgf/number format/precision=2},
ylabel={1 $-$ Test Brier score},
tick label style={font=\fontsize{6}{7}\selectfont},
label style={font=\fontsize{6}{7}\selectfont},
scaled ticks=false,ymajorgrids=true,grid style={black!10},tick align=outside,
legend style={at={(0.5,1.015)},anchor=south,draw=none,font=\fontsize{6}{7}\selectfont,legend columns=3},
bar width=3.6pt,
]

\addplot[
forget plot,ybar,bar shift=0pt,
fill=bsindividual,draw=bsindividual,mark=none,
error bars/.cd,y dir=both,y explicit,error mark=-,
error bar style={gray,line width=0.8pt},
error mark options={mark size=0.20pt,line width=3pt}
] coordinates {
  (0.80,0.846507622718) += (0,0.006149649805) -= (0,0.006149649805)
  (1.80,0.845612010579) += (0,0.006243698461) -= (0,0.006243698461)
  (2.80,0.822749516641) += (0,0.006475288585) -= (0,0.006475288585)
  (3.80,0.831639026053) += (0,0.006799190307) -= (0,0.006799190307)
  (4.80,0.841315923786) += (0,0.006888627751) -= (0,0.006888627751)
  (5.80,0.856114836976) += (0,0.006684512260) -= (0,0.006684512260)
  (6.80,0.853942237656) += (0,0.008472908397) -= (0,0.008472908397)
  (7.80,0.842285515175) += (0,0.007494663503) -= (0,0.007494663503)
  (8.80,0.852425961775) += (0,0.008551409135) -= (0,0.008551409135)
};
\addlegendimage{area legend,fill=bsindividual,draw=bsindividual}
\addlegendentry{Group benchmark}

\addplot[
forget plot,ybar,bar shift=0pt,
fill=bslinear,draw=bslinear,mark=none,
error bars/.cd,y dir=both,y explicit,error mark=-,
error bar style={gray,line width=0.8pt},
error mark options={mark size=0.20pt,line width=3pt}
] coordinates {
  (1.00,0.863438980684) += (0,0.005967085734) -= (0,0.005967085734)
  (2.00,0.853150840021) += (0,0.005874904540) -= (0,0.005874904540)
  (3.00,0.852280103832) += (0,0.004912432189) -= (0,0.004912432189)
  (4.00,0.858891939639) += (0,0.005490530158) -= (0,0.005490530158)
  (5.00,0.861782932923) += (0,0.005926891172) -= (0,0.005926891172)
  (6.00,0.853575575240) += (0,0.005829509310) -= (0,0.005829509310)
  (7.00,0.858651879833) += (0,0.006713524348) -= (0,0.006713524348)
  (8.00,0.855041474279) += (0,0.006621321264) -= (0,0.006621321264)
  (9.00,0.852893627383) += (0,0.007325768788) -= (0,0.007325768788)
};
\addlegendimage{area legend,fill=bslinear,draw=bslinear}
\addlegendentry{Learned linear pooling}

\addplot[
forget plot,ybar,bar shift=0pt,
fill=bslogodds,draw=bslogodds,mark=none,
error bars/.cd,y dir=both,y explicit,error mark=-,
error bar style={gray,line width=0.8pt},
error mark options={mark size=0.20pt,line width=3pt}
] coordinates {
  (1.20,0.857448939470) += (0,0.006795193779) -= (0,0.006795193779)
  (2.20,0.833475588903) += (0,0.006646816302) -= (0,0.006646816302)
  (3.20,0.829732603297) += (0,0.004997287114) -= (0,0.004997287114)
  (4.20,0.836226168081) += (0,0.005757522454) -= (0,0.005757522454)
  (5.20,0.847219952305) += (0,0.006730552668) -= (0,0.006730552668)
  (6.20,0.832910849389) += (0,0.006464043262) -= (0,0.006464043262)
  (7.20,0.847829538121) += (0,0.006704740429) -= (0,0.006704740429)
  (8.20,0.844908740791) += (0,0.006567247007) -= (0,0.006567247007)
  (9.20,0.841861010536) += (0,0.007134657836) -= (0,0.007134657836)
};
\addlegendimage{area legend,fill=bslogodds,draw=bslogodds}
\addlegendentry{Learned log-odds pooling}
\end{axis}
\end{tikzpicture}
\endgroup%
   }
   \caption{Best learned pair versus the group benchmark across the 9 comparison groups at \(\groupThreshold=4{,}000\).
   For each group, we plot test 1 $-$ BS for the group benchmark and the lowest-BS pair under learned linear and log-odds pooling, selected separately for each rule. Higher bars (1 $-$ BS) are better; groups are ordered by decreasing number of forecasters.}
   \label{fig:4000-brier-horizon}
\end{figure}

\rxedit{In \Cref{fig:4000-brier-horizon}, we visualize these group-level results. 
For each comparison group, we report the group benchmark together with the lowest-BS pair under each learned rule, where the best pair is selected separately for linear and log-odds pooling. 
A best-pair point at or above the benchmark bar indicates that the corresponding group contains a successful pair. 
We thus observe directly from \Cref{fig:4000-brier-horizon} that learned linear pooling produces a successful pair in 8 groups while learned log-odds pooling in 5 groups and pairwise aggregation is more significant in G3, G4 and G5.} 

\rxedit{In the following \Cref{tab:group-best-individual-top2-pairs-brier-4000}, we report the group benchmark and the best weak-pair aggregate under each learned rule for each group of $\groupThreshold=4{,}000$, selected by test Brier score (lower is better).
Each Brier entry shows the point estimate and symmetric 95\% CI half-width from 5,000 test-subquestion bootstrap resamples with the selected models and training-fitted coefficients held fixed. The intervals condition on the test-selected pairs. Each pair reports its training-fitted coefficients $(\omega_1,\omega_2)$ in the displayed model order, rounded to four decimals; coefficients are nonnegative and need not sum to one. Bold pair scores are smaller than the best individual's score, with comparisons made before rounding.
}
\begingroup
\setlength{\tabcolsep}{2.0pt}
\renewcommand{\arraystretch}{1.08}
\fontsize{6.0}{7.0}\selectfont
\newcommand{\bestpairline}[1]{%
    \begingroup
    \sbox0{#1}%
    \ifdim\wd0>\linewidth
        \resizebox{\linewidth}{!}{\usebox0}%
    \else
        \usebox0%
    \fi
    \endgroup
}
\newcommand{\brierci}[2]{%
    \begin{tabular}[t]{@{}r@{}}
        #1\\
        $(\pm\,#2)$
    \end{tabular}%
}
\newcommand{\bestpair}[4]{%
    \bestpairline{#1~$+$}\newline
    \bestpairline{#2}\newline
    \bestpairline{$(\omega_1,\omega_2)=(#3,#4)$}%
}
\begin{longtable}{@{}l l@{\hspace{4pt}}
    >{\raggedright\arraybackslash}p{22mm}
    @{\hspace{3pt}}r@{\hspace{4pt}}
    >{\raggedright\arraybackslash}p{34mm}
    @{\hspace{3pt}}r@{\hspace{4pt}}
    >{\raggedright\arraybackslash}p{34mm}
    @{\hspace{3pt}}r@{}
    }
\caption{Group benchmark and the best weak-pair aggregate under each pooling method for each \rxedit{of the size-sorted overall groups at $\groupThreshold=4{,}000$}, selected by test Brier score. }\label{tab:group-best-individual-top2-pairs-brier-4000}
\\
\toprule
Group & Size & Group Benchmark & \shortstack{Brier\\(95\% CI)} & \shortstack{Log-odds pooling\\best pair} & \shortstack{Brier\\(95\% CI)} & \shortstack{Linear pooling\\best pair} & \shortstack{Brier\\(95\% CI)} \\
\midrule
\endfirsthead
\toprule
Group & Size & Group Benchmark & \shortstack{Brier\\(95\% CI)} & \shortstack{Log-odds pooling\\best pair} & \shortstack{Brier\\(95\% CI)} & \shortstack{Linear pooling\\best pair} & \shortstack{Brier\\(95\% CI)} \\
\midrule
\endhead
\midrule
\multicolumn{8}{r}{\fontsize{6.0}{7.0}\selectfont Continued on next page}\\
\endfoot
\bottomrule
\endlastfoot
\multicolumn{8}{@{}l}{\textbf{Overall, threshold 4,000}}\\
\addlinespace[1pt]
1 & 9 & Llama-3.3-70B-Instruct-Turbo & \brierci{0.153}{0.006} 
& \bestpair{O3-Mini}{QwQ-32B-Preview}{0.3409}{0.5087} 
& \brierci{\textbf{0.143}}{0.007} 
& \bestpair{Claude-3-5-Sonnet}{QwQ-32B-Preview}{0.3828}{0.4115} 
& \brierci{\textbf{0.137}}{0.006} \\
2 & 8 & Llama-3.3-70B-Instruct-Turbo & \brierci{0.154}{0.006} 
& \bestpair{Claude-3-7-Sonnet}{GPT-4.1}{0.4461}{0.2871} 
& \brierci{0.167}{0.007} 
& \bestpair{Claude-3-5-Sonnet}{GPT-4.1}{0.4196}{0.2509} 
& \brierci{\textbf{0.147}}{0.006} \\
3 & 7 & Claude-3-Opus & \brierci{0.177}{0.006} 
& \bestpair{Claude-3-5-Sonnet}{Mistral-Large}{0.4287}{0.3415} 
& \brierci{\textbf{0.170}}{0.005} 
& \bestpair{Claude-3-5-Sonnet}{Mistral-Large}{0.5215}{0.1413} 
& \brierci{\textbf{0.148}}{0.005} \\
4 & 6 & Claude-3-Opus & \brierci{0.168}{0.007} 
& \bestpair{Claude-3-5-Sonnet}{GPT-4-Turbo}{0.4946}{0.3134} 
& \brierci{\textbf{0.164}}{0.006} 
& \bestpair{Claude-3-5-Sonnet}{Llama-3-70b-Chat-Hf}{0.5947}{0.0749} 
& \brierci{\textbf{0.141}}{0.005} \\
5 & 6 & Magistral-Medium & \brierci{0.159}{0.007} 
& \bestpair{Claude-3-5-Sonnet}{GPT-4.1}{0.3594}{0.4522} 
& \brierci{\textbf{0.153}}{0.007} 
& \bestpair{Claude-3-5-Sonnet}{GPT-4.1}{0.3725}{0.3621} 
& \brierci{\textbf{0.138}}{0.006} \\
6 & 6 & O4-Mini & \brierci{0.144}{0.007} 
& \bestpair{Claude-3-5-Sonnet}{GPT-4.1}{0.3313}{0.3399} 
& \brierci{0.167}{0.006} 
& \bestpair{Claude-3-5-Sonnet}{GPT-4.1}{0.4427}{0.2476} 
& \brierci{0.146}{0.006} \\
7 & 3 & GPT-4.1 & \brierci{0.146}{0.008} 
& \bestpair{Claude-Opus-4-1}{Qwen3-235B-A22B-Fp8-Tput}{0.4916}{0.2936} 
& \brierci{0.152}{0.007} 
& \bestpair{Claude-Opus-4-1}{Qwen3-235B-A22B-Fp8-Tput}{0.5025}{0.2241} 
& \brierci{\textbf{0.141}}{0.007} \\
8 & 3 & Claude-3-7-Sonnet & \brierci{0.158}{0.007} 
& \bestpair{GPT-4.1}{Qwen3-235B-A22B-Fp8-Tput}{0.4500}{0.2026} 
& \brierci{\textbf{0.155}}{0.007} 
& \bestpair{GPT-4.1}{Qwen3-235B-A22B-Fp8-Tput}{0.4760}{0.2545} 
& \brierci{\textbf{0.145}}{0.007} \\
9 & 3 & GPT-4.1 & \brierci{0.148}{0.009} 
& \bestpair{Claude-Opus-4-1}{Claude-Sonnet-4}{0.4047}{0.4146} 
& \brierci{0.158}{0.007} 
& \bestpair{Claude-Opus-4-1}{Claude-Sonnet-4}{0.4455}{0.2827} 
& \brierci{\textbf{0.147}}{0.007} \\

\end{longtable}
\endgroup

\subsection{Individual Recalibration as a Control}
\label{apx:calibration}
\rxedit{We use an individual-recalibration control to examine whether fitted probability rescaling alone can explain the BS gains under learned linear pooling. In each of the 16 overall comparison groups, we compare the lowest-test-BS calibrated weak forecaster with the best aggregation pair in the group and the group benchmark.}

We calibrate each individually weaker forecaster separately using two-parameter Platt scaling, following the post-processing form studied by \citet{M-26}. For each forecaster, we fit a slope $a$ and an intercept $b$ on its group's original training set by minimizing instance-weighted log loss. We transform its forecast probability $p$ as $\tilde p=\sigma(a\operatorname{logit}(p)+b)$. We then keep the fitted parameters fixed and evaluate the calibrated forecast on the group's common test set.


\rxedit{Following the norm of the evaluation procedure in \Cref{sec:BS-results}}, we compare these calibrated \rxedit{forecasters} with the group benchmark and the aggregation pairs reported in \Cref{sec:BS-results}.
In \Cref{tab:calibration-ablation-group-wins}, we report the fraction of the 16 comparison groups in which the best calibrated forecaster matches or improves on the group benchmark or the best aggregation pair of the group. 

\begin{table}[ht]
\captionsetup{font=normalsize}
\caption{Group-level BS results for individual recalibration at $\groupThreshold=1{,}000$. Cells report fractions and percentages of the 16 overall comparison groups in which the lowest-test-BS calibrated forecaster matches or improves on the group benchmark or the best aggregation pair of the group.}
\label{tab:calibration-ablation-group-wins}
\resulttablesetup
\resizebox{0.545\linewidth}{!}{%
\begin{tabular}{@{}lcc@{}}
\toprule
& \multicolumn{2}{c}{Comparison groups} \\
\cmidrule(lr){2-3}
Calibration rule
& \shortstack{Match or improve\\group benchmark}
& \shortstack{Match or improve\\best aggregation pair} \\
\midrule
Per-model Platt scaling
& \shortstack{7/16\\(43.75\%)}
& \shortstack{2/16\\(12.50\%)} \\
\bottomrule
\end{tabular}

}
\end{table}

\xhdr{Individual recalibration is not that competent}
Using the same match-or-improve criterion as in \Cref{sec:protocol-and-evaluation}, we find that the best calibrated forecaster matches or improves on the group benchmark in 7 of the 16 groups (43.75\%) and on the best aggregation pair in 2 groups (12.50\%).  

These results show that individual-recalibration does not reproduce the best aggregation pair's test BS in most comparison groups. These results show that aggregation can improve forecast accuracy beyond individual recalibration.

\section{Additional Pairwise Aggregation Results}
\label{apx:pairwise-agg-materials}

\xhdr{Best pairs across comparison groups}
\rxedit{In \Cref{fig:overall-size-sorted-thresholds}, we visualize the group-level results reported in \Cref{sec:main-results}. 
For each comparison group, we report the group benchmark together with the lowest-BS pair under each learned rule, where the best pair is selected separately for linear and log-odds pooling. 
A best-pair point at or above the benchmark bar indicates that the corresponding group contains a successful pair. 
We thus observe directly from \Cref{fig:overall-size-sorted-thresholds} that learned linear pooling produces a successful pair in 11 groups while learned log-odds pooling in 8 groups and pairwise aggregation is more significant in G3, G6, G8, G13 and G16. The figure also shows that their relative performance may vary across groups: linear pooling succeeds where log-odds pooling does not in G9, G12, G15, and G16, while the reverse occurs in G11.}


\rxedit{In the following \Cref{tab:group-best-individual-top2-pairs-brier}, we report the group benchmark and the best weak-pair aggregate under each learned rule for each group of $\groupThreshold=1{,}000$, selected by test Brier score (lower is better).
Each Brier entry shows the point estimate and symmetric 95\% CI half-width from 5,000 test-subquestion bootstrap resamples with the selected models and training-fitted coefficients held fixed. The intervals condition on the test-selected pairs. Each pair reports its training-fitted coefficients $(\omega_1,\omega_2)$ in the displayed model order, rounded to four decimals; coefficients are nonnegative and need not sum to one. Bold pair scores are smaller than the best individual's score, with comparisons made before rounding.
}

\begingroup
\setlength{\tabcolsep}{2.0pt}
\renewcommand{\arraystretch}{1.08}
\fontsize{6.0}{7.0}\selectfont
\newcommand{\bestpairline}[1]{%
    \begingroup
    \sbox0{#1}%
    \ifdim\wd0>\linewidth
        \resizebox{\linewidth}{!}{\usebox0}%
    \else
        \usebox0%
    \fi
    \endgroup
}
\newcommand{\brierci}[2]{%
    \begin{tabular}[t]{@{}r@{}}
        #1\\
        $(\pm\,#2)$
    \end{tabular}%
}
\newcommand{\bestpair}[4]{%
    \bestpairline{#1~$+$}\newline
    \bestpairline{#2}\newline
    \bestpairline{$(\omega_1,\omega_2)=(#3,#4)$}%
}
\begin{longtable}{@{}l l@{\hspace{4pt}}
    >{\raggedright\arraybackslash}p{22mm}
    @{\hspace{3pt}}r@{\hspace{4pt}}
    >{\raggedright\arraybackslash}p{34mm}
    @{\hspace{3pt}}r@{\hspace{4pt}}
    >{\raggedright\arraybackslash}p{34mm}
    @{\hspace{3pt}}r@{}
    }
\caption{Group benchmark and the best weak-pair aggregate under each pooling method for each \rxedit{of the size-sorted overall groups at $\groupThreshold=1{,}000$}, selected by test Brier score. }\label{tab:group-best-individual-top2-pairs-brier}
\\
\toprule
Group & Size & Group Benchmark & \shortstack{Brier\\(95\% CI)} & \shortstack{Log-odds pooling\\best pair} & \shortstack{Brier\\(95\% CI)} & \shortstack{Linear pooling\\best pair} & \shortstack{Brier\\(95\% CI)} \\
\midrule
\endfirsthead
\toprule
Group & Size & Group Benchmark & \shortstack{Brier\\(95\% CI)} & \shortstack{Log-odds pooling\\best pair} & \shortstack{Brier\\(95\% CI)} & \shortstack{Linear pooling\\best pair} & \shortstack{Brier\\(95\% CI)} \\
\midrule
\endhead
\midrule
\multicolumn{8}{r}{\fontsize{6.0}{7.0}\selectfont Continued on next page}\\
\endfoot
\bottomrule
\endlastfoot
\multicolumn{8}{@{}l}{\textbf{Overall, threshold 1,000}}\\
\addlinespace[1pt]
1 & 18 & Gemini-3-Pro-Preview & \brierci{0.115}{0.012} 
& \bestpair{Claude-3-7-Sonnet}{GPT-5-Mini}{0.6117}{0.4237} 
& \brierci{0.118}{0.012}
& \bestpair{Claude-Opus-4-1}{GPT-5-Mini}{0.4390}{0.5893} 
& \brierci{0.118}{0.012} \\
2 & 16 & O4-Mini & \brierci{0.147}{0.009} 
& \bestpair{Llama-3.3-70B-Instruct-Turbo}{Meta-Llama-3.1-405B-Instruct-Turbo}{1.0825}{0.4939} 
& \brierci{\textbf{0.139}}{0.009}
& \bestpair{Llama-3.3-70B-Instruct-Turbo}{Meta-Llama-3.1-405B-Instruct-Turbo}{0.3395}{0.4624} 
& \brierci{\textbf{0.143}}{0.008} \\
3 & 16 & Gemini-1.5-Pro & \brierci{0.171}{0.010} 
& \bestpair{Claude-3-5-Sonnet}{Llama-3-8b-Chat-Hf}{0.6165}{0.2859} 
& \brierci{\textbf{0.170}}{0.010}
& \bestpair{Claude-3-5-Sonnet}{GPT-4o}{0.5320}{0.1326} 
& \brierci{\textbf{0.148}}{0.009} \\
4 & 16 & GPT-5-Mini & \brierci{0.120}{0.012} 
& \bestpair{Claude-3-7-Sonnet}{Grok-4}{0.6628}{0.3542} 
& \brierci{\textbf{0.118}}{0.011}
& \bestpair{Claude-Opus-4-1}{Grok-4}{0.5310}{0.4611} 
& \brierci{\textbf{0.118}}{0.011} \\
5 & 16 & Gemini-3-Pro-Preview & \brierci{0.125}{0.011} 
& \bestpair{DeepSeek-V3.1}{GPT-5-Mini}{0.5378}{0.5612} 
& \brierci{0.126}{0.011}
& \bestpair{Claude-Opus-4-1}{GPT-5-Mini}{0.4005}{0.6186} 
& \brierci{0.126}{0.011} \\
6 & 15 & Llama-3.3-70B-Instruct-Turbo & \brierci{0.179}{0.012} 
& \bestpair{DeepSeek-R1}{QwQ-32B-Preview}{0.3993}{0.3646} 
& \brierci{\textbf{0.167}}{0.012} 
& \bestpair{DeepSeek-R1}{Gemini-2.5-Flash-Preview}{0.4370}{0.3543} 
& \brierci{\textbf{0.158}}{0.010} \\
7 & 15 & Grok-4-1-Fast-Reasoning & \brierci{0.147}{0.014} 
& \bestpair{Claude-Haiku-4-5}{GPT-5-Nano}{0.7454}{0.2705} 
& \brierci{\textbf{0.145}}{0.013} 
& \bestpair{GPT-5.2}{Gemini-3.1-Pro-Preview}{0.5315}{0.4918} 
& \brierci{\textbf{0.146}}{0.013} \\
8 & 14 & Llama-3.3-70B-Instruct-Turbo & \brierci{0.177}{0.011} 
& \bestpair{O3-Mini}{QwQ-32B-Preview}{0.3661}{0.4486} 
& \brierci{\textbf{0.164}}{0.013} 
& \bestpair{Claude-3-5-Sonnet}{QwQ-32B-Preview}{0.3989}{0.3729} 
& \brierci{\textbf{0.157}}{0.011} \\
9 & 13 & GPT-5 & \brierci{0.129}{0.010} 
& \bestpair{Claude-Opus-4-1}{Gemini-2.5-Pro}{0.6456}{0.3601} 
& \brierci{0.134}{0.011} 
& \bestpair{Claude-Opus-4-1}{Gemini-2.5-Pro}{0.4733}{0.3017} 
& \brierci{\textbf{0.128}}{0.009} \\
10 & 13 & Grok-4-1-Fast-Reasoning & \brierci{0.135}{0.012} 
& \bestpair{Claude-Haiku-4-5}{GPT-5-Mini}{0.4209}{0.6278} 
& \brierci{\textbf{0.135}}{0.012} 
& \bestpair{GPT-5-Mini}{Gemini-2.5-Pro}{0.4174}{0.5728} 
& \brierci{\textbf{0.135}}{0.012} \\
11 & 11 & O4-Mini & \brierci{0.132}{0.010} 
& \bestpair{Claude-Opus-4}{claude-3-5-haiku}{0.5125}{0.7849} 
& \brierci{\textbf{0.131}}{0.011} 
& \bestpair{Claude-Opus-4}{Magistral-Medium}{0.5112}{0.2484} 
& \brierci{0.133}{0.009} \\
12 & 10 & Gemini-2.5-Pro & \brierci{0.127}{0.013} 
& \bestpair{GPT-4.1}{GPT-5-Mini}{0.1859}{0.6261} 
& \brierci{0.128}{0.012} 
& \bestpair{GPT-4.1}{GPT-5-Mini}{0.2663}{0.7316} 
& \brierci{\textbf{0.127}}{0.013} \\
13 & 9 & Claude-3-Opus & \brierci{0.180}{0.012} 
& \bestpair{Claude-3-5-Sonnet}{Mistral-Large}{0.4045}{0.4091} 
& \brierci{\textbf{0.173}}{0.010} 
& \bestpair{Claude-3-5-Sonnet}{Mistral-Large}{0.4535}{0.2284} 
& \brierci{\textbf{0.152}}{0.009} \\
14 & 9 & Gemini-3.1-Pro-Preview & \brierci{0.155}{0.013} 
& \bestpair{Claude-Haiku-4-5}{Claude-Sonnet-4-6}{0.1206}{0.7256} 
& \brierci{0.157}{0.012} 
& \bestpair{Claude-Sonnet-4-6}{GPT-5-Nano}{0.9308}{0.1264} 
& \brierci{0.156}{0.013} \\
15 & 8 & O4-Mini & \brierci{0.143}{0.011} 
& \bestpair{Claude-3-5-Sonnet}{Magistral-Medium}{0.5238}{0.5429} 
& \brierci{0.147}{0.010} 
& \bestpair{Claude-3-5-Sonnet}{GPT-4.1}{0.4780}{0.2474} 
& \brierci{\textbf{0.137}}{0.010} \\
16 & 7 & Llama-3.3-70B-Instruct-Turbo & \brierci{0.165}{0.011} 
& \bestpair{Claude-3-7-Sonnet}{DeepSeek-R1}{0.4528}{0.4301} 
& \brierci{0.174}{0.010} 
& \bestpair{DeepSeek-R1}{GPT-4.5-Preview}{0.2698}{0.4374} 
& \brierci{\textbf{0.155}}{0.009} 

\end{longtable}
\endgroup

\xhdr{Robust aggregation baselines}
\rxedit{
\Cref{tab:overall-main-results-expanded-continued} below is regarded as an extension of \Cref{tab:overall-main-results-expanded}\aiedit{;} we report the overall two-model BS results across the 16 comparison groups and 1,121 eligible pairs for each robust aggregation rule. In particular, we report the fraction of groups and pairs that match or improve on the group benchmark, the corresponding fractions within 5\% of the benchmark, and the fraction of pairs whose aggregate improves on both constituent forecasters in the same format as \Cref{tab:overall-main-results-expanded}.
}
\newsavebox{\apxMainResultsContinuedBox}
\begingroup
\begin{table}[H]
\centering
\setlength{\tabcolsep}{2.0pt}
\renewcommand{\arraystretch}{1.08}
\fontsize{8.0}{9.5}\selectfont
\captionsetup{font=normalsize}
\caption{Additional robust-rule results corresponding to \Cref{tab:overall-main-results-expanded}}
\label{tab:overall-main-results-expanded-continued}
\sbox{\apxMainResultsContinuedBox}{%
\begin{tabular}{lccccc}
\toprule
& \multicolumn{2}{c}{Groups}
& \multicolumn{3}{c}{Eligible pairs} \\
\cmidrule(lr){2-3}\cmidrule(lr){4-6}
Aggregation rule & \shortstack{Match or improve\\benchmark}
& \shortstack{Within $5\%$\\of benchmark}
& \shortstack{Match or improve\\benchmark}
& \shortstack{Within $5\%$\\of benchmark}
& \shortstack{Improve on\\both members} \\
\midrule
$\alpha$-log-odds & \shortstack{9/16\\(56.25\%)} & \shortstack{15/16\\(93.75\%)} & \shortstack{44/1,121\\(3.93\%)} & \shortstack{351/1,121\\(31.31\%)} & \shortstack{605/1,121\\(53.97\%)} \\
Average prior (AP) & \shortstack{5/16\\(31.25\%)} & \shortstack{14/16\\(87.50\%)} & \shortstack{40/1,121\\(3.57\%)} & \shortstack{336/1,121\\(29.97\%)} & \shortstack{600/1,121\\(53.52\%)} \\
Heuristic prior (HP) & \shortstack{6/16\\(37.50\%)} & \shortstack{14/16\\(87.50\%)} & \shortstack{36/1,121\\(3.21\%)} & \shortstack{326/1,121\\(29.08\%)} & \shortstack{622/1,121\\(55.49\%)} \\
Precision & \shortstack{10/16\\(62.50\%)} & \shortstack{15/16\\(93.75\%)} & \shortstack{48/1,121\\(4.28\%)} & \shortstack{362/1,121\\(32.29\%)} & \shortstack{658/1,121\\(58.70\%)} \\
\bottomrule
\end{tabular}
}
\ifdim\wd\apxMainResultsContinuedBox>\linewidth
  \resizebox{\linewidth}{!}{\usebox{\apxMainResultsContinuedBox}}%
\else
  \resizebox{\wd\apxMainResultsContinuedBox}{!}{\usebox{\apxMainResultsContinuedBox}}%
\fi
\end{table}
\endgroup

\begin{figure}[h]
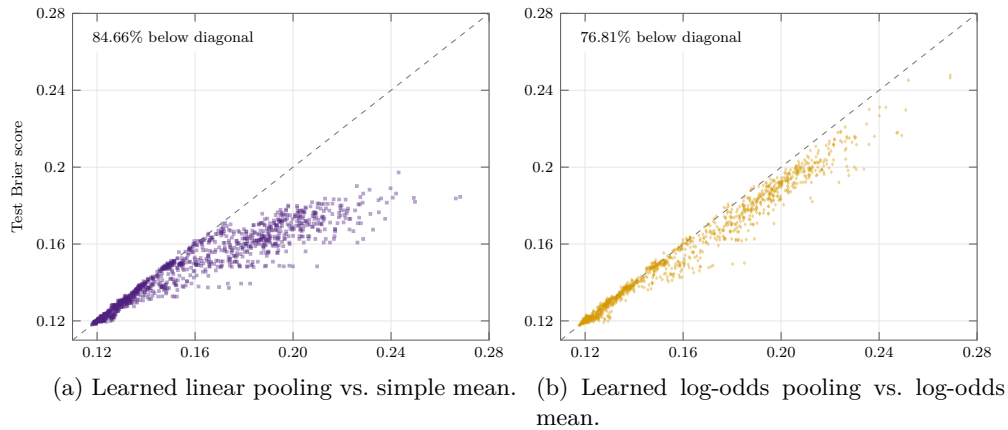

\centering
\hspace*{18pt}
\subfloat[Learned linear pooling vs.\ simple mean.%
\label{fig:learned-vs-simple}]{%
    \hspace*{-18pt}\resizebox{!}{4.8cm}{\input{Paper/figs/sec4-learned-vs-means-a-purple-simple-mean}}%
}
\hspace*{7pt}
\subfloat[Learned log-odds pooling vs.\ log-odds mean.%
\label{fig:learned-vs-logodds-mean}]{%
    \hspace*{-7pt}\resizebox{!}{4.8cm}{\input{Paper/figs/sec4-learned-vs-means-b-gold-logodds-mean}}%
}
\caption{Learned aggregation versus the corresponding equal-weight rule on the same 1,121 pairs. Each point compares the test BS of a learned aggregate with its equal-weight counterpart. Points below the diagonal indicate lower BS under the learned rule.}
\label{fig:sec4-learned-vs-means}
\end{figure}

\xhdr{Learned versus equal-weight pooling}
\rxedit{We compare each learned rule with its equal-weight counterpart on the same 1,121 pairs in the following \Cref{fig:sec4-learned-vs-means}.}
\rxedit{
The horizontal axis reports the BS under the equal-weight rule, while the vertical axis reports the BS under the corresponding learned rule.
A point below the diagonal therefore indicates that learning the coefficients reduces BS for that pair. 
We find that learned linear pooling has lower BS than the simple mean for 949 of the 1,121 pairs (84.66\%), while learned log-odds pooling has lower BS than the log-odds mean for 861 pairs (76.81\%). 
These matched comparisons show that the benefit of learning extends well beyond the pairs that cross the group benchmark.
The figure also shows that learning does not improve every pair, since some points lie above the diagonal. Overall, however, most pairs have lower BS under the learned rules than under their equal-weight counterparts.
}

\raggedbottom
\xhdr{Near-best and no-near pairs}
\rxedit{
In the following \Cref{tab:no-near-pairs}, we demonstrate the near/no-near decomposition of the group- and pair-level within-$5\%$ BS results at $\groupThreshold=1{,}000$. 
Group-level columns report, among the $16$ comparison groups,
the fraction containing a pair within $5\%$ of the benchmark and, separately, the fractions of the within $5\%$ groups whose pair set includes at least one near pair or at least one no-near pair. We note that a group may contain both types, so the latter two fractions need not sum to one.
Pair-level columns decompose the pairs within $5\%$ of
the benchmark into near and no-near pairs, together with the fraction of all pairs meeting the criterion.
}

\newsavebox{\apxNearTableBox}
\begingroup
\begin{table}[H]
\centering
\setlength{\tabcolsep}{2.0pt}
\renewcommand{\arraystretch}{1.08}
\fontsize{8.0}{9.5}\selectfont
\captionsetup{font=normalsize}
\caption{Near/no-near decomposition of the group- and pair-level within-$5\%$ BS results at $\groupThreshold=1{,}000$.}
\label{tab:no-near-pairs}
\sbox{\apxNearTableBox}{%
\begin{tabular}{lcccccc}
\toprule
& \multicolumn{3}{c}{Groups with a pair within $5\%$}
& \multicolumn{3}{c}{Pairs within $5\%$} \\
\cmidrule(lr){2-4}\cmidrule(lr){5-7}
Aggregation rule
& \shortstack{Fraction of\\all groups}
& \shortstack{Share with a\\near pair\\within $5\%$}
& \shortstack{Share with a\\no-near pair\\within $5\%$}
& \shortstack{Fraction of\\all pairs}
& \shortstack{Near-pair\\share}
& \shortstack{No-near-pair\\share} \\
\midrule
\rowcolor{resultshade}
Learned linear pooling
& \shortstack{16/16\\(100.00\%)}
& \shortstack{9/16\\(56.25\%)}
& \shortstack{15/16\\(93.75\%)}
& \shortstack{730/1,121\\(65.12\%)}
& \shortstack{323/730\\(44.25\%)}
& \shortstack{407/730\\(55.75\%)} \\
\rowcolor{resultshade}
Learned log-odds pooling
& \shortstack{15/16\\(93.75\%)}
& \shortstack{9/15\\(60.00\%)}
& \shortstack{15/15\\(100.00\%)}
& \shortstack{467/1,121\\(41.66\%)}
& \shortstack{320/467\\(68.52\%)}
& \shortstack{147/467\\(31.48\%)} \\
\midrule
Simple mean (AM)
& \shortstack{15/16\\(93.75\%)}
& \shortstack{9/15\\(60.00\%)}
& \shortstack{15/15\\(100.00\%)}
& \shortstack{345/1,121\\(30.78\%)}
& \shortstack{250/345\\(72.46\%)}
& \shortstack{95/345\\(27.54\%)} \\
Harmonic mean (HM)
& \shortstack{16/16\\(100.00\%)}
& \shortstack{9/16\\(56.25\%)}
& \shortstack{15/16\\(93.75\%)}
& \shortstack{349/1,121\\(31.13\%)}
& \shortstack{263/349\\(75.36\%)}
& \shortstack{86/349\\(24.64\%)} \\
Log-odds mean (GM)
& \shortstack{15/16\\(93.75\%)}
& \shortstack{9/15\\(60.00\%)}
& \shortstack{15/15\\(100.00\%)}
& \shortstack{375/1,121\\(33.45\%)}
& \shortstack{262/375\\(69.87\%)}
& \shortstack{113/375\\(30.13\%)} \\
\midrule
$\alpha$-log-odds
& \shortstack{15/16\\(93.75\%)}
& \shortstack{9/15\\(60.00\%)}
& \shortstack{15/15\\(100.00\%)}
& \shortstack{351/1,121\\(31.31\%)}
& \shortstack{255/351\\(72.65\%)}
& \shortstack{96/351\\(27.35\%)} \\
Average prior (AP)
& \shortstack{14/16\\(87.50\%)}
& \shortstack{9/14\\(64.29\%)}
& \shortstack{14/14\\(100.00\%)}
& \shortstack{336/1,121\\(29.97\%)}
& \shortstack{251/336\\(74.70\%)}
& \shortstack{85/336\\(25.30\%)} \\
Heuristic prior (HP)
& \shortstack{14/16\\(87.50\%)}
& \shortstack{9/14\\(64.29\%)}
& \shortstack{14/14\\(100.00\%)}
& \shortstack{326/1,121\\(29.08\%)}
& \shortstack{251/326\\(76.99\%)}
& \shortstack{75/326\\(23.01\%)} \\
Precision
& \shortstack{15/16\\(93.75\%)}
& \shortstack{9/15\\(60.00\%)}
& \shortstack{15/15\\(100.00\%)}
& \shortstack{362/1,121\\(32.29\%)}
& \shortstack{261/362\\(72.10\%)}
& \shortstack{101/362\\(27.90\%)} \\
\bottomrule
\end{tabular}}
\ifdim\wd\apxNearTableBox>\linewidth
  \resizebox{\linewidth}{!}{\usebox{\apxNearTableBox}}%
\else
  \resizebox{\wd\apxNearTableBox}{!}{\usebox{\apxNearTableBox}}%
\fi
\end{table}
\endgroup

\xhdr{Calibration results}
\rxedit{
In \Cref{tab:overall-ece-brier-trained} below, we report the group and pair level results evaluated by ECE. We retain the BS-best benchmark, eligible pairs, and BS-trained coefficients used in \Cref{tab:overall-main-results-expanded} and evaluate the same forecasts using ECE. Cells report group- and pair-level rates and improvement over both members.
}
\newsavebox{\apxECETableBox}
\begingroup
\begin{table}[H]
\centering
\setlength{\tabcolsep}{2.0pt}
\renewcommand{\arraystretch}{1.08}
\fontsize{8.0}{9.5}\selectfont
\captionsetup{font=normalsize}
\caption{Two-model ECE results at \(\groupThreshold=1{,}000\), retaining the BS-selected group benchmark, eligible pairs, and BS-trained coefficients.
}
\label{tab:overall-ece-brier-trained}
\sbox{\apxECETableBox}{%
\begin{tabular}{lccccc}
\toprule
& \multicolumn{2}{c}{Groups}
& \multicolumn{3}{c}{Eligible pairs} \\
\cmidrule(lr){2-3}\cmidrule(lr){4-6}
Aggregation rule & \shortstack{Match or improve\\benchmark}
& \shortstack{Within $5\%$\\of benchmark}
& \shortstack{Match or improve\\benchmark}
& \shortstack{Within $5\%$\\of benchmark}
& \shortstack{Improve on\\both members} \\
\midrule
\rowcolor{resultshade}
Learned linear pooling & \shortstack{16/16\\(100.00\%)} & \shortstack{16/16\\(100.00\%)} & \shortstack{772/1,121\\(68.87\%)} & \shortstack{826/1,121\\(73.68\%)} & \shortstack{705/1,121\\(62.89\%)} \\
\rowcolor{resultshade}
Learned log-odds pooling & \shortstack{13/16\\(81.25\%)} & \shortstack{13/16\\(81.25\%)} & \shortstack{291/1,121\\(25.96\%)} & \shortstack{336/1,121\\(29.97\%)} & \shortstack{577/1,121\\(51.47\%)} \\
\midrule
Simple mean (AM) & \shortstack{12/16\\(75.00\%)} & \shortstack{12/16\\(75.00\%)} & \shortstack{271/1,121\\(24.17\%)} & \shortstack{317/1,121\\(28.28\%)} & \shortstack{471/1,121\\(42.02\%)} \\
Harmonic mean (HM) & \shortstack{11/16\\(68.75\%)} & \shortstack{13/16\\(81.25\%)} & \shortstack{277/1,121\\(24.71\%)} & \shortstack{318/1,121\\(28.37\%)} & \shortstack{514/1,121\\(45.85\%)} \\
Log-odds mean (GM) & \shortstack{11/16\\(68.75\%)} & \shortstack{12/16\\(75.00\%)} & \shortstack{313/1,121\\(27.92\%)} & \shortstack{348/1,121\\(31.04\%)} & \shortstack{470/1,121\\(41.93\%)} \\
\midrule
$\alpha$-log-odds & \shortstack{9/16\\(56.25\%)} & \shortstack{10/16\\(62.50\%)} & \shortstack{296/1,121\\(26.40\%)} & \shortstack{336/1,121\\(29.97\%)} & \shortstack{420/1,121\\(37.47\%)} \\
Average prior (AP) & \shortstack{9/16\\(56.25\%)} & \shortstack{9/16\\(56.25\%)} & \shortstack{288/1,121\\(25.69\%)} & \shortstack{325/1,121\\(28.99\%)} & \shortstack{378/1,121\\(33.72\%)} \\
Heuristic prior (HP) & \shortstack{8/16\\(50.00\%)} & \shortstack{9/16\\(56.25\%)} & \shortstack{261/1,121\\(23.28\%)} & \shortstack{311/1,121\\(27.74\%)} & \shortstack{393/1,121\\(35.06\%)} \\
Precision & \shortstack{9/16\\(56.25\%)} & \shortstack{10/16\\(62.50\%)} & \shortstack{300/1,121\\(26.76\%)} & \shortstack{339/1,121\\(30.24\%)} & \shortstack{441/1,121\\(39.34\%)} \\
\bottomrule
\end{tabular}
}
\ifdim\wd\apxECETableBox>\linewidth
  \resizebox{\linewidth}{!}{\usebox{\apxECETableBox}}%
\else
  \resizebox{\wd\apxECETableBox}{!}{\usebox{\apxECETableBox}}%
\fi
\end{table}
\endgroup

\begin{figure}[htbp]
\centering
    \hspace*{22pt}
    \subfloat[Linear pooling.%
    \label{fig:reliability-linear}]{%
        \hspace*{-22pt}\resizebox{!}{5.5cm}{\begin{tikzpicture}
\begin{groupplot}[
forecast axes front,
group style={group size=1 by 2,vertical sep=0.85cm},
width=0.29\linewidth,
height=3.65cm,
xmin=0,
xmax=1,
xtick={0,0.5,1},
tick label style={font=\fontsize{4}{5}\selectfont},
label style={font=\scriptsize},
title style={font=\scriptsize},grid=major,grid style={black!10},scaled ticks=false,label style={font=\fontsize{4.5}{5.5}\selectfont},title style={font=\scriptsize}]

\nextgroupplot[
ymin=0,ymax=1,ytick={0,0.5,1},xticklabels=\empty,ylabel={Yes frequency},]

\addplot[black!50,dashed,mark=none,forget plot] coordinates {(0,0) (1,1)};
\draw[bslinear,line width=1.0pt] (axis cs:0.02270680,0.02270680) -- (axis cs:0.02270680,0.00854701);
\draw[bslinear,line width=1.0pt] (axis cs:0.14031779,0.14031779) -- (axis cs:0.14031779,0.18333333);
\draw[bslinear,line width=1.0pt] (axis cs:0.24368126,0.24368126) -- (axis cs:0.24368126,0.23529412);
\draw[bslinear,line width=1.0pt] (axis cs:0.35954847,0.35954847) -- (axis cs:0.35954847,0.39285714);
\draw[bslinear,line width=1.0pt] (axis cs:0.46116126,0.46116126) -- (axis cs:0.46116126,0.40441176);
\draw[bslinear,line width=1.0pt] (axis cs:0.54027356,0.54027356) -- (axis cs:0.54027356,0.58823529);
\draw[bslinear,line width=1.0pt] (axis cs:0.64003498,0.64003498) -- (axis cs:0.64003498,0.55882353);
\draw[bslinear,line width=1.0pt] (axis cs:0.75107500,0.75107500) -- (axis cs:0.75107500,0.64285714);
\draw[bslinear,line width=1.0pt] (axis cs:0.85286480,0.85286480) -- (axis cs:0.85286480,0.72413793);
\draw[bslinear,line width=1.0pt] (axis cs:0.96464751,0.96464751) -- (axis cs:0.96464751,0.97297297);

\addplot[
    bslinear,mark=square*,mark size=2.1pt,thick,
    mark options={solid, draw=white, line width=1.2pt, fill=bslinear},
] coordinates {(0.022707,0.00854701) (0.140318,0.18333333) (0.243681,0.23529412) (0.359548,0.39285714) (0.461161,0.40441176) (0.540274,0.58823529) (0.640035,0.55882353) (0.751075,0.64285714) (0.852865,0.72413793) (0.964648,0.97297297)};
\node[anchor=south east,fill=white,font=\fontsize{4}{5}\selectfont] at (rel axis cs:0.98,0.02) {ECE $=0.0370$};

\nextgroupplot[height=2.4cm,ymin=0,ymax=45,ytick={0,20,40},
ylabel={Forecasts (\%)},ybar,bar width=8pt]
\addplot[fill=bslinear,draw=none] coordinates {(0.050000,32.68156425) (0.150000,4.18994413) (0.250000,2.37430168) (0.350000,3.91061453) (0.450000,9.49720670) (0.550000,36.80167598) (0.650000,2.37430168) (0.750000,0.97765363) (0.850000,2.02513966) (0.950000,5.16759777)};

\end{groupplot}
\end{tikzpicture}}%
    }
    \hspace*{9pt}
    \subfloat[Log-odds pooling.%
    \label{fig:reliability-logodds}]{%
        \hspace*{-9pt}\resizebox{!}{5.5cm}{\begin{tikzpicture}
\begin{groupplot}[
forecast axes front,
group style={group size=1 by 2,vertical sep=0.85cm},
width = 0.3\linewidth,
height=3.65cm,
xmin=0,
xmax=1,
xtick={0,0.5,1},
tick label style={font=\fontsize{4}{5}\selectfont},
label style={font=\scriptsize},
title style={font=\scriptsize},
grid=major,
grid style={black!10},
scaled ticks=false,
label style={font=\fontsize{4.5}{5.5}\selectfont},
title style={font=\scriptsize}
]

\nextgroupplot[
ymin=0,ymax=1,ytick={0,0.5,1},xticklabels=\empty,]
\addplot[black!50,dashed,mark=none,forget plot] coordinates {(0,0) (1,1)};
\draw[bslogodds,line width=1.0pt] (axis cs:0.02916408,0.02916408) -- (axis cs:0.02916408,0.01094092);
\draw[bslogodds,line width=1.0pt] (axis cs:0.15343752,0.15343752) -- (axis cs:0.15343752,0.17021277);
\draw[bslogodds,line width=1.0pt] (axis cs:0.24875956,0.24875956) -- (axis cs:0.24875956,0.16666667);
\draw[bslogodds,line width=1.0pt] (axis cs:0.34477212,0.34477212) -- (axis cs:0.34477212,0.22500000);
\draw[bslogodds,line width=1.0pt] (axis cs:0.46863618,0.46863618) -- (axis cs:0.46863618,0.39880952);
\draw[bslogodds,line width=1.0pt] (axis cs:0.53056223,0.53056223) -- (axis cs:0.53056223,0.59474672);
\draw[bslogodds,line width=1.0pt] (axis cs:0.63683061,0.63683061) -- (axis cs:0.63683061,0.55555556);
\draw[bslogodds,line width=1.0pt] (axis cs:0.75195379,0.75195379) -- (axis cs:0.75195379,0.73333333);
\draw[bslogodds,line width=1.0pt] (axis cs:0.85541160,0.85541160) -- (axis cs:0.85541160,0.84615385);
\draw[bslogodds,line width=1.0pt] (axis cs:0.95683577,0.95683577) -- (axis cs:0.95683577,0.97183099);
\addplot[
    bslogodds,mark=diamond*,mark size=2.1pt,thick,
    mark options={solid, draw=white, line width=0.6pt, fill=bslogodds},
] 
coordinates {(0.029164,0.01094092) (0.153438,0.17021277) (0.248760,0.16666667) (0.344772,0.22500000) (0.468636,0.39880952) (0.530562,0.59474672) (0.636831,0.55555556) (0.751954,0.73333333) (0.855412,0.84615385) (0.956836,0.97183099)};
\node[anchor=south east,fill=white,font=\fontsize{4}{5}\selectfont] at (rel axis cs:0.98,0.02) {ECE $=0.0472$};

\nextgroupplot[height=2.4cm,ymin=0,ymax=45,ytick={0,20,40},
ybar,bar width=8pt]
\addplot[fill=bslogodds,draw=none] coordinates {(0.050000,31.91340782) (0.150000,3.28212291) (0.250000,3.35195531) (0.350000,2.79329609) (0.450000,11.73184358) (0.550000,37.22067039) (0.650000,1.88547486) (0.750000,1.04748603) (0.850000,1.81564246) (0.950000,4.95810056)};

\end{groupplot}
\end{tikzpicture}}%
    }
    \hspace*{9pt}
    \subfloat[Group benchmark.%
    \label{fig:reliability-benchmark}]{%
        \hspace*{-9pt}\resizebox{!}{5.5cm}{\begin{tikzpicture}
\begin{groupplot}[
forecast axes front,
group style={group size=1 by 2,vertical sep=0.85cm},
width=0.3\linewidth,
height=3.65cm,
xmin=0,
xmax=1,
xtick={0,0.5,1},
tick label style={font=\fontsize{4}{5}\selectfont},
label style={font=\scriptsize},
title style={font=\scriptsize},grid=major,grid style={black!10},scaled ticks=false,label style={font=\fontsize{4.5}{5.5}\selectfont},title style={font=\scriptsize}
]

\nextgroupplot[
ymin=0,ymax=1,ytick={0,0.5,1},xticklabels=\empty,]
\addplot[black!50,dashed,mark=none,forget plot] coordinates {(0,0) (1,1)};
\draw[bsindividual,line width=1.0pt] (axis cs:0.01054479,0.01054479) -- (axis cs:0.01054479,0.01679104);
\draw[bsindividual,line width=1.0pt] (axis cs:0.13538235,0.13538235) -- (axis cs:0.13538235,0.11764706);
\draw[bsindividual,line width=1.0pt] (axis cs:0.24788462,0.24788462) -- (axis cs:0.24788462,0.34615385);
\draw[bsindividual,line width=1.0pt] (axis cs:0.35190476,0.35190476) -- (axis cs:0.35190476,0.45238095);
\draw[bsindividual,line width=1.0pt] (axis cs:0.46051948,0.46051948) -- (axis cs:0.46051948,0.55194805);
\draw[bsindividual,line width=1.0pt] (axis cs:0.51982582,0.51982582) -- (axis cs:0.51982582,0.58196721);
\draw[bsindividual,line width=1.0pt] (axis cs:0.62631579,0.62631579) -- (axis cs:0.62631579,0.44736842);
\draw[bsindividual,line width=1.0pt] (axis cs:0.74142857,0.74142857) -- (axis cs:0.74142857,0.71428571);
\draw[bsindividual,line width=1.0pt] (axis cs:0.84875000,0.84875000) -- (axis cs:0.84875000,0.81250000);
\draw[bsindividual,line width=1.0pt] (axis cs:0.98268831,0.98268831) -- (axis cs:0.98268831,0.98701299);

\addplot[
    bsindividual,mark=*,mark size=2.1pt,thick,
    mark options={solid, draw=white, line width=1.2pt, fill=bsindividual},
] coordinates {(0.010545,0.01679104) (0.135382,0.11764706) (0.247885,0.34615385) (0.351905,0.45238095) (0.460519,0.55194805) (0.519826,0.58196721) (0.626316,0.44736842) (0.741429,0.71428571) (0.848750,0.81250000) (0.982688,0.98701299)};
\node[anchor=south east,fill=white,font=\fontsize{4}{5}\selectfont] at (rel axis cs:0.98,0.02) {ECE $=0.0443$};

\nextgroupplot[height=2.4cm,ymin=0,ymax=45,ytick={0,20,40},
ybar,bar width=8pt]
\addplot[fill=bsindividual,draw=none] coordinates {(0.050000,37.43016760) (0.150000,2.37430168) (0.250000,1.81564246) (0.350000,2.93296089) (0.450000,10.75418994) (0.550000,34.07821229) (0.650000,2.65363128) (0.750000,1.46648045) (0.850000,1.11731844) (0.950000,5.37709497)};

\end{groupplot}
\end{tikzpicture}}%
    }

\caption{Reliability and forecast distributions for the rule-specific eligible pairs and group benchmark in G1.
}
\label{fig:overall-reliability-1000}
\end{figure}
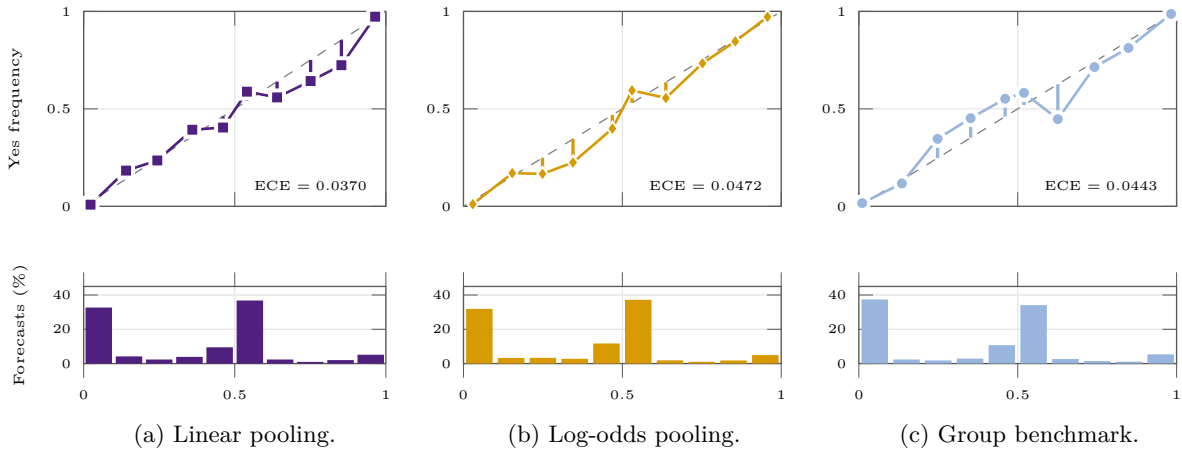

\flushbottom
\rxedit{
Below is \Cref{fig:overall-reliability-1000}\aiedit{, showing} the additional reliability diagrams for a representative comparison group G1 of \Cref{sec:main-results}.
The linear pair is Claude-Opus-4-1 with GPT-5-Mini; the log-odds pair is Claude-3-7-Sonnet with GPT-5-Mini; the group benchmark
is Gemini-3-Pro-Preview. We select the pairs by BS.
The top panels use ten equal-width probability bins.
The dashed diagonal indicates perfect calibration, and vertical segments show the gap between each bin's mean forecast and its Yes frequency. 
Empty bins have no marker.
The bottom panels show the percentage of test forecasts in each bin. The three panels evaluate the same 533 test subquestions in G1.
}

\section{Case Study: When Learned Linear Pooling Succeeds}
\label{sec:case-study}
\begin{figure}[h]
\centering
\captionsetup[subfloat]{font=footnotesize,justification=centering}

\newsavebox{\caseStudyTopPanelA}
\newsavebox{\caseStudyTopPanelB}
\newlength{\caseStudyTopRowHeight}
\newlength{\caseStudyTopRowDepth}

\pgfmathsetlengthmacro{\caseStudyPlotWidth}{0.37*\linewidth}
\pgfmathsetlengthmacro{\caseStudyPlotHeight}{0.28*\linewidth}
\def\caseStudyPanelAScale{0.6}
\pgfmathsetlengthmacro{\caseStudyIndividualWidth}{\caseStudyPlotWidth/\caseStudyPanelAScale}
\pgfmathsetlengthmacro{\caseStudyIndividualHeight}{\caseStudyPlotHeight/\caseStudyPanelAScale}
\pgfmathsetlengthmacro{\caseStudyHeatmapXUnit}{\caseStudyPlotWidth/7}
\pgfmathsetlengthmacro{\caseStudyHeatmapYUnit}{\caseStudyPlotHeight/7}

\sbox{\caseStudyTopPanelA}{%
    \scalebox{\caseStudyPanelAScale}{%
\begin{tikzpicture}[baseline=(current axis.south)]
\begin{axis}[
forecast axes front,
scale only axis,
width=\caseStudyIndividualWidth,height=\caseStudyIndividualHeight,
xmin=0.45,xmax=9.55,ymin=0.740,ymax=0.842,
xtick={1,...,9},
xticklabels={{Claude 3 Opus},{Mistral Large},{Claude 3.5 Sonnet},{Llama 3 70B},{GPT-4 Turbo},{GPT-4o},{Qwen3 235B},{Llama 3 8B},{Claude 3 Haiku}},
x tick label style={font=\footnotesize,rotate=38,anchor=east},
ytick={0.74,0.79,0.84},
yticklabel style={
/pgf/number format/fixed zerofill,
/pgf/number format/precision=2
},
ylabel={1 $-$ Test Brier score
},
tick label style={font=\small},label style={font=\small},
scaled ticks=false,ymajorgrids=true,
grid style={black!10, /pgfplots/on layer=axis grid},tick align=outside,
bar width=17pt,
]
\addplot[forget plot,ybar,bar shift=0pt,
fill=a-orange,
draw=a-orange!90!black,
mark=none,
error bars/.cd,y dir=both,y explicit,error mark=-,error bar style={gray,line width=0.8pt},error mark options={mark size=0.40pt,line width=5.60pt}
] 
coordinates {(1,0.820252304383) += (0,0.012052556863) -= (0,0.012052556863)};

\addplot[forget plot,ybar,bar shift=0pt,
fill=mistral-orange,
draw=mistral-orange!90!black,
mark=none,
error bars/.cd,y dir=both,y explicit,error mark=-,error bar style={gray,line width=0.8pt},error mark options={mark size=0.40pt,line width=5.60pt}
] 
coordinates {(2,0.812662067816) += (0,0.011798290813) -= (0,0.011798290813)};

\addplot[forget plot,ybar,bar shift=0pt,
fill=a-orange,
draw=a-orange!90!black,
mark=none,
error bars/.cd,y dir=both,y explicit,error mark=-,error bar style={gray,line width=0.8pt},error mark options={mark size=0.40pt,line width=5.60pt}
] 
coordinates {(3,0.811048166002) += (0,0.013469298002) -= (0,0.013469298002)};

\addplot[forget plot,ybar,bar shift=0pt,
fill=meta-blue,
draw=meta-blue!90!black,
mark=none,
error bars/.cd,y dir=both,y explicit,error mark=-,error bar style={gray,line width=0.8pt},error mark options={mark size=0.40pt,line width=5.60pt}
]
coordinates {(4,0.795640113308) += (0,0.012715805107) -= (0,0.012715805108)};

\addplot[forget plot,ybar,bar shift=0pt,
fill=o-green,
draw=o-green!90!black,
mark=none,
error bars/.cd,y dir=both,y explicit,error mark=-,error bar style={gray,line width=0.8pt},error mark options={mark size=0.40pt,line width=5.60pt}
] 
coordinates {(5,0.794901161350) += (0,0.014928948408) -= (0,0.014928948407)};

\addplot[forget plot,ybar,bar shift=0pt,
fill=o-green,
draw=o-green!90!black,
mark=none,
error bars/.cd,y dir=both,y explicit,error mark=-,error bar style={gray,line width=0.8pt},error mark options={mark size=0.40pt,line width=5.60pt}
] 
coordinates {(6,0.784374110976) += (0,0.014380216411) -= (0,0.014380216412)};

\addplot[forget plot,ybar,bar shift=0pt,
fill=ali-purple,
draw=ali-purple!90!black,
mark=none,
error bars/.cd,y dir=both,y explicit,error mark=-,error bar style={gray,line width=0.8pt},error mark options={mark size=0.40pt,line width=5.60pt}
] 
coordinates {(7,0.771209444999) += (0,0.013934066662) -= (0,0.013934066661)};

\addplot[forget plot,ybar,bar shift=0pt,
fill=meta-blue,
draw=meta-blue!90!black,
mark=none,
error bars/.cd,y dir=both,y explicit,error mark=-,error bar style={gray,line width=0.8pt},error mark options={mark size=0.40pt,line width=5.60pt}
]
coordinates {(8,0.767982990508) += (0,0.015051782462) -= (0,0.015051782462)};

\addplot[forget plot,ybar,bar shift=0pt,
fill=a-orange,
draw=a-orange!90!black,
mark=none,
error bars/.cd,y dir=both,y explicit,error mark=-,error bar style={gray,line width=0.8pt},error mark options={mark size=0.40pt,line width=5.60pt}
]
coordinates {(9,0.764232913268) += (0,0.011208360374) -= (0,0.011208360373)};

\end{axis}

\end{tikzpicture}}%
}
\sbox{\caseStudyTopPanelB}{%
\begin{tikzpicture}[x=\caseStudyHeatmapXUnit,y=\caseStudyHeatmapYUnit,baseline=0pt]
\definecolor{g13gainpurple}{RGB}{83,39,131}
\definecolor{g13gaingold}{RGB}{240,174,15}
\definecolor{g13gainwhite}{RGB}{250,249,247}
\definecolor{g13gainr9c2}{RGB}{242,201,99}
\definecolor{g13gainr9c3}{RGB}{239,182,36}
\definecolor{g13gainr9c4}{RGB}{244,220,162}
\definecolor{g13gainr9c5}{RGB}{246,230,194}
\definecolor{g13gainr9c6}{RGB}{244,226,179}
\definecolor{g13gainr9c7}{RGB}{246,236,211}
\definecolor{g13gainr9c8}{RGB}{117,80,152}
\definecolor{g13gainr8c2}{RGB}{242,201,99}
\definecolor{g13gainr8c3}{RGB}{239,182,36}
\definecolor{g13gainr8c4}{RGB}{244,221,164}
\definecolor{g13gainr8c5}{RGB}{246,236,213}
\definecolor{g13gainr8c6}{RGB}{244,229,190}
\definecolor{g13gainr8c7}{RGB}{246,238,220}
\definecolor{g13gainr7c2}{RGB}{242,200,95}
\definecolor{g13gainr7c3}{RGB}{239,182,36}
\definecolor{g13gainr7c4}{RGB}{244,219,153}
\definecolor{g13gainr7c5}{RGB}{244,224,173}
\definecolor{g13gainr7c6}{RGB}{244,225,178}
\definecolor{g13gainr6c2}{RGB}{242,198,89}
\definecolor{g13gainr6c3}{RGB}{239,182,36}
\definecolor{g13gainr6c4}{RGB}{243,213,140}
\definecolor{g13gainr6c5}{RGB}{245,222,168}
\definecolor{g13gainr5c2}{RGB}{242,201,99}
\definecolor{g13gainr5c3}{RGB}{240,180,30}
\definecolor{g13gainr5c4}{RGB}{244,220,160}
\definecolor{g13gainr4c2}{RGB}{242,200,97}
\definecolor{g13gainr4c3}{RGB}{240,180,32}
\definecolor{g13gainr3c2}{RGB}{240,174,15}
\fill[g13gainr9c2] (0.03,6.03) rectangle (0.97,6.97);
\fill[g13gainr9c3] (1.02,6.03) rectangle (1.98,6.97);
\fill[g13gainr9c4] (2.02,6.03) rectangle (2.98,6.97);
\fill[g13gainr9c5] (3.02,6.03) rectangle (3.98,6.97);
\fill[g13gainr9c6] (4.03,6.03) rectangle (4.97,6.97);
\fill[g13gainr9c7] (5.03,6.03) rectangle (5.97,6.97);
\fill[g13gainr9c8] (6.03,6.03) rectangle (6.97,6.97);
\fill[g13gainr8c2] (0.03,5.03) rectangle (0.97,5.97);
\fill[g13gainr8c3] (1.02,5.03) rectangle (1.98,5.97);
\fill[g13gainr8c4] (2.02,5.03) rectangle (2.98,5.97);
\fill[g13gainr8c5] (3.02,5.03) rectangle (3.98,5.97);
\fill[g13gainr8c6] (4.03,5.03) rectangle (4.97,5.97);
\fill[g13gainr8c7] (5.03,5.03) rectangle (5.97,5.97);
\fill[g13gainr7c2] (0.03,4.03) rectangle (0.97,4.97);
\fill[g13gainr7c3] (1.02,4.03) rectangle (1.98,4.97);
\fill[g13gainr7c4] (2.02,4.03) rectangle (2.98,4.97);
\fill[g13gainr7c5] (3.02,4.03) rectangle (3.98,4.97);
\fill[g13gainr7c6] (4.03,4.03) rectangle (4.97,4.97);
\fill[g13gainr6c2] (0.03,3.02) rectangle (0.97,3.98);
\fill[g13gainr6c3] (1.02,3.02) rectangle (1.98,3.98);
\fill[g13gainr6c4] (2.02,3.02) rectangle (2.98,3.98);
\fill[g13gainr6c5] (3.02,3.02) rectangle (3.98,3.98);
\fill[g13gainr5c2] (0.03,2.02) rectangle (0.97,2.98);
\fill[g13gainr5c3] (1.02,2.02) rectangle (1.98,2.98);
\fill[g13gainr5c4] (2.02,2.02) rectangle (2.98,2.98);
\fill[g13gainr4c2] (0.03,1.02) rectangle (0.97,1.98);
\fill[g13gainr4c3] (1.02,1.02) rectangle (1.98,1.98);
\fill[g13gainr3c2] (0.03,0.03) rectangle (0.97,0.97);
\node[font=\fontsize{6}{7}\selectfont,text=white] at (0.5,6.5) {+0.094};
\node[font=\fontsize{6}{7}\selectfont,text=white] at (1.5,6.5) {+0.135};
\node[font=\fontsize{6}{7}\selectfont,text=black!75] at (2.5,6.5) {+0.052};
\node[font=\fontsize{6}{7}\selectfont,text=black!75] at (3.5,6.5) {+0.030};
\node[font=\fontsize{6}{7}\selectfont,text=black!75] at (4.5,6.5) {+0.040};
\node[font=\fontsize{6}{7}\selectfont,text=black!75] at (5.5,6.5) {+0.020};
\node[font=\fontsize{6}{7}\selectfont,text=white] at (6.5,6.5) {-0.019};
\node[font=\fontsize{6}{7}\selectfont,text=white] at (0.5,5.5) {+0.094};
\node[font=\fontsize{6}{7}\selectfont,text=white] at (1.5,5.5) {+0.135};
\node[font=\fontsize{6}{7}\selectfont,text=black!75] at (2.5,5.5) {+0.050};
\node[font=\fontsize{6}{7}\selectfont,text=black!75] at (3.5,5.5) {+0.018};
\node[font=\fontsize{6}{7}\selectfont,text=black!75] at (4.5,5.5) {+0.033};
\node[font=\fontsize{6}{7}\selectfont,text=black!75] at (5.5,5.5) {+0.013};
\node[font=\fontsize{6}{7}\selectfont,text=white] at (0.5,4.5) {+0.097};
\node[font=\fontsize{6}{7}\selectfont,text=white] at (1.5,4.5) {+0.135};
\node[font=\fontsize{6}{7}\selectfont,text=black!75] at (2.5,4.5) {+0.058};
\node[font=\fontsize{6}{7}\selectfont,text=black!75] at (3.5,4.5) {+0.044};
\node[font=\fontsize{6}{7}\selectfont,text=black!75] at (4.5,4.5) {+0.042};
\node[font=\fontsize{6}{7}\selectfont,text=white] at (0.5,3.5) {+0.100};
\node[font=\fontsize{6}{7}\selectfont,text=white] at (1.5,3.5) {+0.135};
\node[font=\fontsize{6}{7}\selectfont,text=black!75] at (2.5,3.5) {+0.066};
\node[font=\fontsize{6}{7}\selectfont,text=black!75] at (3.5,3.5) {+0.048};
\node[font=\fontsize{6}{7}\selectfont,text=white] at (0.5,2.5) {+0.095};
\node[font=\fontsize{6}{7}\selectfont,text=white] at (1.5,2.5) {+0.139};
\node[font=\fontsize{6}{7}\selectfont,text=black!75] at (2.5,2.5) {+0.053};
\node[font=\fontsize{6}{7}\selectfont,text=white] at (0.5,1.5) {+0.096};
\node[font=\fontsize{6}{7}\selectfont,text=white] at (1.5,1.5) {+0.138};
\node[font=\fontsize{6}{7}\selectfont,text=white] at (0.5,0.5) {+0.152};
\draw[black!65,line width=0.35pt] (0,0) -- (7,0);
\draw[black!65,line width=0.35pt] (0,0) -- (0,7);
\draw[black!65,line width=0.35pt] (0.5,0) -- (0.5,-0.09);
\node[font=\fontsize{6}{7}\selectfont,anchor=north] at (0.5,-0.13) {2};
\draw[black!65,line width=0.35pt] (1.5,0) -- (1.5,-0.09);
\node[font=\fontsize{6}{7}\selectfont,anchor=north] at (1.5,-0.13) {3};
\draw[black!65,line width=0.35pt] (2.5,0) -- (2.5,-0.09);
\node[font=\fontsize{6}{7}\selectfont,anchor=north] at (2.5,-0.13) {4};
\draw[black!65,line width=0.35pt] (3.5,0) -- (3.5,-0.09);
\node[font=\fontsize{6}{7}\selectfont,anchor=north] at (3.5,-0.13) {5};
\draw[black!65,line width=0.35pt] (4.5,0) -- (4.5,-0.09);
\node[font=\fontsize{6}{7}\selectfont,anchor=north] at (4.5,-0.13) {6};
\draw[black!65,line width=0.35pt] (5.5,0) -- (5.5,-0.09);
\node[font=\fontsize{6}{7}\selectfont,anchor=north] at (5.5,-0.13) {7};
\draw[black!65,line width=0.35pt] (6.5,0) -- (6.5,-0.09);
\node[font=\fontsize{6}{7}\selectfont,anchor=north] at (6.5,-0.13) {8};
\draw[black!65,line width=0.35pt] (0,0.5) -- (-0.09,0.5);
\node[font=\fontsize{6}{7}\selectfont,anchor=east] at (-0.15,0.5) {3};
\draw[black!65,line width=0.35pt] (0,1.5) -- (-0.09,1.5);
\node[font=\fontsize{6}{7}\selectfont,anchor=east] at (-0.15,1.5) {4};
\draw[black!65,line width=0.35pt] (0,2.5) -- (-0.09,2.5);
\node[font=\fontsize{6}{7}\selectfont,anchor=east] at (-0.15,2.5) {5};
\draw[black!65,line width=0.35pt] (0,3.5) -- (-0.09,3.5);
\node[font=\fontsize{6}{7}\selectfont,anchor=east] at (-0.15,3.5) {6};
\draw[black!65,line width=0.35pt] (0,4.5) -- (-0.09,4.5);
\node[font=\fontsize{6}{7}\selectfont,anchor=east] at (-0.15,4.5) {7};
\draw[black!65,line width=0.35pt] (0,5.5) -- (-0.09,5.5);
\node[font=\fontsize{6}{7}\selectfont,anchor=east] at (-0.15,5.5) {8};
\draw[black!65,line width=0.35pt] (0,6.5) -- (-0.09,6.5);
\node[font=\fontsize{6}{7}\selectfont,anchor=east] at (-0.15,6.5) {9};

\shade[bottom color=g13gainpurple,top color=g13gainwhite] (7.28,0.75) rectangle (7.56,3.50);
\shade[bottom color=g13gainwhite,top color=g13gaingold] (7.28,3.50) rectangle (7.56,6.25);
\draw[black!65,line width=0.35pt] (7.56,1.67) -- (7.66,1.67);
\node[font=\fontsize{5.5}{6.5}\selectfont,anchor=west]
    at (7.72,1.67) {-0.02};

\draw[black!65,line width=0.35pt] (7.56,3.50) -- (7.66,3.50);
\node[font=\fontsize{5.5}{6.5}\selectfont,anchor=west]
    at (7.72,3.50) {0.00};

\draw[black!65,line width=0.35pt] (7.56,6.25) -- (7.66,6.25);
\node[font=\fontsize{5.5}{6.5}\selectfont,anchor=west]
    at (7.72,6.25) {0.16};
\node[font=\fontsize{6}{7}\selectfont,rotate=90] at (8.55,3.5) {Fractional BS gain};
\end{tikzpicture}%
}

\setlength{\caseStudyTopRowHeight}{\ht\caseStudyTopPanelA}
\ifdim\ht\caseStudyTopPanelB>\caseStudyTopRowHeight
    \setlength{\caseStudyTopRowHeight}{\ht\caseStudyTopPanelB}
\fi
\setlength{\caseStudyTopRowDepth}{\dp\caseStudyTopPanelA}
\ifdim\dp\caseStudyTopPanelB>\caseStudyTopRowDepth
    \setlength{\caseStudyTopRowDepth}{\dp\caseStudyTopPanelB}
\fi

\subfloat[Individual forecasts \rxedit{1 - test BS}.%
\label{fig:sec4-case-G13-a}]{%
    \makebox[0.46\linewidth][c]{%
        \raisebox{0pt}[\caseStudyTopRowHeight][\caseStudyTopRowDepth]{%
            \usebox{\caseStudyTopPanelA}}%
    }%
}
\hfill%
\subfloat[Learned-linear BS gain by pair ranks.%
\label{fig:sec4-case-G13-b}]{%
    \makebox[0.48\linewidth][c]{%
        \raisebox{0pt}[\caseStudyTopRowHeight][\caseStudyTopRowDepth]{%
            \usebox{\caseStudyTopPanelB}}%
    }%
}

\caption{
Individual and pairwise forecasting performance in Group G13. 
(a) Individual forecasters\aiedit{'} performance. 
(b) Fractional test-BS reduction of learned-linear pairs relative to the benchmark, indexed by the ranks of their two members; positive values favor the pair. 
}
\label{fig:sec4-case-G13}
\end{figure}
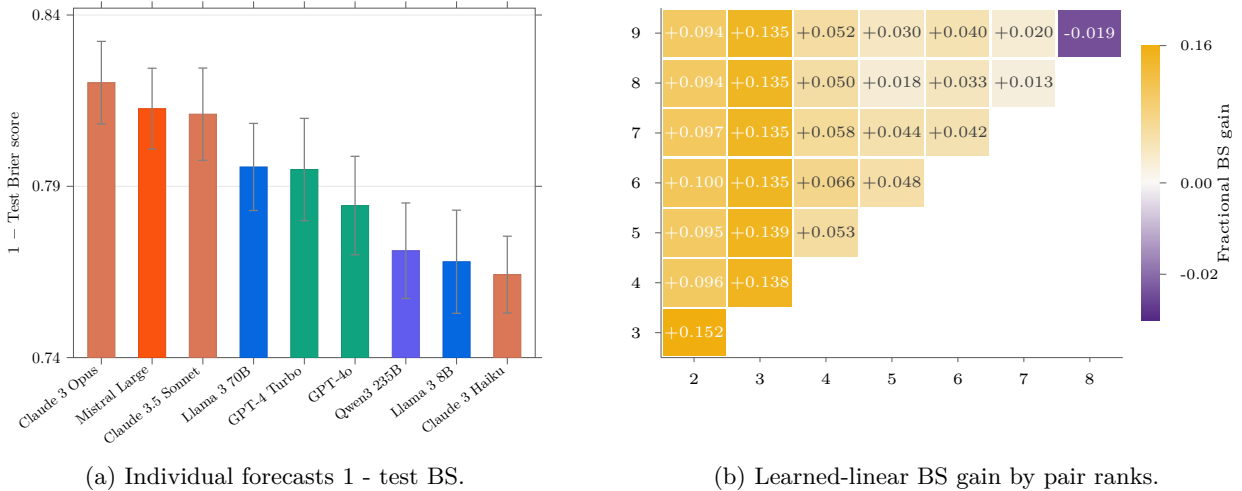

In this section, under our pairwise aggregation results, we zoom into Group G13 to understand why the choice of pooling rule can matter even when the constituent forecasters are held fixed. 
This group contains 9 forecasters with 1,557 shared subquestions, of which 779 belong to the test set. 
In \Cref{fig:sec4-case-G13-a}, we report the 1 $-$ BS of each individual forecaster, where we can see that the group benchmark is Claude 3 Opus. 
The remaining eight forecasters form 28 eligible pairs. 
In \Cref{fig:sec4-case-G13-b}, we report their aggregated performance.
As we can see,  linear pooling succeeds broadly across pair compositions. 
Learned linear pooling matches or improves on the group benchmark for 27 of the 28 pairs (96.43\%), compared with only 2 pairs (7.14\%) under learned log-odds pooling. 
\Cref{fig:sec4-case-G13-a} shows the group benchmark has test \(1-\mathrm{BS}\) \aiedit{of} $0.820$. 
\Cref{fig:sec4-case-G13-b} further shows that the linear-pooling gains are not concentrated among pairs containing the strongest of these weaker forecasters. 
The improved forecast performance appear\aiedit{s} across almost the entire pair-rank matrix, including pairs formed by relatively low-ranked individuals; only one pair has a negative gain. 
Thus, the strong performance of linear pooling is not driven by a small number of favorable pair compositions.

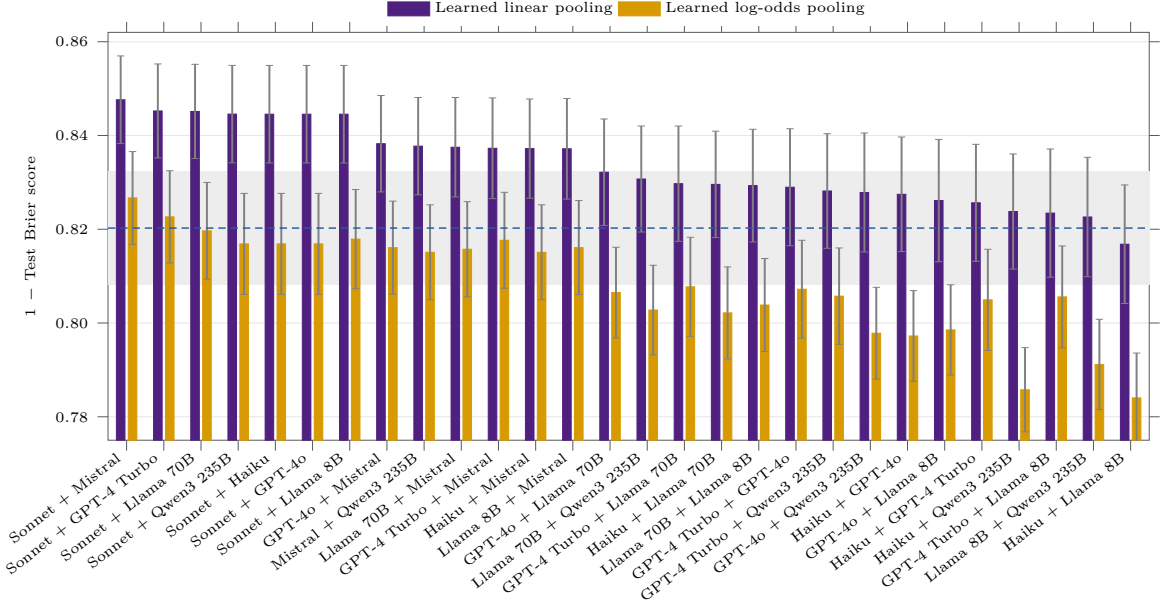
\begin{figure}[h]
    \centering
    \resizebox{0.95\linewidth}{!}{%
\begin{tikzpicture}
\begin{axis}[
forecast axes front,
width=\linewidth,height=0.45\linewidth,
xmin=0.5,xmax=28.5,ymin=0.775,ymax=0.862,
grid style={black!10, /pgfplots/on layer=axis grid},
xtick={1,...,28},
xticklabels={{Sonnet + Mistral},{Sonnet + GPT-4 Turbo},{Sonnet + Llama 70B},{Sonnet + Qwen3 235B},{Sonnet + Haiku},{Sonnet + GPT-4o},{Sonnet + Llama 8B},{GPT-4o + Mistral},{Mistral + Qwen3 235B},{Llama 70B + Mistral},{GPT-4 Turbo + Mistral},{Haiku + Mistral},{Llama 8B + Mistral},{GPT-4o + Llama 70B},{Llama 70B + Qwen3 235B},{GPT-4 Turbo + Llama 70B},{Haiku + Llama 70B},{Llama 70B + Llama 8B},{GPT-4 Turbo + GPT-4o},{GPT-4 Turbo + Qwen3 235B},{GPT-4o + Qwen3 235B},{Haiku + GPT-4o},{GPT-4o + Llama 8B},{Haiku + GPT-4 Turbo},{Haiku + Qwen3 235B},{GPT-4 Turbo + Llama 8B},{Llama 8B + Qwen3 235B},{Haiku + Llama 8B}},
x tick label style={font=\tiny,rotate=38,anchor=east},
ytick={0.78,0.80,0.82,0.84,0.86},
yticklabel style={
/pgf/number format/fixed zerofill,
/pgf/number format/precision=2
},
ylabel={1 $-$ Test Brier score},
tick label style={font=\fontsize{6}{7}\selectfont},label style={font=\fontsize{6}{7}\selectfont},
scaled ticks=false,ymajorgrids=true,
tick align=outside,
legend style={at={(0.5,1.015)},anchor=south,draw=none,font=\fontsize{6}{7}\selectfont,legend columns=2},
bar width=3.5pt,
]
\path[fill=black!7,draw=none] (axis cs:0.5,0.808199747520) rectangle (axis cs:28.5,0.832304861246);

\addplot[forget plot,ybar,bar shift=0pt,fill=bslinear,draw=bslinear,mark=none,
error bars/.cd,y dir=both,y explicit,error mark=-,error bar style={gray,line width=0.8pt},error mark options={mark size=0.20pt,line width=3pt}
] 
coordinates {(0.84,0.847623482313) += (0,0.009353720096) -= (0,0.009353720096) (1.84,0.845236366608) += (0,0.010029351572) -= (0,0.010029351571) (2.84,0.845133023079) += (0,0.010056364465) -= (0,0.010056364465) (3.84,0.844564570337) += (0,0.010394579423) -= (0,0.010394579424) (4.84,0.844543704820) += (0,0.010403851663) -= (0,0.010403851662) (5.84,0.844543704820) += (0,0.010403851663) -= (0,0.010403851662) (6.84,0.844543704820) += (0,0.010403851663) -= (0,0.010403851662) (7.84,0.838249098750) += (0,0.010287816673) -= (0,0.010287816673) (8.84,0.837729834012) += (0,0.010385978520) -= (0,0.010385978519) (9.84,0.837494611382) += (0,0.010606282974) -= (0,0.010606282973) (10.84,0.837272464347) += (0,0.010742847981) -= (0,0.010742847981) (11.84,0.837200737410) += (0,0.010585308958) -= (0,0.010585308958) (12.84,0.837171081576) += (0,0.010733695152) -= (0,0.010733695152) (13.84,0.832174338128) += (0,0.011351300170) -= (0,0.011351300171) (14.84,0.830720381909) += (0,0.011303795736) -= (0,0.011303795736) (15.84,0.829719190514) += (0,0.012295834606) -= (0,0.012295834607) (16.84,0.829573104543) += (0,0.011353616821) -= (0,0.011353616821) (17.84,0.829311022069) += (0,0.012021748635) -= (0,0.012021748634) (18.84,0.828966010655) += (0,0.012480340211) -= (0,0.012480340210) (19.84,0.828164747037) += (0,0.012221403374) -= (0,0.012221403374) (20.84,0.827850682165) += (0,0.012686455345) -= (0,0.012686455346) (21.84,0.827455125123) += (0,0.012222280134) -= (0,0.012222280134) (22.84,0.826118878956) += (0,0.013039671102) -= (0,0.013039671101) (23.84,0.825647312210) += (0,0.012474104630) -= (0,0.012474104630) (24.84,0.823787662566) += (0,0.012283232869) -= (0,0.012283232869) (25.84,0.823447560464) += (0,0.013682556874) -= (0,0.013682556874) (26.84,0.822619673907) += (0,0.012734504863) -= (0,0.012734504863) (27.84,0.816811364726) += (0,0.012646640484) -= (0,0.012646640484)};
\addlegendimage{area legend,fill=bslinear,draw=bslinear}
\addlegendentry{Learned linear pooling}
\addplot[forget plot,ybar,bar shift=0pt,fill=bslogodds,draw=bslogodds,mark=none,
error bars/.cd,y dir=both,y explicit,error mark=-,error bar style={gray,line width=0.8pt},error mark options={mark size=0.20pt,line width=3pt}
] 
coordinates {(1.16,0.826680186987) += (0,0.009891824181) -= (0,0.009891824182) (2.16,0.822651135720) += (0,0.009831734315) -= (0,0.009831734315) (3.16,0.819674341574) += (0,0.010319881705) -= (0,0.010319881705) (4.16,0.816879374987) += (0,0.010773617074) -= (0,0.010773617073) (5.16,0.816879374987) += (0,0.010773617074) -= (0,0.010773617073) (6.16,0.816879374987) += (0,0.010773617074) -= (0,0.010773617073) (7.16,0.817902435679) += (0,0.010583712396) -= (0,0.010583712396) (8.16,0.816076482455) += (0,0.009929014528) -= (0,0.009929014528) (9.16,0.815091324189) += (0,0.010121233654) -= (0,0.010121233655) (10.16,0.815736865859) += (0,0.010145294133) -= (0,0.010145294133) (11.16,0.817647379032) += (0,0.010237180733) -= (0,0.010237180732) (12.16,0.815091324193) += (0,0.010121233656) -= (0,0.010121233656) (13.16,0.816090720050) += (0,0.010033206241) -= (0,0.010033206240) (14.16,0.806493735256) += (0,0.009668343305) -= (0,0.009668343304) (15.16,0.802772233814) += (0,0.009553396507) -= (0,0.009553396507) (16.16,0.807721779125) += (0,0.010573756414) -= (0,0.010573756414) (17.16,0.802164994900) += (0,0.009809271605) -= (0,0.009809271605) (18.16,0.803837227234) += (0,0.009923298645) -= (0,0.009923298646) (19.16,0.807208875919) += (0,0.010447356630) -= (0,0.010447356631) (20.16,0.805737289056) += (0,0.010273513008) -= (0,0.010273513007) (21.16,0.797824577986) += (0,0.009799224670) -= (0,0.009799224669) (22.16,0.797236554075) += (0,0.009678027119) -= (0,0.009678027120) (23.16,0.798527546010) += (0,0.009621701441) -= (0,0.009621701441) (24.16,0.804949946672) += (0,0.010784165655) -= (0,0.010784165655) (25.16,0.785764721128) += (0,0.008984050528) -= (0,0.008984050528) (26.16,0.805577660522) += (0,0.010869130037) -= (0,0.010869130037) (27.16,0.791163980294) += (0,0.009634730368) -= (0,0.009634730367) (28.16,0.784034069241) += (0,0.009564697952) -= (0,0.009564697951)};
\addlegendimage{area legend,fill=bslogodds,draw=bslogodds}
\addlegendentry{Learned log-odds pooling}
\addplot[forget plot,forecastblue!85!black,densely dashed,semithick] coordinates {(0.5,0.820252304383) (28.5,0.820252304383)};
\end{axis}
\end{tikzpicture}
    }%
    \caption{
    Learned linear and log-odds pooling on the same 28 eligible pairs of this section. The dashed line and gray band are the benchmark $1$ - Test BS and its 95\% CI, respectively.}
    \label{fig:sec4-case-G13-c}
\end{figure}

In addition, we also observe that the pooling rule matters even for the same pair. 
\Cref{fig:sec4-case-G13-c} compares learned linear and log-odds pooling on the same 28 pairs. We observe higher test \(1-\mathrm{BS}\) under linear pooling in all 28 matched comparisons, with differences ranging from 0.0179 to 0.038. 
The best observed linear aggregate reaches \(1-\mathrm{BS}=0.848\), compared with 0.827 under learned log-odds pooling. 
These matched comparisons show that the aggregation rule itself can substantially affect whether a pair of weaker forecasters closes the gap to the group benchmark.

\begin{figure}[H]
\centering
\hspace*{22pt}
\subfloat[Linear pooling.%
\label{fig:case-reliability-linear}]{%
    \hspace*{-22pt}\resizebox{!}{5.45cm}{\begin{tikzpicture}
\begin{groupplot}[
forecast axes front,
group style={group size=1 by 2,vertical sep=0.85cm},
width=0.30\linewidth,height=3.65cm,
xmin=0,xmax=1,xtick={0,0.5,1},
tick label style={font=\fontsize{4}{5}\selectfont},
label style={font=\fontsize{4.5}{5.5}\selectfont},
title style={font=\scriptsize},
grid=major,grid style={black!10},scaled ticks=false
]
\nextgroupplot[
ymin=0,ymax=1,ytick={0,0.50,1},
xticklabels=\empty,ylabel={Yes frequency}
]
\addplot[
black!50,dashed,mark=none,forget plot
]
    coordinates {(0,0) (1,1)};
\draw[bslinear,line width=1pt]
    (axis cs:0.02613676,0.02613676) --
    (axis cs:0.02613676,0.01070336);
\draw[bslinear,line width=1pt]
    (axis cs:0.13733722,0.13733722) --
    (axis cs:0.13733722,0.20320856);
\draw[bslinear,line width=1pt]
    (axis cs:0.26376750,0.26376750) --
    (axis cs:0.26376750,0.25217391);
\draw[bslinear,line width=1pt]
    (axis cs:0.35133618,0.35133618) --
    (axis cs:0.35133618,0.31536388);
\draw[bslinear,line width=1pt]
    (axis cs:0.44213366,0.44213366) --
    (axis cs:0.44213366,0.45945946);
\draw[bslinear,line width=1pt]
    (axis cs:0.55377985,0.55377985) --
    (axis cs:0.55377985,0.47478992);
\draw[bslinear,line width=1pt]
    (axis cs:0.62509047,0.62509047) --
    (axis cs:0.62509047,0.89843750);
\addplot[
    bslinear,
    mark=square*,
    mark size=2.1pt,
    thick,
    mark options={solid, draw=white, line width=1.2pt, fill=bslinear},
    ]
    coordinates {
    (0.02613676,0.01070336) (0.13733722,0.20320856)
    (0.26376750,0.25217391) (0.35133618,0.31536388)
    (0.44213366,0.45945946) (0.55377985,0.47478992)
    (0.62509047,0.89843750)
    };
\node[anchor=south east,fill=white,font=\fontsize{4}{5}\selectfont]
    at (rel axis cs:0.98,0.02) {ECE $=0.0465$};

\nextgroupplot[
height=2.4cm,ymin=0,ymax=50,ytick={0,20,40},
ylabel={Forecasts (\%)},
ybar,bar width=8pt
]
\addplot[fill=bslinear,draw=none]
    coordinates {
    (0.050000,25.31939605) (0.150000,7.23964383)
    (0.250000,4.45218738) (0.350000,43.08943089)
    (0.450000,5.72977158) (0.550000,9.21409214)
    (0.650000,4.95547813)
    };
\end{groupplot}
\end{tikzpicture}}%
}
\hspace*{9pt}
\subfloat[Log-odds pooling.%
\label{fig:case-reliability-logodds}]{%
    \hspace*{-9pt}\resizebox{!}{5.45cm}{\begin{tikzpicture}
\begin{groupplot}[
forecast axes front,
group style={group size=1 by 2,vertical sep=0.85cm},
width=0.30\linewidth,height=3.65cm,
xmin=0,xmax=1,xtick={0,0.5,1},
tick label style={font=\fontsize{4}{5}\selectfont},
label style={font=\fontsize{4.5}{5.5}\selectfont},
title style={font=\scriptsize},
grid=major,grid style={black!10},scaled ticks=false
]
\nextgroupplot[
ymin=0,ymax=1,ytick={0,0.5,1},
xticklabels=\empty
]
\addplot[black!50,dashed,mark=none,forget plot]
    coordinates {(0,0) (1,1)};
\draw[bslogodds,line width=1pt]
    (axis cs:0.07084999,0.07084999) --
    (axis cs:0.07084999,0.00000000);
\draw[bslogodds,line width=1pt]
    (axis cs:0.14039795,0.14039795) --
    (axis cs:0.14039795,0.01492537);
\draw[bslogodds,line width=1pt]
    (axis cs:0.24606681,0.24606681) --
    (axis cs:0.24606681,0.14285714);
\draw[bslogodds,line width=1pt]
    (axis cs:0.33925178,0.33925178) --
    (axis cs:0.33925178,0.23484848);
\draw[bslogodds,line width=1pt]
    (axis cs:0.47030151,0.47030151) --
    (axis cs:0.47030151,0.28138528);
\draw[bslogodds,line width=1pt]
    (axis cs:0.53866948,0.53866948) --
    (axis cs:0.53866948,0.32567050);
\draw[bslogodds,line width=1pt]
    (axis cs:0.64624501,0.64624501) --
    (axis cs:0.64624501,0.46323529);
\draw[bslogodds,line width=1pt]
    (axis cs:0.75702954,0.75702954) --
    (axis cs:0.75702954,0.46258503);
\draw[bslogodds,line width=1pt]
    (axis cs:0.85706322,0.85706322) --
    (axis cs:0.85706322,0.77777778);
\draw[bslogodds,line width=1pt]
    (axis cs:0.92241735,0.92241735) --
    (axis cs:0.92241735,1.00000000);
\addplot[
    bslogodds,
    mark=diamond*,
    mark size=2.1pt,
    thick,
    mark options={solid, draw=white, line width=0.6pt, fill=bslogodds},
    ]
    coordinates {
    (0.07084999,0.00000000) (0.14039795,0.01492537)
    (0.24606681,0.14285714) (0.33925178,0.23484848)
    (0.47030151,0.28138528) (0.53866948,0.32567050)
    (0.64624501,0.46323529) (0.75702954,0.46258503)
    (0.85706322,0.77777778) (0.92241735,1.00000000)
    };
\node[anchor=south east,fill=white,font=\fontsize{4}{5}\selectfont]
    at (rel axis cs:0.98,0.02) {ECE $=0.1656$};

\nextgroupplot[
height=2.4cm,ymin=0,ymax=50,ytick={0,20,40},
ybar,bar width=8pt
]
\addplot[fill=bslogodds,draw=none]
    coordinates {
    (0.050000,15.60201316) (0.150000,7.78164925)
    (0.250000,5.14905149) (0.350000,5.11033682)
    (0.450000,8.94308943) (0.550000,40.41811847)
    (0.650000,5.26519551) (0.750000,5.69105691)
    (0.850000,4.18118467) (0.950000,1.85830430)
    };
\end{groupplot}
\end{tikzpicture}}%
}
\hspace*{9pt}
\subfloat[Group benchmark.%
\label{fig:case-reliability-benchmark}]{%
    \hspace*{-9pt}\resizebox{!}{5.45cm}{\begin{tikzpicture}
\begin{groupplot}[
forecast axes front,
group style={group size=1 by 2,vertical sep=0.85cm},
width=0.30\linewidth,height=3.65cm,
xmin=0,xmax=1,xtick={0,0.5,1},
tick label style={font=\fontsize{4}{5}\selectfont},
label style={font=\fontsize{4.5}{5.5}\selectfont},
title style={font=\scriptsize},
grid=major,grid style={black!10},scaled ticks=false
]
\nextgroupplot[
ymin=0,ymax=1,ytick={0,0.5,1},
xticklabels=\empty
]
\addplot[black!50,dashed,mark=none,forget plot]
    coordinates {(0,0) (1,1)};
\draw[bsindividual,line width=1pt]
    (axis cs:0.02589778,0.02589778) --
    (axis cs:0.02589778,0.10831721);
\draw[bsindividual,line width=1pt]
    (axis cs:0.11530121,0.11530121) --
    (axis cs:0.11530121,0.09696970);
\draw[bsindividual,line width=1pt]
    (axis cs:0.22217491,0.22217491) --
    (axis cs:0.22217491,0.18213058);
\draw[bsindividual,line width=1pt]
    (axis cs:0.31538246,0.31538246) --
    (axis cs:0.31538246,0.25000000);
\draw[bsindividual,line width=1pt]
    (axis cs:0.42593259,0.42593259) --
    (axis cs:0.42593259,0.26548673);
\draw[bsindividual,line width=1pt]
    (axis cs:0.52046464,0.52046464) --
    (axis cs:0.52046464,0.29537367);
\draw[bsindividual,line width=1pt]
    (axis cs:0.61865469,0.61865469) --
    (axis cs:0.61865469,0.44012945);
\draw[bsindividual,line width=1pt]
    (axis cs:0.71434994,0.71434994) --
    (axis cs:0.71434994,0.56741573);
\draw[bsindividual,line width=1pt]
    (axis cs:0.81622333,0.81622333) --
    (axis cs:0.81622333,0.72222222);
\draw[bsindividual,line width=1pt]
    (axis cs:0.94069801,0.94069801) --
    (axis cs:0.94069801,0.84210526);
\addplot[
    bsindividual,
    mark=*,
    mark size=2.1pt,
    thick,
    mark options={solid, draw=white, line width=1.2pt, fill=bsindividual},
    ]
    coordinates {
    (0.02589778,0.10831721) (0.11530121,0.09696970)
    (0.22217491,0.18213058) (0.31538246,0.25000000)
    (0.42593259,0.26548673) (0.52046464,0.29537367)
    (0.61865469,0.44012945) (0.71434994,0.56741573)
    (0.81622333,0.72222222) (0.94069801,0.84210526)
    };
\node[anchor=south east,fill=white,font=\fontsize{4}{5}\selectfont]
    at (rel axis cs:0.98,0.02) {ECE $=0.1197$};

\nextgroupplot[
height=2.4cm,ymin=0,ymax=50,ytick={0,20,40},
ybar,bar width=8pt
]
\addplot[fill=bsindividual,draw=none]
    coordinates {
    (0.050000,20.01548587) (0.150000,12.77584204)
    (0.250000,11.26596980) (0.350000,5.26519551)
    (0.450000,4.37475803) (0.550000,21.75764615)
    (0.650000,11.96283391) (0.750000,6.89121177)
    (0.850000,3.48432056) (0.950000,2.20673635)
    };
\end{groupplot}
\end{tikzpicture}}%
}
\caption{
\rxedit{
Reliability and forecast distributions for a fixed eligible pair under each learned pooling rule and the group benchmark in G13.
}
}\label{fig:sec4-case-reliability}
\end{figure}
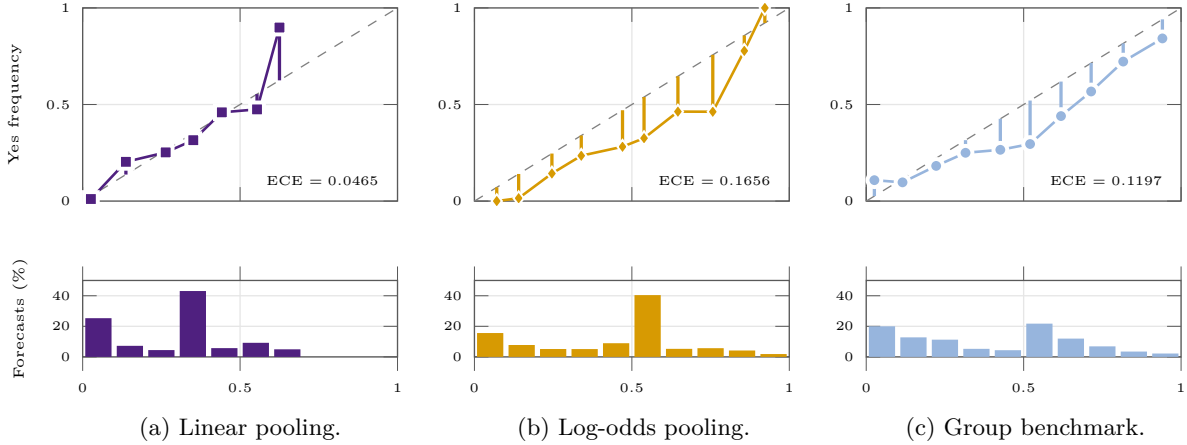

\rxedit{
\Cref{fig:sec4-case-reliability} looks into G13 through the lens \aiedit{of} calibration. \Cref{fig:case-reliability-linear} and \Cref{fig:case-reliability-logodds} combine Claude-3-5-Sonnet with Llama-3-70b-Chat-Hf using learned linear and log-odds pooling, respectively; panel \Cref{fig:case-reliability-benchmark} shows the BS-best individual, Claude-3-Opus. The three panels evaluate the same 779 test subquestions in G13. The binning and plotting conventions follow \Cref{fig:overall-reliability-1000}.
} 

\section{Additional Multi-Model Aggregation Results}
\label{apx:multi-aggregation}

\begin{figure}[h]
\captionsetup[subfloat]{font=footnotesize,justification=centering}
\noindent
\begin{minipage}[t]{0.4\linewidth}
    \vspace{0pt}
    \captionsetup[subfloat]{margin={0.21\linewidth,0pt}}
    \subfloat[Group-level success\\by number of forecasters.%
    \label{fig:sec4-three-models-a}]{%
        \resizebox{\linewidth}{!}{
\begin{tikzpicture}
\begin{axis}[
forecast axes front,
    width=6.4cm, height=4.8cm,
    xmin=1.85, xmax=5.15,
    xtick={2,3,4,5},
    tick label style={font=\scriptsize},
    label style={font=\scriptsize},
    yticklabel={\pgfmathprintnumber{\tick}\%},
    scaled ticks=false,
    ymajorgrids=true,
    grid style={black!10},
    tick align=outside,
    every axis plot/.append style={mark options={solid}},
    ymin=0, ymax=100,
    ytick={0,25,50,75,100},
    ylabel={Successful groups (\%)},
    legend to name=leg:three-models,
    legend columns=1,
    legend cell align=left,
    legend style={
        font=\scriptsize, draw=none, fill=none,
        inner sep=0pt, column sep=2pt, row sep=1pt
    },
]
        \addplot[
            bslinear, solid, thick,
            mark=square*, mark size=3.6pt,
            mark options={solid, draw=white, line width=1.2pt, fill=bslinear}
        ] 
        coordinates {
            (2,68.75) (3,75.00) (4,68.75) (5,68.75)
        };
        \addlegendentry{\shortstack[l]{Learned linear\\pooling}}

        \addplot[
            bslogodds, solid, thick,
            mark=diamond*, mark size=3.6pt,
            mark options={solid, draw=white, line width=1.2pt, fill=bslogodds}
        ] coordinates {
            (2,50.00) (3,68.75) (4,68.75) (5,68.75)
        };
        \addlegendentry{\shortstack[l]{Learned log-odds\\pooling}}

        \addplot[
            forecastblue!85!black, solid, thick,
            mark=triangle*, mark size=3.6pt,
            mark options={solid, draw=white, line width=1.2pt, fill=forecastblue!85!black}
        ] coordinates {
            (2,50.00) (3,50.00) (4,56.25) (5,50.00)
        };
        \addlegendentry{Simple mean}

        \addplot[
           black!60, solid, thick,
           mark=*, mark size=3.6pt,
           mark options={solid, draw=white, line width=2pt, fill=black!60}
        ] 
        coordinates {
            (2,50.00) (3,50.00) (4,43.75) (5,50.00)
        };
        \addlegendentry{Log-odds mean}

        \addplot[
            forecastteal!85!black, solid, thick,
            mark=pentagon*, mark size=3.6pt,
            mark options={solid, draw=white, line width=1.2pt, fill=forecastteal!85!black}
        ] coordinates {
            (2,50.00) (3,50.00) (4,43.75) (5,25.00)
        };
        \addlegendentry{Harmonic mean}

        \addplot[
            forecastred!85!black, solid, thick,
            mark=star, mark size=3.6pt,
            mark options={solid, draw=forecastred!85!black, fill=white}
        ] coordinates {
            (2,50.00) (3,56.25) (4,68.75) (5,56.25)
        };
        \addlegendentry{Median}

        \addlegendimage{line legend,forecastblue!85!black,densely dashed,semithick,mark=none}
        \addlegendentry{\shortstack[l]{Group Benchmark}}

\end{axis}
\end{tikzpicture}}%
    }
\end{minipage}\hfill%
\begin{minipage}[t]{0.4\linewidth}
    \vspace{0pt}
    \captionsetup[subfloat]{margin={0.22\linewidth,0pt}}
    \subfloat[Best performance in G13\\by number of forecasters.%
    \label{fig:sec4-three-models-b}]{%
        \resizebox{\linewidth}{!}{
\begin{tikzpicture}
\begin{axis}[
forecast axes front,
width=6.4cm,height=4.8cm,
xmin=1.5,xmax=5.5,ymin=0.79,ymax=0.86,
yticklabel style={
/pgf/number format/fixed zerofill,
/pgf/number format/precision=2
},
xtick={2,3,4,5},
ytick={0.79,0.82,0.86},
ylabel={1 $-$ Test Brier score},
tick label style={font=\scriptsize},label style={font=\scriptsize},
scaled ticks=false,ymajorgrids=true,grid style={black!10},tick align=outside,
bar width=3pt,
]

\path[fill=black!7,draw=none] (axis cs:1.5,0.808199747520) rectangle (axis cs:5.5,0.832304861246);
\addplot+[forget plot,
ybar,
bar shift=0pt,
fill=bslinear,
draw=bslinear,
mark=none,
error bars/.cd,y dir=both,y explicit,error mark=-,error bar style={gray,line width=0.60pt},error mark options={mark size=0.20pt,line width=2.80pt}
] 
coordinates {(1.730,0.847623482313) += (0,0.009353720096) -= (0,0.009353720096) (2.730,0.847623482319) += (0,0.009353720032) -= (0,0.009353720032) (3.730,0.847623482330) += (0,0.009353719976) -= (0,0.009353719976) (4.730,0.847623482373) += (0,0.009353719699) -= (0,0.009353719699)};

\addplot+[forget plot,
ybar,
bar shift=0pt,
fill=bslogodds,
draw=bslogodds,
mark=none,
error bars/.cd,y dir=both,y explicit,error mark=-,error bar style={gray,line width=0.60pt},error mark options={mark size=0.20pt,line width=2.80pt}
] 
coordinates {(1.838,0.826680186987) += (0,0.009891824181) -= (0,0.009891824181) (2.838,0.827166120931) += (0,0.009810467515) -= (0,0.009810467515) (3.838,0.827166121088) += (0,0.009810467409) -= (0,0.009810467409) (4.838,0.827166121121) += (0,0.009810467276) -= (0,0.009810467276)};

\addplot+[forget plot,
ybar,
bar shift=0pt,
fill=forecastblue!85!black,
draw=forecastblue!85!black,
mark=none,
error bars/.cd,y dir=both,y explicit,error mark=-,error bar style={gray,line width=0.60pt},error mark options={mark size=0.20pt,line width=2.80pt}
] 
coordinates {(1.946,0.825841829006) += (0,0.010215057487) -= (0,0.010215057487) (2.946,0.824783600645) += (0,0.010240446319) -= (0,0.010240446319) (3.946,0.820982830380) += (0,0.010344368548) -= (0,0.010344368548) (4.946,0.818559146756) += (0,0.010395728950) -= (0,0.010395728950)};

\addplot+[forget plot,
ybar,
bar shift=0pt,
fill=black!60,
draw=black!60,
mark=none,
error bars/.cd,y dir=both,y explicit,error mark=-,error bar style={gray,line width=0.60pt},error mark options={mark size=0.20pt,line width=2.80pt}
] 
coordinates {(2.054,0.822691428164) += (0,0.011485166923) -= (0,0.011485166923) (3.054,0.820495664340) += (0,0.011700739743) -= (0,0.011700739743) (4.054,0.816433894180) += (0,0.012108659848) -= (0,0.012108659848) (5.054,0.813281744198) += (0,0.012300534330) -= (0,0.012300534330)};

\addplot+[forget plot,
ybar,
bar shift=0pt,
fill=forecastteal!85!black,
draw=forecastteal!85!black,
mark=none,
error bars/.cd,y dir=both,y explicit,error mark=-,error bar style={gray,line width=0.60pt},error mark options={mark size=0.20pt,line width=2.80pt}
] 
coordinates {(2.162,0.826814975834) += (0,0.011623126518) -= (0,0.011623126518) (3.162,0.819463405639) += (0,0.012258486321) -= (0,0.012258486321) (4.162,0.813887926918) += (0,0.012071445376) -= (0,0.012071445376) (5.162,0.809464147539) += (0,0.012839721681) -= (0,0.012839721681)};

\addplot+[forget plot,
ybar,
bar shift=0pt,
fill=forecastred!85!black,
draw=forecastred!85!black,
mark=none,
error bars/.cd,y dir=both,y explicit,error mark=-,error bar style={gray,line width=0.60pt},error mark options={mark size=0.20pt,line width=2.80pt}
] 
coordinates {(2.270,0.825841829006) += (0,0.010215057487) -= (0,0.010215057487) (3.270,0.823526100032) += (0,0.011486863239) -= (0,0.011486863239) (4.270,0.821286650669) += (0,0.010357794982) -= (0,0.010357794982) (5.270,0.815877062418) += (0,0.011476846583) -= (0,0.011476846583)};

\addplot[forget plot,forecastblue!85!black,densely dashed,semithick] coordinates {(1.5,0.820252304383) (5.5,0.820252304383)};
\end{axis}
\end{tikzpicture}}%
    }
\end{minipage}\hfill%
\begin{minipage}[t]{0.18\linewidth}
    \vspace{10pt}
    \raggedright
    \resizebox{0.87\linewidth}{!}{\pgfplotslegendfromname{leg:three-models}}
\end{minipage}
\caption{
Aggregation performance across exact pool sizes. (a) Fraction of the 16 comparison groups containing at least one pool that matches or improves on the group benchmark in terms of test BS, for pool sizes \(k\in\{2,3,4,5\}\). 
(b) Best test \(1-\mathrm{BS}\) in group G13 for each aggregation rule and pool size. The dashed line marks the group benchmark in group G13. 
For each aggregation rule and pool size, the best-performing pool is selected separately.
}
\label{fig:sec4-three-models}
\end{figure}
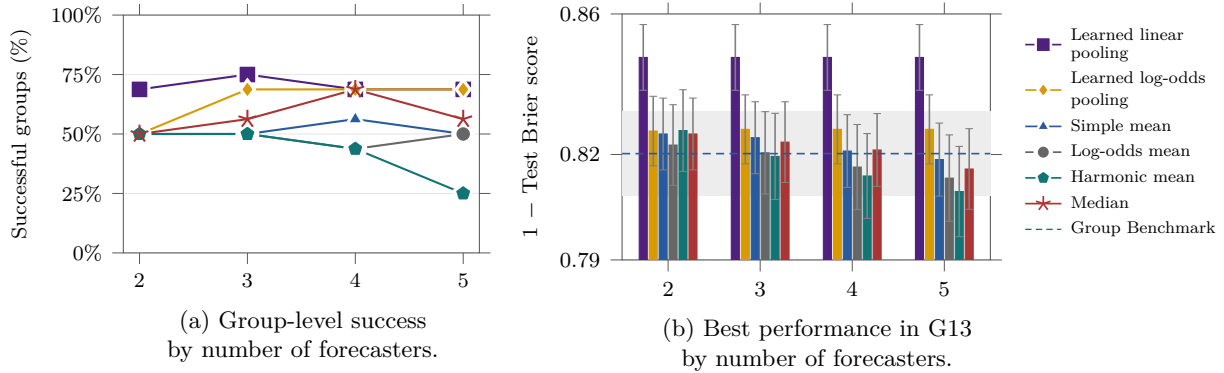

We provide additional comparisons across aggregation rules for the multi-model experiments in \Cref{sec:multi-aggregation}. 
In \Cref{fig:sec4-three-models}, we report results for exact pool sizes \(k\in\{2,3,4,5\}\). 
For each aggregation rule and pool size, the best-performing pool is selected separately. 
Thus, the curves compare the best available pools at each exact size rather than tracking a fixed pool as additional forecasters are added.
In \Cref{fig:sec4-three-models}, we additionally compare the learned rules with the classical aggregation rules. 
At the group level, learned linear pooling remains among the strongest-performing rules across all pool sizes, whereas the classical rules exhibit more variable performance as the pool grows. 
In particular, their success rates do not generally increase with pool size, and some decline for larger pools. \Cref{fig:sec4-three-models-b} shows a similar pattern within G13: the best performance under the classical rules fluctuates or decreases as the pool grows and generally remains below that of learned linear pooling. 
Thus, the lack of systematic gains from larger pools is also visible across the fixed aggregation rules, rather than being specific to the learned rules.

\section{Practical Constraints}
\label{apx:prac-constraint-materials}

We next examine whether useful weak-pair aggregates remain available when access to the strongest individual forecaster is constrained in practice. 
We consider two restrictions: {\em lower input price} and {\em open-weight access}. 
In both analyses, we start from the same two-model comparisons as in \Cref{sec:BS-results} and keep the fitted aggregation rules and group benchmarks fixed. 
\xhdr{Lower input price} 
For each model, we use the published API input price, measured per million input tokens, at the time the specific model version (snapshot) evaluated in our experiments was released. 
We use input price as a simple and comparable cost proxy because the forecasting task typically requires only a short probability output. 
For a pair to satisfy the lower-input-price constraint, the sum of the two models' published input prices must be strictly below that of the group benchmark. We then ask whether such a cheaper pair can still match or approach the benchmark after aggregation. 
\xhdr{Open-weight access} We also consider settings in which the group benchmark is closed-weight but locally deployable models are required. We restrict attention to the 10 comparison groups with a closed-weight benchmark and at least one eligible open-weight pair, and evaluate aggregates formed only from such open-weight forecasters. Our open-weight classifications are reported in \Cref{tab:all-evaluated-models}.

\xhdr{Group-level results}
In \Cref{tab:price-and-availability}, we summarize the performance under both constraints. 
Under the lower-input-price restriction, learned linear pooling identifies a pair that matches or improves on the benchmark in 2 of the 16 groups and comes within 5\% in 5 groups; learned log-odds pooling does so in 1 and 3 groups, respectively. 
Under the open-weight restriction, learned linear pooling matches or improves on the benchmark in 3 of the 10 eligible groups and comes within 5\% in 4, compared with 1 and 3 groups for learned log-odds pooling. 
Thus, although these restrictions substantially narrow the candidate set, they still leave comparison groups in which two individually weaker models provide a competitive alternative to the stronger benchmark


\begin{table}[h]
\captionsetup{font=normalsize}
\caption{\wtedit{Group-level results under practical constraints. Entries report the fraction of groups with at least one constrained pair that matches or improves on, or comes within 5\% of, the group benchmark. Denominators are 16 groups for lower input price and 10 groups for open weights.}}
\label{tab:price-and-availability}
\label{tab:overall-input-cost-outcomes}
\label{tab:overall-two-open-weight-vs-closed}
\resulttablesetup
\resizebox{0.80\linewidth}{!}{%
\begin{tabular}{@{}lcccc@{}}
\toprule
 
& \multicolumn{2}{c}{Lower input price}
& \multicolumn{2}{c}{Open weights} \\
\cmidrule(lr){2-3}\cmidrule(lr){4-5}
Aggregation rule
& \shortstack{Match or improve\\benchmark}
& \shortstack{Within $5\%$\\of benchmark}
& \shortstack{Match or improve\\benchmark}
& \shortstack{Within $5\%$\\of benchmark} \\
\midrule

Learned linear pooling
& \shortstack{2/16\\(12.50\%)}
& \shortstack{5/16\\(31.25\%)}
& \shortstack{3/10\\(30.00\%)}
& \shortstack{4/10\\(40.00\%)} \\

Learned log-odds pooling
& \shortstack{1/16\\(6.25\%)}
& \shortstack{3/16\\(18.75\%)}
& \shortstack{1/10\\(10.00\%)}
& \shortstack{3/10\\(30.00\%)} \\

\bottomrule
\end{tabular}

%
}
\end{table}

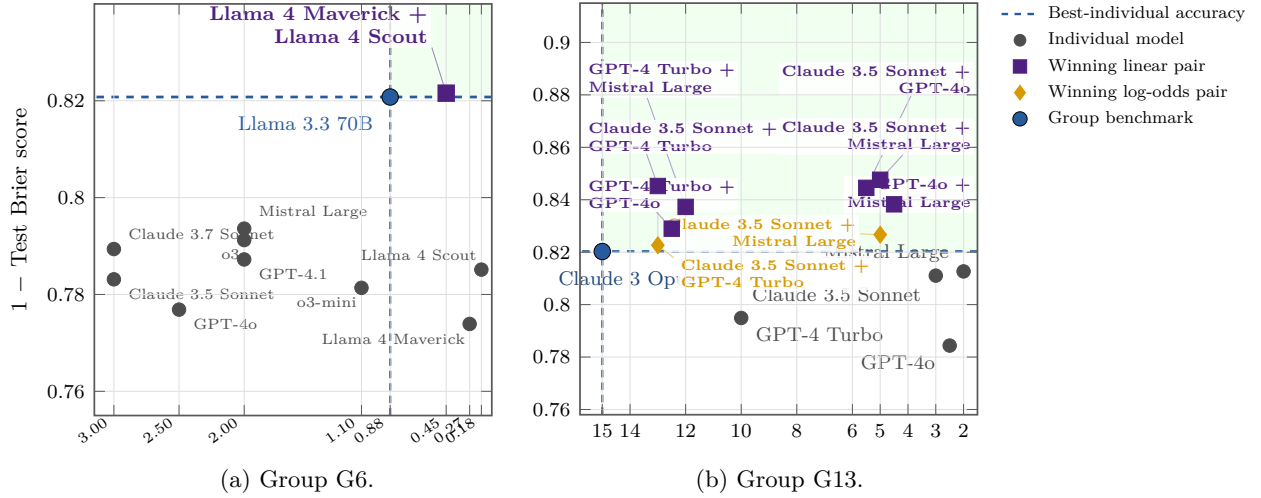
\begin{figure}[h]
\captionsetup[subfloat]{font=footnotesize,justification=centering}
\newsavebox{\pricePanelARaw}
\newsavebox{\pricePanelBRaw}
\newsavebox{\pricePanelA}
\newsavebox{\pricePanelB}
\sbox{\pricePanelARaw}{
\begin{tikzpicture}
\begin{axis}[
    width=7.2cm,
    height=7.4cm,
    xmin=0.10,xmax=3.15,
    x dir=reverse,
    ymin=0.755,ymax=0.840,
    ylabel={1 $-$ Test Brier score},
    xtick={0.18,0.27,0.45,0.88,1.1,2,2.5,3},
    xticklabels={0.18,0.27,0.45,0.88,1.10,2.00,2.50,3.00},
    x tick label style={rotate=35,anchor=east,font=\tiny},
    ytick={0.76,0.78,0.80,0.82},
    y tick label style={/pgf/number format/fixed,/pgf/number format/precision=2},
    grid=both,
    major grid style={gray!25},minor grid style={gray!12},
    axis line style={black!65,line width=0.7pt},tick style={black!65},
    tick label style={font=\scriptsize},label style={font=\footnotesize},
    clip=false,axis on top
]

\path[fill=green!6,draw=none]
    (axis cs:0.10,0.8207778517) rectangle (axis cs:0.88000000,0.840);
\addplot[dashed,line width=1.1pt,color=forecastblue!85!black,mark=none]
    coordinates {(0.10,0.8207778517) (3.15,0.8207778517)};
\addplot[densely dashed,line width=0.9pt,color=forecastblue!65!black,mark=none,forget plot]
    coordinates {(0.88000000,0.755) (0.88000000,0.840)};

\addplot[only marks,mark=*,mark size=2.7pt,color=black!70]
    coordinates {(2.00000000,0.7936024130) (2.00000000,0.7912310923) (3.00000000,0.7893840265) (2.00000000,0.7872310519) (0.18000000,0.7851330004) (3.00000000,0.7831281372) (1.10000000,0.7813720503) (2.50000000,0.7768714972) (0.27000000,0.7739137384)};
\addplot[only marks,mark=square*,mark size=3.3pt,color=bslinear]
    coordinates {(0.45000000,0.8215718233)};
\addplot[only marks,mark=*,mark size=3.3pt,
    mark options={fill=forecastblue!85!black,draw=black},color=forecastblue!85!black]
    coordinates {(0.88000000,0.8207778517)};

\node[anchor=south west,font=\tiny,text=black!70,xshift=2pt] at (axis cs:2.0,0.7936024129667568) {Mistral Large};
\node[anchor=north east,font=\tiny,text=black!70,xshift=2pt] at (axis cs:2.0,0.7912310922706995) {o3};
\node[anchor=south west,font=\tiny,text=black!70,xshift=2pt] at (axis cs:3.0,0.7893840265241547) {Claude 3.7 Sonnet};
\node[anchor=north west,font=\tiny,text=black!70,xshift=2pt] at (axis cs:2.0,0.7872310518515151) {GPT-4.1};
\node[anchor=south east,font=\tiny,text=black!70,xshift=2pt] at (axis cs:0.18,0.7851330004312717) {Llama 4 Scout};
\node[anchor=north west,font=\tiny,text=black!70,xshift=2pt] at (axis cs:3.0,0.7831281371979475) {Claude 3.5 Sonnet};
\node[anchor=north east,font=\tiny,text=black!70,xshift=2pt] at (axis cs:1.1,0.7813720502933796) {o3-mini};
\node[anchor=north west,font=\tiny,text=black!70,xshift=2pt] at (axis cs:2.5,0.7768714972204672) {GPT-4o};
\node[anchor=north east,font=\tiny,text=black!70,xshift=2pt] at (axis cs:0.27,0.7739137384039759) {Llama 4 Maverick};
\node[anchor=north east,font=\scriptsize,text=forecastblue!85!black,xshift=-3pt,yshift=-4pt]
    at (axis cs:0.88000000,0.8207778517) {Llama 3.3 70B};
\draw[bslinear!55,thin]
    (axis cs:0.45,0.8215718233) -- (axis cs:0.56,0.8310);
\node[anchor=south east,font=\scriptsize\bfseries,text=bslinear,
    align=right,fill=white,fill opacity=0.88,text opacity=1,inner sep=1.5pt]
    at (axis cs:0.56,0.8310)
    {Llama 4 Maverick +\\Llama 4 Scout};
\end{axis}
\end{tikzpicture}}
\sbox{\pricePanelBRaw}{
\pgfplotsset{grid style={on background layer}}
\begin{tikzpicture}
\begin{axis}[
    width=7.2cm,
    height=7.4cm,
    xmin=1.5,xmax=15.8,
    x dir=reverse,
    ymin=0.758,ymax=0.915,
    xtick={2,3,4,5,6,8,10,12,14,15},
    ytick={0.76,0.78,0.80,0.82,0.84,0.86,0.88,0.90},
    y tick label style={/pgf/number format/fixed,/pgf/number format/precision=2},
    grid=both,
    major grid style={gray!25},
    minor grid style={gray!12},
    axis line style={black!65,line width=0.7pt},
    tick style={black!65},
    tick label style={font=\scriptsize},
    label style={font=\footnotesize},
    legend to name=leg:input-price-common,
    legend columns=1,
    legend style={font=\scriptsize,draw=none,fill=none,
        inner sep=0pt,column sep=2pt,row sep=1pt},
    legend cell align=left,
    clip=false,
    axis on top
]

\path[fill=green!6,draw=none]
    (axis cs:1.5,0.8202523044) rectangle (axis cs:15.00000000,0.915);

\addplot[dashed,line width=1.1pt,color=forecastblue!85!black,mark=none]
    coordinates {(1.5,0.8202523044) (15.8,0.8202523044)};
\addlegendentry{Best-individual accuracy}
\addplot[densely dashed,line width=0.9pt,color=forecastblue!65!black,mark=none,forget plot]
    coordinates {(15.00000000,0.758) (15.00000000,0.915)};

\addplot[gray!35,thin,forget plot] coordinates {(5.00000000,0.8266801870) (5.00000000,0.8476234823)};
\addplot[gray!35,thin,forget plot] coordinates {(13.00000000,0.8226511357) (13.00000000,0.8452363666)};

\addplot[only marks,mark=*,mark size=2.7pt,color=black!70]
    coordinates {(3.00000000,0.8110481660)};
\addlegendentry{Individual model}
\addplot[only marks,mark=*,mark size=2.7pt,color=black!70,forget plot]
    coordinates {(10.00000000,0.7949011614) (2.50000000,0.7843741110)};
\addplot[only marks,mark=*,mark size=2.7pt,color=black!70,forget plot]
    coordinates {(2.00000000,0.8126620678)};

\addplot[only marks,mark=square*,mark size=3.1pt,color=bslinear]
    coordinates {(4.50000000,0.8382490987) (5.00000000,0.8476234823) (5.50000000,0.8445437048) (12.00000000,0.8372724643) (12.50000000,0.8289660107) (13.00000000,0.8452363666)};
\addlegendentry{Winning linear pair}
\addplot[only marks,mark=diamond*,mark size=3.4pt,color=bslogodds]
    coordinates {(5.00000000,0.8266801870) (13.00000000,0.8226511357)};
\addlegendentry{Winning log-odds pair}

\addplot[only marks,mark=*,mark size=3.3pt,
    mark options={fill=forecastblue!85!black,draw=black},color=forecastblue!85!black]
    coordinates {(15.00000000,0.8202523044)};
\addlegendentry{Group benchmark}

\node[anchor=south east,font=\scriptsize,text=black!70,xshift=-2pt]
    at (axis cs:2,0.8126620678) {Mistral Large};
\node[anchor=north east,font=\scriptsize,text=black!70,xshift=-2pt]
    at (axis cs:2.5,0.7843741110) {GPT-4o};
\node[anchor=north east,font=\scriptsize,text=black!70,xshift=-2pt,yshift=-1pt]
    at (axis cs:3,0.8110481660) {Claude 3.5 Sonnet};
\node[anchor=north west,font=\scriptsize,text=black!70,xshift=2pt]
    at (axis cs:10,0.7949011614) {GPT-4 Turbo};
\node[anchor=north,
font=\scriptsize,
text=forecastblue!65!black,
xshift=10pt,
yshift=-4pt]
    at (axis cs:15,0.8202523044) {Claude 3 Opus};

\draw[bslinear!45,thin] (axis cs:12.50000000,0.8289660107) -- (axis cs:13.62,0.842);
\draw[bslinear!45,thin] (axis cs:13.00000000,0.8452363666) -- (axis cs:13.62,0.864);
\draw[bslinear!45,thin] (axis cs:12.00000000,0.8372724643) -- (axis cs:13.62,0.886);
\node[anchor=west,font=\tiny\bfseries,text=bslinear,align=left,fill=white,fill opacity=0.88,text opacity=1,inner sep=1.2pt]
    at (axis cs:15.62,0.842) {GPT-4 Turbo +\\GPT-4o};
\node[anchor=west,font=\tiny\bfseries,text=bslinear,align=left,fill=white,fill opacity=0.88,text opacity=1,inner sep=1.2pt]
    at (axis cs:15.62,0.864) {Claude 3.5 Sonnet +\\GPT-4 Turbo};
\node[anchor=west,font=\tiny\bfseries,text=bslinear,align=left,fill=white,fill opacity=0.88,text opacity=1,inner sep=1.2pt]
    at (axis cs:15.62,0.886) {GPT-4 Turbo +\\Mistral Large};
\draw[bslinear!45,thin] (axis cs:4.50000000,0.8382490987) -- (axis cs:3.60,0.842);
\draw[bslinear!45,thin] (axis cs:5.00000000,0.8476234823) -- (axis cs:3.60,0.864);
\draw[bslinear!45,thin] (axis cs:5.50000000,0.8445437048) -- (axis cs:3.60,0.886);
\node[anchor=east,font=\tiny\bfseries,text=bslinear,align=right,fill=white,fill opacity=0.88,text opacity=1,inner sep=1.2pt]
    at (axis cs:1.62,0.842) {GPT-4o +\\Mistral Large};
\node[anchor=east,font=\tiny\bfseries,text=bslinear,align=right,fill=white,fill opacity=0.88,text opacity=1,inner sep=1.2pt]
    at (axis cs:1.62,0.864) {Claude 3.5 Sonnet +\\Mistral Large};
\node[anchor=east,font=\tiny\bfseries,text=bslinear,align=right,fill=white,fill opacity=0.88,text opacity=1,inner sep=1.2pt]
    at (axis cs:1.62,0.886) {Claude 3.5 Sonnet +\\GPT-4o};
\draw[bslogodds!55,thin] (axis cs:5.05000000,0.8266801870) -- (axis cs:5.72,0.827);
\node[anchor=east,font=\tiny\bfseries,text=bslogodds,align=right,fill=white,fill opacity=0.90,text opacity=1,inner sep=1.2pt]
    at (axis cs:5.75,0.827) {Claude 3.5 Sonnet +\\Mistral Large};
\draw[bslogodds!55,thin] (axis cs:12.93000000,0.8220) -- (axis cs:12.42,0.8178);
\node[anchor=north west,font=\tiny\bfseries,text=bslogodds,align=left,fill=white,fill opacity=0.90,text opacity=1,inner sep=1.2pt]
    at (axis cs:12.30,0.8185) {Claude 3.5 Sonnet +\\GPT-4 Turbo};
\end{axis}
\end{tikzpicture}}
\pgfmathsetmacro{\pricePanelScale}{0.78*\the\linewidth/(\the\wd\pricePanelARaw+\the\wd\pricePanelBRaw)}
\sbox{\pricePanelA}{\scalebox{\pricePanelScale}{\usebox{\pricePanelARaw}}}
\sbox{\pricePanelB}{\scalebox{\pricePanelScale}{\usebox{\pricePanelBRaw}}}
\newlength{\pricePanelHeight}
\setlength{\pricePanelHeight}{\dimexpr\ht\pricePanelA+\dp\pricePanelA\relax}
\ifdim\dimexpr\ht\pricePanelB+\dp\pricePanelB\relax>\pricePanelHeight
    \setlength{\pricePanelHeight}{\dimexpr\ht\pricePanelB+\dp\pricePanelB\relax}
\fi
\noindent
\begin{minipage}[t]{\wd\pricePanelA}
    \vspace{0pt}
    \captionsetup[subfloat]{margin={0.18\linewidth,0pt}}
    \subfloat[Group G6.%
    \label{fig:g07-input-price-one-minus-brier}]{%
        \vbox to\pricePanelHeight{\hbox{\usebox{\pricePanelA}}\vfil}%
    }
\end{minipage}\hfill%
\begin{minipage}[t]{\wd\pricePanelB}
    \vspace{0pt}
    \captionsetup[subfloat]{margin={0.12\linewidth,0pt}}
    \subfloat[Group G13.%
    \label{fig:g13-input-price-one-minus-brier}]{%
        \vbox to\pricePanelHeight{\hbox{\usebox{\pricePanelB}}\vfil}%
    }
\end{minipage}\hfill%
\begin{minipage}[t]{0.20\linewidth}
    \vspace{0pt}
    \raggedright
    \resizebox{\linewidth}{!}{\pgfplotslegendfromname{leg:input-price-common}}
\end{minipage}
\caption{Input-price versus BS frontiers for groups. \Cref{fig:g07-input-price-one-minus-brier} shows G6 (1,512 common subquestions) and \Cref{fig:g13-input-price-one-minus-brier} shows G13 (1,557 common subquestions) at the 1,000-coverage threshold.
The x-axis \aiedit{shows} published input list prices (USD per million tokens), decreasing from left to right.
Gray circles are individual models; the blue circle is each group's BS-best model.
Purple squares and gold diamonds show, respectively, linear and log-odds input-cheaper weak-model pairs whose learned aggregate strictly exceeds the benchmark on the test half; \Cref{fig:g07-input-price-one-minus-brier} contains only one such pair.
Pair cost is the sum of the two members' standard input-token rates.
Complete pair names are attached directly to their aggregate points, and method-specific scores example appear in \Cref{tab:practical-best-pairs}.}
\label{fig:input-price-frontiers}
\end{figure}

\xhdr{Price-accuracy trade-offs}
In \Cref{fig:input-price-frontiers}, we provide a more detailed view of the lower-input-price constraint in groups G6 and G13. 
The figure places individual models and successful constrained pairs in the same price–accuracy space, where the cost of a pair is the sum of the published input prices of its two members. 
A successful constrained pair therefore provides both lower input cost and higher forecast accuracy than the group benchmark. 
As we can see, G6 contains one such pair, whereas G13 contains several alternatives across different price levels.     

The gains can be substantial when such alternatives exist. In G13, for example, Claude-3-5-Sonnet combined with Mistral-Large has a total input price of \$5 per million tokens, compared with \$15 for the Claude-3-Opus benchmark. The pair reduces BS by 15.23\% under learned linear pooling and by 3.58\% under learned log-odds pooling. Figure 9 therefore illustrates that weak-to-strong aggregation can generate a meaningful price–accuracy trade-off rather than merely a near-benchmark substitute

\xhdr{Best-performing constrained pairs} 
In \Cref{tab:practical-best-pairs}, we report, for each practical constraint and learned aggregation rule, the constrained pair with the largest relative BS reduction. These examples complement the group-level results in \Cref{tab:price-and-availability} by showing the magnitude of the gains when a successful constrained alternative exists. 
Under the open-weight constraint, for example, the best learned-linear pair combines Mistral-Large with Qwen3-235B-A22B and reduces BS by 9.72\% relative to the Claude-3-Opus benchmark. The corresponding learned log-odds example combines Llama-3.3-70B with Llama-3.1-405B and reduces BS by 4.87\% relative to O4-Mini.

\newsavebox{\apxPracticalTableBox}
\begingroup
\begin{table}[H]
\centering
\setlength{\tabcolsep}{2.0pt}
\renewcommand{\arraystretch}{1.08}
\fontsize{8.0}{9.5}\selectfont
\captionsetup{font=normalsize}
\caption{Pairs with the largest relative BS reduction under each practical constraint and learned rule. We select each example retrospectively relative to its own group's benchmark. Both members are individually weaker than that benchmark under BS. The final column reports the percentage reduction in BS; larger is better. Model names are shortened for readability.}
\label{tab:practical-best-pairs}
\sbox{\apxPracticalTableBox}{%
\begin{tabular}{@{}lll l rrr@{}}
\toprule
\shortstack[l]{Constraint\\and group} & Learned rule & Pair & Group benchmark
& \shortstack{Benchmark\\BS $\downarrow$} & \shortstack{Pair\\BS $\downarrow$}
& \shortstack{Relative BS\\reduction $\uparrow$} \\
\midrule
\shortstack[l]{Input price\\G13} & Learned linear & \shortstack[l]{Claude-3-5-Sonnet\\+ Mistral-Large} & Claude-3-Opus & 0.1797 & 0.1524 & 15.23\% \\
\shortstack[l]{Input price\\G13} & Learned log-odds & \shortstack[l]{Claude-3-5-Sonnet\\+ Mistral-Large} & Claude-3-Opus & 0.1797 & 0.1733 & 3.58\% \\
\midrule
\shortstack[l]{Open weights\\G13} & Learned linear & \shortstack[l]{Mistral-Large\\+ Qwen3-235B-A22B} & Claude-3-Opus & 0.1797 & 0.1623 & 9.72\% \\
\shortstack[l]{Open weights\\G2} & Learned log-odds & \shortstack[l]{Llama-3.3-70B\\+ Llama-3.1-405B} & O4-Mini & 0.1466 & 0.1395 & 4.87\% \\
\bottomrule
\end{tabular}
}
\ifdim\wd\apxPracticalTableBox>\linewidth
  \resizebox{\linewidth}{!}{\usebox{\apxPracticalTableBox}}%
\else
  \resizebox{\wd\apxPracticalTableBox}{!}{\usebox{\apxPracticalTableBox}}%
\fi
\end{table}
\endgroup

\section{Additional Category-Specific Results}
\label{apx:event-type-materials}

\wtedit{We provide additional details for the category-specific analysis in \Cref{sec:event-type}. 
We first summarize the available data and comparison groups across all question categories, and then report the full group- and pair-level results for Politics and Finance, together with detailed group compositions and group-level comparisons.}

\xhdr{Category-specific data and comparison groups}
\wtedit{In \Cref{tab:event-type-data-statistics}, we summarize the available data after constructing comparison groups separately within each question category at the shared-coverage threshold \(N=1{,}000\). \rxedit{We note that the $6$ categories do not cover all the subquestions as some subquestions cannot be categorized properly.}
Politics and Finance yield 12 and 9 eligible maximal groups, respectively, while Climate yields only two groups and Science, Sports, and Entertainment yield none. This limited coverage motivates our focus on Politics and Finance in \Cref{sec:event-type}. The table also reports the number of distinct forecasters and the ranges of group sizes and shared subquestions within each category}
\newsavebox{\apxEventDataTableBox}
\begingroup
\begin{table}[htbp]
\centering
\setlength{\tabcolsep}{2.0pt}
\renewcommand{\arraystretch}{1.08}
\fontsize{8.0}{9.5}\selectfont
\captionsetup{font=normalsize}
\caption{
    Dataset statistics by question category at
    $\groupThreshold=1{,}000$.
}
\label{tab:event-type-data-statistics}
\sbox{\apxEventDataTableBox}{%
\begin{tabular}{lrrrrr}
    \toprule
    Question\\Category
    & \shortstack[r]{Category\\subquestions}
    & \shortstack[r]{Distinct\\forecasters}
    & \shortstack[r]{No.\ of maximal\\groups}
    & \shortstack[r]{Forecasters per group\\(min--max)}
    & \shortstack[r]{Shared subquestions\\per group (min--max)} \\
    \midrule
    Politics
        & $5{,}264$ & $25$ & $12$ & $3$--$9$  & $1{,}003$--$1{,}337$ \\
    Finance
        & $7{,}222$ & $33$ & $9$  & $5$--$16$ & $1{,}085$--$1{,}260$ \\
    Climate
        & $2{,}220$ & $5$  & $2$  & $3$--$3$  & $1{,}275$--$1{,}308$ \\
    Science
        & $459$     & $0$  & $0$  & -- & -- \\
    Sports
        & $1{,}300$ & $0$  & $0$  & -- & -- \\
    Entertainment
        & $512$     & $0$  & $0$  & -- & -- \\
    \bottomrule
\end{tabular}%
}
\ifdim\wd\apxEventDataTableBox>\linewidth
  \resizebox{\linewidth}{!}{\usebox{\apxEventDataTableBox}}%
\else
  \resizebox{\wd\apxEventDataTableBox}{!}{\usebox{\apxEventDataTableBox}}%
\fi
\end{table}
\endgroup

\xhdr{Aggregate performance within categories}
\wtedit{In \Cref{tab:event-type-threshold-1000}, we report the complete group- and pair-level results for learned linear and learned log-odds pooling in Politics and Finance. 
For ECE, we retain the BS-selected benchmark, eligible pairs, and fitted coefficients, so the calibration results evaluate exactly the same aggregates rather than reselecting models or refitting the pooling rules. These results complement the summary in \Cref{sec:event-type} by showing how frequently the category-level patterns appear across individual eligible pairs as well as across comparison groups.}

\begin{table}[h]
\captionsetup{font=normalsize}
\caption{
Pairwise aggregation within Politics and Finance at \(\groupThreshold=1{,}000\). Cells report group- and pair-level rates for aggregates that match or improve on the category-specific BS-best group benchmark under BS or ECE. For ECE, we keep the BS-selected group benchmark, eligible pairs, and fitted coefficients.
}
\label{tab:event-type-threshold-1000}
\resulttablesetup
\resizebox{0.99\linewidth}{!}{
\begin{tabular}{llcccc}
\toprule
& & \multicolumn{2}{c}{BS: match or improve benchmark}
& \multicolumn{2}{c}{ECE: match or improve benchmark} \\
\cmidrule(lr){3-4}\cmidrule(lr){5-6}
Question category & Aggregation rule & Groups & Pairs & Groups & Pairs \\
\midrule

\multirow{2}{*}{Politics}
& Learned linear pooling 
& \shortstack{8/12\\(66.67\%)} 
& \shortstack{23/103\\(22.33\%)} 
& \shortstack{12/12\\(100.00\%)} 
& \shortstack{57/103\\(55.34\%)} \\

& Learned log-odds pooling 
& \shortstack{8/12\\(66.67\%)} 
& \shortstack{17/103\\(16.50\%)} 
& \shortstack{10/12\\(83.33\%)} 
& \shortstack{44/103\\(42.72\%)} \\

\midrule

\multirow{2}{*}{Finance}
& Learned linear pooling 
& \shortstack{9/9\\(100.00\%)} 
& \shortstack{235/287\\(81.88\%)} 
& \shortstack{8/9\\(88.89\%)} 
& \shortstack{281/287\\(97.91\%)} \\

& Learned log-odds pooling 
& \shortstack{5/9\\(55.56\%)} 
& \shortstack{31/287\\(10.80\%)} 
& \shortstack{5/9\\(55.56\%)} 
& \shortstack{42/287\\(14.63\%)} \\

\bottomrule
\end{tabular}
}
\end{table}

\xhdr{Group composition}
In \Cref{tab:event-type-group-membership-n1000}, we report  the complete membership of the Politics and Finance comparison groups used in the analysis. 
The groups differ in both size and model composition, and some forecasters appear in multiple groups.
These details are useful for interpreting the category comparison because the Politics and Finance groups are constructed separately and therefore do not hold the set of forecasters fixed across categories.
Accordingly, the differences reported in \Cref{sec:event-type} should be interpreted descriptively rather than as a causal effect of question category.

\begingroup
\setlength{\tabcolsep}{2.0pt}
\renewcommand{\arraystretch}{1.08}
\fontsize{6.0}{7.0}\selectfont
\begin{longtable}{@{}l l l@{\hspace{10pt}} >{\raggedright\arraybackslash}p{\dimexpr\textwidth-8\tabcolsep-33mm\relax}@{}}
\caption{Complete membership of the size-sorted Politics and Finance groups at the 1,000-common-subquestion threshold.}\label{tab:event-type-group-membership-n1000}

\\
\toprule
Group & Size & Subquestions & Models \\
\midrule
\endfirsthead
\toprule
Group & Size & Subquestions & Models \\
\midrule
\endhead
\midrule
\multicolumn{4}{r}{\fontsize{6.0}{7.0}\selectfont Continued on next page}\\
\endfoot
\bottomrule
\endlastfoot
\multicolumn{4}{@{}l}{\textbf{Politics, threshold 1,000}}\\
\addlinespace[1pt]
1 & 9 & 1,116 & Claude-3-5-Sonnet; Claude-3-7-Sonnet; DeepSeek-R1; DeepSeek-V3; GPT-4o; Llama-3.3-70B-Instruct-Turbo; Mistral-Large; O3-Mini; QwQ-32B-Preview \\
2 & 8 & 1,035 & Claude-3-5-Sonnet; Claude-3-7-Sonnet; GPT-4.1; GPT-4o; Llama-3.3-70B-Instruct-Turbo; Llama-4-Maverick-17B-128E-Instruct-FP8; Llama-4-Scout-17B-16E-Instruct; Mistral-Large \\
3 & 7 & 1,337 & Claude-3-5-Sonnet; Claude-3-Haiku; Claude-3-Opus; GPT-4-Turbo; GPT-4o; Mistral-Large; Qwen3-235B-A22B-Fp8-Tput \\
4 & 6 & 1,129 & Claude-3-5-Sonnet; Claude-3-Haiku; Claude-3-Opus; GPT-4-Turbo; GPT-4o; Llama-3-70b-Chat-Hf \\
5 & 5 & 1,032 & Claude-3-5-Sonnet; Claude-Sonnet-4; GPT-4.1; Kimi-K2-Instruct; Magistral-Medium \\
6 & 5 & 1,011 & Claude-3-5-Sonnet; GPT-4.1; Llama-4-Maverick-17B-128E-Instruct-FP8; Llama-4-Scout-17B-16E-Instruct; O4-Mini \\
7 & 5 & 1,003 & Claude-3-5-Sonnet; GPT-4.1; Kimi-K2-Instruct; Magistral-Medium; Qwen3-235B-A22B-Fp8-Tput \\
8 & 4 & 1,313 & Claude-3-5-Sonnet; GPT-4.1; O4-Mini; Qwen3-235B-A22B-Fp8-Tput \\
9 & 4 & 1,120 & Claude-Sonnet-4; GPT-4.1; Kimi-K2-Instruct; Qwen3-235B-A22B-Fp8-Tput \\
10 & 4 & 1,071 & GPT-4.1; GPT-5-Mini; GPT-5-Nano; Gemini-2.5-Pro \\
11 & 3 & 1,072 & Claude-Opus-4-1; GPT-4.1; Qwen3-235B-A22B-Fp8-Tput \\
12 & 3 & 1,067 & Claude-3-7-Sonnet; GPT-4.1; Qwen3-235B-A22B-Fp8-Tput \\
\midrule
\addlinespace[3pt]
\multicolumn{4}{@{}l}{\textbf{Finance, threshold 1,000}}\\
\addlinespace[1pt]
1 & 16 & 1,170 & Claude-3-5-Sonnet; Claude-3-7-Sonnet; Claude-3-Haiku; Claude-3-Opus; GPT-4-Turbo; GPT-4.1; GPT-4o; Llama-3.2-3B-Instruct-Turbo; Llama-3.3-70B-Instruct-Turbo; Llama-4-Maverick-17B-128E-Instruct-FP8; Llama-4-Scout-17B-16E-Instruct; Meta-Llama-3.1-405B-Instruct-Turbo; Mistral-Large; O4-Mini; Qwen2.5-72B-Instruct-Turbo; Qwen3-235B-A22B-Fp8-Tput \\
2 & 12 & 1,171 & Claude-3-5-Sonnet; Claude-3-7-Sonnet; DeepSeek-R1; DeepSeek-V3; GPT-4o; Gemini-2.0-Flash-Lite-001; Gemini-2.5-Pro-Exp; Grok-beta; Llama-3.3-70B-Instruct-Turbo; Mistral-Large; O3-Mini; QwQ-32B-Preview \\
3 & 11 & 1,191 & Claude-3-5-Sonnet; Claude-3-7-Sonnet; DeepSeek-R1; DeepSeek-V3; GPT-4o; Llama-3.3-70B-Instruct-Turbo; Llama-4-Maverick-17B-128E-Instruct-FP8; Llama-4-Scout-17B-16E-Instruct; Mistral-Large; O3-Mini; QwQ-32B-Preview \\
4 & 8 & 1,182 & Claude-3-5-Sonnet; Claude-3-Haiku; Claude-3-Opus; GPT-4-Turbo; GPT-4o; Llama-3-70b-Chat-Hf; Mistral-Large; Qwen3-235B-A22B-Fp8-Tput \\
5 & 7 & 1,230 & Claude-3-5-Sonnet; Claude-3-Haiku; Claude-3-Opus; GPT-4-Turbo; GPT-4o; Llama-3-70b-Chat-Hf; Llama-3-8b-Chat-Hf \\
6 & 7 & 1,174 & Claude-3-5-Sonnet; Claude-Sonnet-4; GPT-4.1; Kimi-K2-Instruct; Magistral-Medium; O4-Mini; Qwen3-235B-A22B-Fp8-Tput \\
7 & 7 & 1,157 & Claude-3-5-Sonnet; Claude-Opus-4-1; Claude-Sonnet-4; GPT-4.1; Kimi-K2-Instruct; Magistral-Medium; Qwen3-235B-A22B-Fp8-Tput \\
8 & 6 & 1,260 & Claude-3-5-Sonnet; Claude-3-7-Sonnet; DeepSeek-R1; DeepSeek-V3; GPT-4.5-Preview; Llama-3.3-70B-Instruct-Turbo \\
9 & 5 & 1,085 & Claude-Opus-4-1; GPT-4.1; GPT-5-Mini; GPT-5-Nano; Gemini-2.5-Pro \\

\end{longtable}
\endgroup

\xhdr{Best weak-pair aggregates by group}
In \Cref{tab:group-best-individual-top2-pairs-brier-politics-and-finance}, we provide the detailed group-level comparisons underlying the category-specific results. 
For each group, we report the BS-best individual benchmark and the lowest-test-BS weak pair under learned linear and learned log-odds pooling, together with the fitted coefficients and bootstrap confidence intervals. 
This allows us to examine not only whether a category contains successful weak-to-strong aggregates, but also how the identity and performance of the best pair vary across groups.

\begingroup
\setlength{\tabcolsep}{2.0pt}
\renewcommand{\arraystretch}{1.08}
\fontsize{6.0}{7.0}\selectfont
\newcommand{\bestpairline}[1]{%
    \begingroup
    \sbox0{#1}%
    \ifdim\wd0>\linewidth
        \resizebox{\linewidth}{!}{\usebox0}%
    \else
        \usebox0%
    \fi
    \endgroup
}
\newcommand{\brierci}[2]{%
    \begin{tabular}[t]{@{}r@{}}
        #1\\
        $(\pm\,#2)$
    \end{tabular}%
}
\newcommand{\bestpair}[4]{%
    \bestpairline{#1~$+$}\newline
    \bestpairline{#2}\newline
    \bestpairline{$(\omega_1,\omega_2)=(#3,#4)$}%
}
\begin{longtable}{@{}l l@{\hspace{4pt}}
    >{\raggedright\arraybackslash}p{22mm}
    @{\hspace{3pt}}r@{\hspace{4pt}}
    >{\raggedright\arraybackslash}p{34mm}
    @{\hspace{3pt}}r@{\hspace{4pt}}
    >{\raggedright\arraybackslash}p{34mm}
    @{\hspace{3pt}}r@{}
    }
\caption{
\rxedit{
Group benchmark and the best weak-pair aggregate under each pooling method for each \rxedit{of the size-sorted Politics and Finance groups at the 1,000-common-subquestion threshold}, selected by test Brier score. 
}
}\label{tab:group-best-individual-top2-pairs-brier-politics-and-finance}
\\
\toprule
Group & Size & Group Benchmark & \shortstack{Brier\\(95\% CI)} & \shortstack{Log-odds pooling\\best pair} & \shortstack{Brier\\(95\% CI)} & \shortstack{Linear pooling\\best pair} & \shortstack{Brier\\(95\% CI)} \\
\midrule
\endfirsthead
\toprule
Group & Size & Group Benchmark & \shortstack{Brier\\(95\% CI)} & \shortstack{Log-odds pooling\\best pair} & \shortstack{Brier\\(95\% CI)} & \shortstack{Linear pooling\\best pair} & \shortstack{Brier\\(95\% CI)} \\
\midrule
\endhead
\midrule
\multicolumn{8}{r}{\fontsize{6.0}{7.0}\selectfont Continued on next page}\\
\endfoot
\bottomrule
\endlastfoot
\multicolumn{8}{@{}l}{\textbf{Politics, threshold 1,000}}\\
\addlinespace[1pt]
1 & 9 & O3-Mini & \brierci{0.093}{0.017} 
& \bestpair{Mistral-Large}{QwQ-32B-Preview}{0.5087}{0.5308} 
& \brierci{\textbf{0.091}}{0.013} 
& \bestpair{Mistral-Large}{QwQ-32B-Preview}{0.5694}{0.4300} 
& \brierci{\textbf{0.090}}{0.012} \\
2 & 8 & Claude-3-5-Sonnet & \brierci{0.111}{0.017} 
& \bestpair{Claude-3-7-Sonnet}{GPT-4.1}{0.4040}{0.5103} 
& \brierci{\textbf{0.101}}{0.015} 
& \bestpair{GPT-4.1}{Mistral-Large}{0.5201}{0.3794} 
& \brierci{\textbf{0.103}}{0.016} \\
3 & 7 & Claude-3-5-Sonnet & \brierci{0.109}{0.014} 
& \bestpair{Claude-3-Opus}{Mistral-Large}{0.6572}{0.4858} 
& \brierci{\textbf{0.105}}{0.013} 
& \bestpair{Claude-3-Opus}{Mistral-Large}{0.4971}{0.4154} 
& \brierci{\textbf{0.102}}{0.011} \\
4 & 6 & Claude-3-5-Sonnet & \brierci{0.111}{0.017} 
& \bestpair{Claude-3-Opus}{Llama-3-70b-Chat-Hf}{0.8566}{0.2804} 
& \brierci{\textbf{0.105}}{0.014} 
& \bestpair{Claude-3-Opus}{Llama-3-70b-Chat-Hf}{0.7538}{0.1329} 
& \brierci{\textbf{0.104}}{0.012} \\
5 & 5 & GPT-4.1 & \brierci{0.094}{0.017} 
& \bestpair{Kimi-K2-Instruct}{Magistral-Medium}{0.5169}{0.5702} 
& \brierci{\textbf{0.092}}{0.014} 
& \bestpair{Claude-3-5-Sonnet}{Magistral-Medium}{0.6958}{0.2150} 
& \brierci{\textbf{0.093}}{0.013} \\
6 & 5 & O4-Mini & \brierci{0.088}{0.012} 
& \bestpair{Claude-3-5-Sonnet}{GPT-4.1}{0.7774}{0.3183} 
& \brierci{0.093}{0.014} 
& \bestpair{Claude-3-5-Sonnet}{GPT-4.1}{0.6453}{0.2540} 
& \brierci{0.091}{0.012} \\
7 & 5 & GPT-4.1 & \brierci{0.093}{0.018} 
& \bestpair{Kimi-K2-Instruct}{Magistral-Medium}{0.5071}{0.5840} 
& \brierci{\textbf{0.091}}{0.014} 
& \bestpair{Claude-3-5-Sonnet}{Magistral-Medium}{0.6966}{0.2022} 
& \brierci{\textbf{0.091}}{0.013} \\
8 & 4 & O4-Mini & \brierci{0.088}{0.010} 
& \bestpair{Claude-3-5-Sonnet}{GPT-4.1}{0.8719}{0.2802} 
& \brierci{0.093}{0.013} & \bestpair{Claude-3-5-Sonnet}{GPT-4.1}{0.7033}{0.2140} 
& \brierci{0.091}{0.011} \\
9 & 4 & GPT-4.1 & \brierci{0.089}{0.016} 
& \bestpair{Kimi-K2-Instruct}{Qwen3-235B-A22B-Fp8-Tput}{0.5563}{0.6586} 
& \brierci{0.098}{0.014} 
& \bestpair{Kimi-K2-Instruct}{Qwen3-235B-A22B-Fp8-Tput}{0.4537}{0.4680} 
& \brierci{0.096}{0.012} \\
10 & 4 & GPT-5-Mini & \brierci{0.075}{0.014} 
& \bestpair{GPT-4.1}{Gemini-2.5-Pro}{0.3139}{0.7821} 
& \brierci{\textbf{0.074}}{0.014} 
& \bestpair{GPT-4.1}{Gemini-2.5-Pro}{0.2453}{0.7142} 
& \brierci{\textbf{0.074}}{0.013} \\
11 & 3 & GPT-4.1 & \brierci{0.088}{0.017} 
& \bestpair{Claude-Opus-4-1}{Qwen3-235B-A22B-Fp8-Tput}{0.5158}{0.4456} 
& \brierci{\textbf{0.084}}{0.015} 
& \bestpair{Claude-Opus-4-1}{Qwen3-235B-A22B-Fp8-Tput}{0.4742}{0.3733} 
& \brierci{\textbf{0.081}}{0.014} \\
12 & 3 & GPT-4.1 & \brierci{0.109}{0.018} 
& \bestpair{Claude-3-7-Sonnet}{Qwen3-235B-A22B-Fp8-Tput}{0.5874}{0.5654} 
& \brierci{0.112}{0.016} 
& \bestpair{Claude-3-7-Sonnet}{Qwen3-235B-A22B-Fp8-Tput}{0.3616}{0.4461} 
& \brierci{0.112}{0.015} \\
\midrule
\addlinespace[3pt]
\multicolumn{8}{@{}l}{\textbf{Finance, threshold 1,000}}\\
\addlinespace[1pt]
1 & 16 & Llama-3.3-70B-Instruct-Turbo & \brierci{0.211}{0.011} 
& \bestpair{Llama-4-Scout-17B-16E-Instruct}{O4-Mini}{0.3082}{0.8885} 
& \brierci{0.220}{0.011} 
& \bestpair{Claude-3-7-Sonnet}{Llama-4-Scout-17B-16E-Instruct}{0.3486}{0.3024} 
& \brierci{\textbf{0.206}}{0.013} \\
2 & 12 & Llama-3.3-70B-Instruct-Turbo & \brierci{0.210}{0.011} 
& \bestpair{DeepSeek-R1}{QwQ-32B-Preview}{0.2377}{0.3719} 
& \brierci{\textbf{0.199}}{0.012} 
& \bestpair{O3-Mini}{QwQ-32B-Preview}{0.3449}{0.4018} 
& \brierci{\textbf{0.195}}{0.014} \\
3 & 11 & Llama-3.3-70B-Instruct-Turbo & \brierci{0.215}{0.012} 
& \bestpair{DeepSeek-R1}{QwQ-32B-Preview}{0.4337}{0.2798} 
& \brierci{\textbf{0.207}}{0.012} 
& \bestpair{DeepSeek-R1}{QwQ-32B-Preview}{0.5413}{0.2715} 
& \brierci{\textbf{0.202}}{0.013} \\
4 & 8 & Claude-3-Opus & \brierci{0.247}{0.010} 
& \bestpair{Claude-3-5-Sonnet}{Mistral-Large}{0.1765}{0.1133} 
& \brierci{\textbf{0.246}}{0.003} 
& \bestpair{Claude-3-5-Sonnet}{GPT-4o}{0.5378}{0.0000} 
& \brierci{\textbf{0.201}}{0.012} \\
5 & 7 & Claude-3-Opus & \brierci{0.238}{0.009} 
& \bestpair{Claude-3-5-Sonnet}{GPT-4-Turbo}{0.1751}{0.1803} 
& \brierci{0.241}{0.004} 
& \bestpair{Claude-3-5-Sonnet}{Llama-3-70b-Chat-Hf}{0.4631}{0.0836} 
& \brierci{\textbf{0.200}}{0.011} \\
6 & 7 & O4-Mini & \brierci{0.222}{0.009} 
& \bestpair{GPT-4.1}{Magistral-Medium}{0.4252}{0.6643} 
& \brierci{0.225}{0.010} 
& \bestpair{Claude-3-5-Sonnet}{GPT-4.1}{0.4053}{0.1665} 
& \brierci{\textbf{0.204}}{0.012} \\
7 & 7 & Magistral-Medium & \brierci{0.233}{0.010} 
& \bestpair{Claude-Sonnet-4}{GPT-4.1}{0.1522}{0.4135} 
& \brierci{\textbf{0.228}}{0.008} 
& \bestpair{Claude-3-5-Sonnet}{GPT-4.1}{0.4321}{0.1317} 
& \brierci{\textbf{0.206}}{0.012} \\
8 & 6 & Llama-3.3-70B-Instruct-Turbo & \brierci{0.213}{0.010} 
& \bestpair{Claude-3-7-Sonnet}{GPT-4.5-Preview}{0.1910}{0.3798} 
& \brierci{0.237}{0.005} 
& \bestpair{Claude-3-5-Sonnet}{Claude-3-7-Sonnet}{0.3737}{0.1702} 
& \brierci{\textbf{0.205}}{0.011} \\
9 & 5 & GPT-4.1 & \brierci{0.225}{0.011} 
& \bestpair{Claude-Opus-4-1}{Gemini-2.5-Pro}{0.5254}{0.3201} 
& \brierci{\textbf{0.220}}{0.009} 
& \bestpair{Claude-Opus-4-1}{Gemini-2.5-Pro}{0.4559}{0.3650} 
& \brierci{\textbf{0.221}}{0.010} \\

\end{longtable}
\endgroup

\begin{figure}[h]
\centering
\newsavebox{\eventtypeBSPanelARaw}
\newsavebox{\eventtypeBSPanelBRaw}
\newsavebox{\eventtypeBSPanelA}
\newsavebox{\eventtypeBSPanelB}
\sbox{\eventtypeBSPanelARaw}{\resizebox{!}{5.5cm}{
\begin{tikzpicture}
\begin{axis}[
forecast axes front,
    width=6.0cm,
    height=5.0cm,
    y dir=reverse,
    scaled x ticks=false,
    tick align=outside,
    tick pos=left,
    axis x line*=bottom,
    axis y line*=left,
    axis line style={black!65,/pgfplots/on layer=axis foreground},
    tick label style={font=\fontsize{4.5}{5.4}\selectfont},
  x tick label style={
    font=\fontsize{4.5}{5.4}\selectfont,
    /pgf/number format/fixed,
    /pgf/number format/precision=2,
    /pgf/number format/fixed zerofill
},
    y tick label style={font=\fontsize{4.5}{5.4}\selectfont},
    label style={font=\fontsize{5.2}{6.2}\selectfont},
    title style={font=\small},
    xmajorgrids=true,
    grid style={bsgrid,line width=0.35pt},
    unbounded coords=jump,
    xmin=0.07,
    xmax=0.12,
    xtick={0.07,0.08,0.09,0.10,0.11,0.12},
    ylabel={Comparison group},
    ymin=0.4,
    ymax=12.6,
    ytick={1,...,12},
    yticklabels={
        G1,G2,G3,G4,G5,G6,
        G7,G8,G9,G10,G11,G12
    },
    legend to name=eventtypelegend,
    legend style={
        legend columns=1,
        draw=none,
        fill=none,
        font=\fontsize{6}{7}\selectfont,
        inner sep=0pt,
        column sep=2pt,
        row sep=1pt
    },
    legend cell align=left,
]
\path[fill=black!7,fill opacity=0.72,draw=none,/pgfplots/on layer=axis background]
    (axis cs:0.07,1.5) rectangle (axis cs:0.12,2.5)
    (axis cs:0.07,3.5) rectangle (axis cs:0.12,4.5)
    (axis cs:0.07,5.5) rectangle (axis cs:0.12,6.5)
    (axis cs:0.07,7.5) rectangle (axis cs:0.12,8.5)
    (axis cs:0.07,9.5) rectangle (axis cs:0.12,10.5)
    (axis cs:0.07,11.5) rectangle (axis cs:0.12,12.5);
\addplot[
    no marks,
    bsgrid,
    line width=1pt,
    forget plot
] coordinates {
    (0.08998955,1) (0.09349671,1) (nan,nan)
    (0.09349671,0.72) (0.09349671,1) (nan,nan)
    (0.08998955,1) (0.08998955,1.28) (nan,nan)

    (0.10113483,2) (0.11074494,2) (nan,nan)
    (0.11074494,1.72) (0.11074494,2) (nan,nan)
    (0.10285658,2) (0.10285658,2.28) (nan,nan)

    (0.10209654,3) (0.10866883,3) (nan,nan)
    (0.10866883,2.72) (0.10866883,3) (nan,nan)
    (0.10209654,3) (0.10209654,3.28) (nan,nan)

    (0.10381425,4) (0.11074157,4) (nan,nan)
    (0.11074157,3.72) (0.11074157,4) (nan,nan)
    (0.10381425,4) (0.10381425,4.28) (nan,nan)

    (0.09231555,5) (0.09398653,5) (nan,nan)
    (0.09398653,4.72) (0.09398653,5) (nan,nan)
    (0.09298373,5) (0.09298373,5.28) (nan,nan)

    (0.08764833,6) (0.09312854,6) (nan,nan)
    (0.08764833,5.72) (0.08764833,6) (nan,nan)
    (0.09121336,6) (0.09121336,6.28) (nan,nan)

    (0.09081564,7) (0.09279240,7) (nan,nan)
    (0.09279240,6.72) (0.09279240,7) (nan,nan)
    (0.09081564,7) (0.09081564,7.28) (nan,nan)

    (0.08820606,8) (0.09349246,8) (nan,nan)
    (0.08820606,7.72) (0.08820606,8) (nan,nan)
    (0.09074299,8) (0.09074299,8.28) (nan,nan)

    (0.08916792,9) (0.09773953,9) (nan,nan)
    (0.08916792,8.72) (0.08916792,9) (nan,nan)
    (0.09552431,9) (0.09552431,9.28) (nan,nan)

    (0.07378488,10) (0.07537446,10) (nan,nan)
    (0.07537446,9.72) (0.07537446,10) (nan,nan)
    (0.07378488,10) (0.07378488,10.28) (nan,nan)

    (0.08080061,11) (0.08808537,11) (nan,nan)
    (0.08808537,10.72) (0.08808537,11) (nan,nan)
    (0.08080061,11) (0.08080061,11.28) (nan,nan)

    (0.10884647,12) (0.11171556,12) (nan,nan)
    (0.10884647,11.72) (0.10884647,12) (nan,nan)
    (0.11171556,12) (0.11171556,12.28) (nan,nan)
};

\addplot[
    only marks,
    bsindividual,
    forget plot,
    mark=*,
    mark size=1.6pt
] coordinates {
    (0.09349671,0.72)
    (0.11074494,1.72)
    (0.10866883,2.72)
    (0.11074157,3.72)
    (0.09398653,4.72)
    (0.08764833,5.72)
    (0.09279240,6.72)
    (0.08820606,7.72)
    (0.08916792,8.72)
    (0.07537446,9.72)
    (0.08808537,10.72)
    (0.10884647,11.72)
};

\addplot[
    only marks,
    bslogodds,
    forget plot,
    mark=diamond*,
    mark size=1.9pt
] coordinates {
    (0.09094779,1)
    (0.10113483,2)
    (0.10549153,3)
    (0.10502191,4)
    (0.09231555,5)
    (0.09312854,6)
    (0.09092709,7)
    (0.09349246,8)
    (0.09773953,9)
    (0.07392310,10)
    (0.08441964,11)
    (0.11164805,12)
};

\addplot[
    only marks,
    bslinear,
    forget plot,
    mark=square*,
    mark size=1.4pt
] coordinates {
    (0.08998955,1.28)
    (0.10285658,2.28)
    (0.10209654,3.28)
    (0.10381425,4.28)
    (0.09298373,5.28)
    (0.09121336,6.28)
    (0.09081564,7.28)
    (0.09074299,8.28)
    (0.09552431,9.28)
    (0.07378488,10.28)
    (0.08080061,11.28)
    (0.11171556,12.28)
};

\addlegendimage{only marks,bsindividual,mark=*,mark size=1.6pt}
\addlegendentry{\shortstack[l]{Group\\benchmark}}
\addlegendimage{only marks,bslinear,mark=square*,mark size=1.4pt}
\addlegendentry{\shortstack[l]{Learned linear\\ pooling}}
\addlegendimage{only marks,bslogodds,mark=diamond*,mark size=1.9pt}
\addlegendentry{\shortstack[l]{Learned log-odds\\ pooling}}
\end{axis}
\end{tikzpicture}}}
\sbox{\eventtypeBSPanelBRaw}{\resizebox{!}{5.5cm}{

\begin{tikzpicture}
\begin{axis}[
forecast axes front,
width=6.0cm,
    height=5.0cm,
    y dir=reverse,
    scaled x ticks=false,
    tick align=outside,
    tick pos=left,
    axis x line*=bottom,
    axis y line*=left,
    axis line style={black!65,/pgfplots/on layer=axis foreground},
    tick label style={font=\fontsize{4.5}{5.4}\selectfont},
    x tick label style={
        font=\fontsize{4.5}{5.4}\selectfont,
        /pgf/number format/fixed,
        /pgf/number format/precision=2
    },
    y tick label style={font=\fontsize{4.5}{5.4}\selectfont},
    label style={font=\fontsize{5.2}{6.2}\selectfont},
    title style={font=\small},
    xmajorgrids=true,
    grid style={bsgrid,line width=0.35pt},
    unbounded coords=jump,
    xmin=0.19,
    xmax=0.25,
    xtick={0.19,0.21,0.23,0.25},
    ymin=0.4,
    ymax=9.6,
    ytick={1,...,9},
    yticklabels={G1,G2,G3,G4,G5,G6,G7,G8,G9}
]
\path[fill=black!7,fill opacity=0.72,draw=none,/pgfplots/on layer=axis background]
    (axis cs:0.19,1.5) rectangle (axis cs:0.25,2.5)
    (axis cs:0.19,3.5) rectangle (axis cs:0.25,4.5)
    (axis cs:0.19,5.5) rectangle (axis cs:0.25,6.5)
    (axis cs:0.19,7.5) rectangle (axis cs:0.25,8.5);
\addplot[
    no marks,
    bsgrid,
    line width=1pt,
    forget plot
] coordinates {
    (0.20560051,1) (0.21952203,1) (nan,nan)
    (0.21134516,0.72) (0.21134516,1) (nan,nan)
    (0.20560051,1) (0.20560051,1.28) (nan,nan)

    (0.19521397,2) (0.21035470,2) (nan,nan)
    (0.21035470,1.72) (0.21035470,2) (nan,nan)
    (0.19521397,2) (0.19521397,2.28) (nan,nan)

    (0.20164503,3) (0.21507863,3) (nan,nan)
    (0.21507863,2.72) (0.21507863,3) (nan,nan)
    (0.20164503,3) (0.20164503,3.28) (nan,nan)

    (0.20124646,4) (0.24696408,4) (nan,nan)
    (0.24696408,3.72) (0.24696408,4) (nan,nan)
    (0.20124646,4) (0.20124646,4.28) (nan,nan)

    (0.19951936,5) (0.24133249,5) (nan,nan)
    (0.23761423,4.72) (0.23761423,5) (nan,nan)
    (0.19951936,5) (0.19951936,5.28) (nan,nan)

    (0.20368020,6) (0.22454560,6) (nan,nan)
    (0.22241754,5.72) (0.22241754,6) (nan,nan)
    (0.20368020,6) (0.20368020,6.28) (nan,nan)

    (0.20578994,7) (0.23279780,7) (nan,nan)
    (0.23279780,6.72) (0.23279780,7) (nan,nan)
    (0.20578994,7) (0.20578994,7.28) (nan,nan)

    (0.20507890,8) (0.23749093,8) (nan,nan)
    (0.21269312,7.72) (0.21269312,8) (nan,nan)
    (0.20507890,8) (0.20507890,8.28) (nan,nan)

    (0.22049128,9) (0.22538345,9) (nan,nan)
    (0.22538345,8.72) (0.22538345,9) (nan,nan)
    (0.22130595,9) (0.22130595,9.28) (nan,nan)
};
\addplot[
    only marks,
    bsindividual,
    mark=*,
    mark size=1.6pt
] coordinates {
    (0.21134516,0.72)
    (0.21035470,1.72)
    (0.21507863,2.72)
    (0.24696408,3.72)
    (0.23761423,4.72)
    (0.22241754,5.72)
    (0.23279780,6.72)
    (0.21269312,7.72)
    (0.22538345,8.72)
};
\addplot[
    only marks,
    bslogodds,
    mark=diamond*,
    mark size=1.9pt
] coordinates {
    (0.21952203,1)
    (0.19882391,2)
    (0.20688645,3)
    (0.24601119,4)
    (0.24133249,5)
    (0.22454560,6)
    (0.22842176,7)
    (0.23749093,8)
    (0.22049128,9)
};
\addplot[
    only marks,
    bslinear,
    mark=square*,
    mark size=1.4pt
] coordinates {
    (0.20560051,1.28)
    (0.19521397,2.28)
    (0.20164503,3.28)
    (0.20124646,4.28)
    (0.19951936,5.28)
    (0.20368020,6.28)
    (0.20578994,7.28)
    (0.20507890,8.28)
    (0.22130595,9.28)
};

\end{axis}
\end{tikzpicture}}}
\pgfmathsetmacro{\eventtypeBSScale}{0.82*\the\linewidth/(\the\wd\eventtypeBSPanelARaw+\the\wd\eventtypeBSPanelBRaw)}
\sbox{\eventtypeBSPanelA}{\scalebox{\eventtypeBSScale}{\usebox{\eventtypeBSPanelARaw}}}
\sbox{\eventtypeBSPanelB}{\scalebox{\eventtypeBSScale}{\usebox{\eventtypeBSPanelBRaw}}}
\noindent
\begin{minipage}[t]{\wd\eventtypeBSPanelA}
\vspace{0pt}\centering
\hspace*{22pt}
\subfloat[Politics: BS.%
\label{fig:event-type-politics-bs}]{%
    \hspace*{-22pt}\usebox{\eventtypeBSPanelA}%
}
\end{minipage}\hfill%
\begin{minipage}[t]{\wd\eventtypeBSPanelB}
\vspace{0pt}\centering
\hspace*{6pt}
\subfloat[Finance: BS.%
\label{fig:event-type-finance-bs}]{%
    \hspace*{-6pt}\usebox{\eventtypeBSPanelB}%
}
\end{minipage}\hfill%
\begin{minipage}[t]{0.16\linewidth}
\vspace{0pt}\raggedright
\resizebox{0.85\linewidth}{!}{\pgfplotslegendfromname{eventtypelegend}}
\end{minipage}
\caption{Test BS of the BS-best group benchmark and the lowest-test-BS pair under each learned rule in \Cref{fig:event-type-politics-bs} Politics and \Cref{fig:event-type-finance-bs} Finance.
We select the pair separately for learned linear and learned log-odds pooling in each category-specific comparison group. Lower BS is better.
The pale horizontal segments connect the three point estimates within a group; they are not confidence intervals.
Group labels are local to each category and ordered by decreasing group size.}
\label{fig:event-type-size-sorted-groups}
\end{figure}
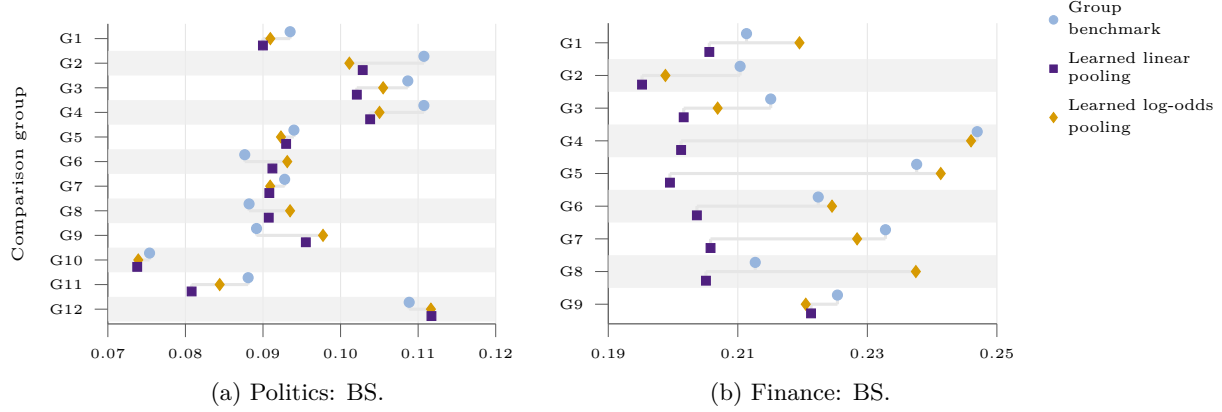

In \Cref{fig:event-type-size-sorted-groups}, we visualize the comparisons in \Cref{tab:group-best-individual-top2-pairs-brier-politics-and-finance}. 
The figure makes the contrast between the two categories more transparent: learned linear pooling performs consistently close to or better than the benchmark across the Finance groups, whereas the Politics results are more heterogeneous. 
The performance of learned log-odds pooling also varies more substantially across groups in both categories. 
Because the groups differ in both questions and constituent forecasters, these patterns should again be interpreted as descriptive differences across the resulting comparison groups.

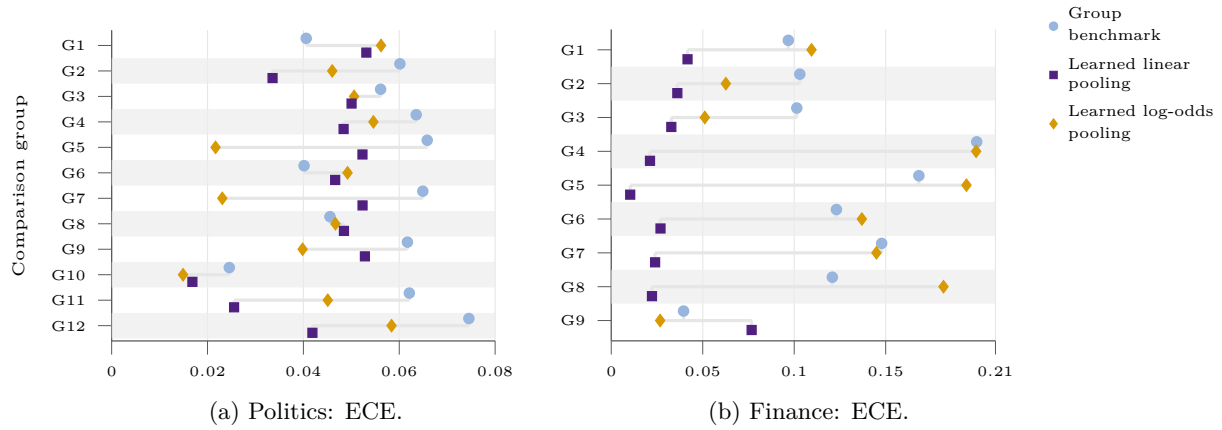
\begin{figure}[h]
\centering

\newsavebox{\eventtypeECEPanelARaw}
\newsavebox{\eventtypeECEPanelBRaw}
\newsavebox{\eventtypeECEPanelA}
\newsavebox{\eventtypeECEPanelB}

\sbox{\eventtypeECEPanelARaw}{%
    \resizebox{!}{5.5cm}{%

\begin{tikzpicture}
\begin{axis}[
forecast axes front,
width=5.8cm,
    height=5.0cm,
    y dir=reverse,
    scaled x ticks=false,
    tick align=outside,
    tick pos=left,
    axis x line*=bottom,
    axis y line*=left,
    axis line style={black!65,/pgfplots/on layer=axis foreground},
    tick label style={font=\fontsize{4.5}{5.4}\selectfont},
    x tick label style={
        font=\fontsize{4.5}{5.4}\selectfont,
        /pgf/number format/fixed,
        /pgf/number format/precision=2
    },
    y tick label style={font=\fontsize{4.5}{5.4}\selectfont},
    label style={font=\fontsize{5.2}{6.2}\selectfont},
    title style={font=\small},
    xmajorgrids=true,
    grid style={bsgrid,line width=0.35pt},
    unbounded coords=jump,
    xmin=0.00,
    xmax=0.08,
    xtick={0.00,0.02,0.04,0.06,0.08},
    ylabel={Comparison group},
    ymin=0.4,
    ymax=12.6,
    ytick={1,...,12},
    yticklabels={
        G1,G2,G3,G4,G5,G6,
        G7,G8,G9,G10,G11,G12
    }
]
\path[fill=black!7,fill opacity=0.72,draw=none,/pgfplots/on layer=axis background]
    (axis cs:0.00,1.5) rectangle (axis cs:0.08,2.5)
    (axis cs:0.00,3.5) rectangle (axis cs:0.08,4.5)
    (axis cs:0.00,5.5) rectangle (axis cs:0.08,6.5)
    (axis cs:0.00,7.5) rectangle (axis cs:0.08,8.5)
    (axis cs:0.00,9.5) rectangle (axis cs:0.08,10.5)
    (axis cs:0.00,11.5) rectangle (axis cs:0.08,12.5);
\addplot[
    no marks,
    bsgrid,
    line width=1pt,
    forget plot
] coordinates {
    (0.04058933,1) (0.05620720,1) (nan,nan)
    (0.04058933,0.72) (0.04058933,1) (nan,nan)
    (0.05313544,1) (0.05313544,1.28) (nan,nan)

    (0.03356657,2) (0.06014036,2) (nan,nan)
    (0.06014036,1.72) (0.06014036,2) (nan,nan)
    (0.03356657,2) (0.03356657,2.28) (nan,nan)

    (0.05004981,3) (0.05611024,3) (nan,nan)
    (0.05611024,2.72) (0.05611024,3) (nan,nan)
    (0.05004981,3) (0.05004981,3.28) (nan,nan)

    (0.04838793,4) (0.06353696,4) (nan,nan)
    (0.06353696,3.72) (0.06353696,4) (nan,nan)
    (0.04838793,4) (0.04838793,4.28) (nan,nan)

    (0.02168024,5) (0.06582939,5) (nan,nan)
    (0.06582939,4.72) (0.06582939,5) (nan,nan)
    (0.05232604,5) (0.05232604,5.28) (nan,nan)

    (0.04014146,6) (0.04921755,6) (nan,nan)
    (0.04014146,5.72) (0.04014146,6) (nan,nan)
    (0.04663144,6) (0.04663144,6.28) (nan,nan)

    (0.02308743,7) (0.06491516,7) (nan,nan)
    (0.06491516,6.72) (0.06491516,7) (nan,nan)
    (0.05234098,7) (0.05234098,7.28) (nan,nan)

    (0.04554761,8) (0.04846248,8) (nan,nan)
    (0.04554761,7.72) (0.04554761,8) (nan,nan)
    (0.04846248,8) (0.04846248,8.28) (nan,nan)

    (0.03984026,9) (0.06170483,9) (nan,nan)
    (0.06170483,8.72) (0.06170483,9) (nan,nan)
    (0.05283612,9) (0.05283612,9.28) (nan,nan)

    (0.01489565,10) (0.02454583,10) (nan,nan)
    (0.02454583,9.72) (0.02454583,10) (nan,nan)
    (0.01684766,10) (0.01684766,10.28) (nan,nan)

    (0.02554234,11) (0.06210154,11) (nan,nan)
    (0.06210154,10.72) (0.06210154,11) (nan,nan)
    (0.02554234,11) (0.02554234,11.28) (nan,nan)

    (0.04188370,12) (0.07452425,12) (nan,nan)
    (0.07452425,11.72) (0.07452425,12) (nan,nan)
    (0.04188370,12) (0.04188370,12.28) (nan,nan)
};
\addplot[
    only marks,
    bsindividual,
    mark=*,
    mark size=1.6pt
] coordinates {
    (0.04058933,0.72)
    (0.06014036,1.72)
    (0.05611024,2.72)
    (0.06353696,3.72)
    (0.06582939,4.72)
    (0.04014146,5.72)
    (0.06491516,6.72)
    (0.04554761,7.72)
    (0.06170483,8.72)
    (0.02454583,9.72)
    (0.06210154,10.72)
    (0.07452425,11.72)
};
\addplot[
    only marks,
    bslogodds,
    mark=diamond*,
    mark size=1.9pt
] coordinates {
    (0.05620720,1)
    (0.04601923,2)
    (0.05058305,3)
    (0.05460974,4)
    (0.02168024,5)
    (0.04921755,6)
    (0.02308743,7)
    (0.04668758,8)
    (0.03984026,9)
    (0.01489565,10)
    (0.04509152,11)
    (0.05839480,12)
};
\addplot[
    only marks,
    bslinear,
    mark=square*,
    mark size=1.4pt
] coordinates {
    (0.05313544,1.28)
    (0.03356657,2.28)
    (0.05004981,3.28)
    (0.04838793,4.28)
    (0.05232604,5.28)
    (0.04663144,6.28)
    (0.05234098,7.28)
    (0.04846248,8.28)
    (0.05283612,9.28)
    (0.01684766,10.28)
    (0.02554234,11.28)
    (0.04188370,12.28)
};

\end{axis}
\end{tikzpicture}}}
\sbox{\eventtypeECEPanelBRaw}{%
    \resizebox{!}{5.5cm}{%

\begin{tikzpicture}
\begin{axis}[
forecast axes front,
    width=5.8cm,
    height=5.0cm,
    y dir=reverse,
    scaled x ticks=false,
    tick align=outside,
    tick pos=left,
    axis x line*=bottom,
    axis y line*=left,
    axis line style={black!65,/pgfplots/on layer=axis foreground},
    tick label style={font=\fontsize{4.5}{5.4}\selectfont},
    x tick label style={
        font=\fontsize{4.5}{5.4}\selectfont,
        /pgf/number format/fixed,
        /pgf/number format/precision=2
    },
    y tick label style={font=\fontsize{4.5}{5.4}\selectfont},
    label style={font=\fontsize{5.2}{6.2}\selectfont},
    title style={font=\small},
    xmajorgrids=true,
    grid style={bsgrid,line width=0.35pt},
    unbounded coords=jump,
    xmin=0.00,
    xmax=0.21,
    xtick={0.00,0.05,0.10,0.15,0.21},
    ymin=0.4,
    ymax=9.6,
    ytick={1,...,9},
    yticklabels={G1,G2,G3,G4,G5,G6,G7,G8,G9}
]
\path[fill=black!7,fill opacity=0.72,draw=none,/pgfplots/on layer=axis background]
    (axis cs:0.00,1.5) rectangle (axis cs:0.21,2.5)
    (axis cs:0.00,3.5) rectangle (axis cs:0.21,4.5)
    (axis cs:0.00,5.5) rectangle (axis cs:0.21,6.5)
    (axis cs:0.00,7.5) rectangle (axis cs:0.21,8.5);
\addplot[
    no marks,
    bsgrid,
    line width=1pt,
    forget plot
] coordinates {
    (0.04169452,1) (0.10946338,1) (nan,nan)
    (0.09677279,0.72) (0.09677279,1) (nan,nan)
    (0.04169452,1) (0.04169452,1.28) (nan,nan)

    (0.03612729,2) (0.10306291,2) (nan,nan)
    (0.10306291,1.72) (0.10306291,2) (nan,nan)
    (0.03612729,2) (0.03612729,2.28) (nan,nan)

    (0.03283744,3) (0.10142069,3) (nan,nan)
    (0.10142069,2.72) (0.10142069,3) (nan,nan)
    (0.03283744,3) (0.03283744,3.28) (nan,nan)

    (0.02118196,4) (0.19975404,4) (nan,nan)
    (0.19975404,3.72) (0.19975404,4) (nan,nan)
    (0.02118196,4) (0.02118196,4.28) (nan,nan)

    (0.01036905,5) (0.19414363,5) (nan,nan)
    (0.16815098,4.72) (0.16815098,5) (nan,nan)
    (0.01036905,5) (0.01036905,5.28) (nan,nan)

    (0.02693006,6) (0.13698142,6) (nan,nan)
    (0.12307994,5.72) (0.12307994,6) (nan,nan)
    (0.02693006,6) (0.02693006,6.28) (nan,nan)

    (0.02397753,7) (0.14779948,7) (nan,nan)
    (0.14779948,6.72) (0.14779948,7) (nan,nan)
    (0.02397753,7) (0.02397753,7.28) (nan,nan)

    (0.02225986,8) (0.18160762,8) (nan,nan)
    (0.12083472,7.72) (0.12083472,8) (nan,nan)
    (0.02225986,8) (0.02225986,8.28) (nan,nan)

    (0.02673915,9) (0.07672938,9) (nan,nan)
    (0.03953812,8.72) (0.03953812,9) (nan,nan)
    (0.07672938,9) (0.07672938,9.28) (nan,nan)
};

\addplot[
    only marks,
    bsindividual,
    mark=*,
    mark size=1.6pt
] coordinates {
    (0.09677279,0.72)
    (0.10306291,1.72)
    (0.10142069,2.72)
    (0.19975404,3.72)
    (0.16815098,4.72)
    (0.12307994,5.72)
    (0.14779948,6.72)
    (0.12083472,7.72)
    (0.03953812,8.72)
};

\addplot[
    only marks,
    bslogodds,
    mark=diamond*,
    mark size=1.9pt
] coordinates {
    (0.10946338,1)
    (0.06264481,2)
    (0.05115217,3)
    (0.19943137,4)
    (0.19414363,5)
    (0.13698142,6)
    (0.14494739,7)
    (0.18160762,8)
    (0.02673915,9)
};

\addplot[
    only marks,
    bslinear,
    mark=square*,
    mark size=1.4pt
] coordinates {
    (0.04169452,1.28)
    (0.03612729,2.28)
    (0.03283744,3.28)
    (0.02118196,4.28)
    (0.01036905,5.28)
    (0.02693006,6.28)
    (0.02397753,7.28)
    (0.02225986,8.28)
    (0.07672938,9.28)
};

\end{axis}
\end{tikzpicture}}}

\pgfmathsetmacro{\eventtypeECEScale}{0.82*\the\linewidth/(\the\wd\eventtypeECEPanelARaw+\the\wd\eventtypeECEPanelBRaw)}
\sbox{\eventtypeECEPanelA}{%
    \scalebox{\eventtypeECEScale}{\usebox{\eventtypeECEPanelARaw}}}
\sbox{\eventtypeECEPanelB}{%
    \scalebox{\eventtypeECEScale}{\usebox{\eventtypeECEPanelBRaw}}}

\noindent
\begin{minipage}[t]{\wd\eventtypeECEPanelA}
\vspace{0pt}\centering
\hspace*{28pt}%
\subfloat[Politics: ECE.%
\label{fig:event-type-politics-ece}]{%
    \hspace*{-28pt}\usebox{\eventtypeECEPanelA}%
}
\end{minipage}\hfill%
\begin{minipage}[t]{\wd\eventtypeECEPanelB}
\vspace{0pt}\centering
\hspace*{12pt}%
\subfloat[Finance: ECE.%
\label{fig:event-type-finance-ece}]{%
    \hspace*{-12pt}\usebox{\eventtypeECEPanelB}%
}
\end{minipage}\hfill%
\begin{minipage}[t]{0.16\linewidth}
\vspace{0pt}\raggedright
\resizebox{0.85\linewidth}{!}{%
    \pgfplotslegendfromname{eventtypelegend}}
\end{minipage}
\caption{
Test ECE of the BS-best individual and the learned linear and
log-odds pairs in the Politics and Finance comparison groups.
}
\label{fig:event-type-ece-groups}
\end{figure}

\xhdr{Calibration of the BS-selected aggregates}
Lastly, in \Cref{fig:event-type-ece-groups}, we evaluate the same group benchmarks and rule-specific weak pairs selected by test BS in \Cref{fig:event-type-size-sorted-groups} using ECE, without reselection or refitting. 
We observe a broadly similar category pattern for calibration: learned linear pooling performs particularly well in Finance, where the pair-level ECE success rate is substantially higher than in Politics. 
This mirrors the corresponding BS comparison and suggests that the stronger aggregation gains observed in Finance are generally accompanied by improved calibration as well.

\end{document}